\documentclass[10pt,twocolumn,letterpaper]{article}

\usepackage{cvpr}              

\definecolor{cvprblue}{rgb}{0.21,0.49,0.74}
\usepackage[pagebackref,breaklinks,colorlinks,allcolors=cvprblue]{hyperref}

\usepackage{amsmath, amssymb, amsthm}
\usepackage{graphicx}
\usepackage{diagbox}
\usepackage{hyperref}
\usepackage{multirow}
\usepackage{makecell}
\usepackage{tabularx}

\usepackage{bbding}
\usepackage{array}
\usepackage{booktabs}
\usepackage{hyperref}

\usepackage{algorithm}
\usepackage{algorithmic}
\usepackage{booktabs}
\usepackage{multirow}
\usepackage{amsmath}
\usepackage{amssymb}
\usepackage{colortbl}
\usepackage{xcolor}
\usepackage{graphicx}  
\usepackage{float}  
\usepackage[accsupp]{axessibility}  
\usepackage{tablefootnote}

\def\paperID{2431} 
\def\confName{CVPR}
\def\confYear{2025}

\title{OmniDocBench: Benchmarking Diverse PDF Document Parsing with Comprehensive Annotations\vspace{-1.5em}}

\author{
Linke Ouyang$^{1*}$\quad
Yuan Qu$^{1*}$\quad
Hongbin Zhou$^{1*}$\quad
Jiawei Zhu$^{1*}$\quad
Rui Zhang$^{1*}$\quad
Qunshu Lin$^{2*}$\quad\\
Bin Wang$^{1*\dag}$\quad
Zhiyuan Zhao$^{1}$
Man Jiang$^{1}$
Xiaomeng Zhao$^{1}$
Jin Shi$^{1}$
Fan Wu$^{1}$
Pei Chu$^{1}$
Minghao Liu$^{3}$\\
Zhenxiang Li$^{1}$
Chao Xu$^{1}$
Bo Zhang$^{1}$
Botian Shi$^{1}$
Zhongying Tu$^{1}$
Conghui He$^{1\ddagger}$\\
{\small $^1$Shanghai AI Laboratory\quad
$^2$Abaka AI\quad
$^3$2077AI\quad}
}

\begin{document}
\maketitle
\renewcommand{\thefootnote}{}
\footnotetext{$^*$ The authors contributed equally.}
\footnotetext{$^\dag$ Project lead.}
\footnotetext{$^\ddagger$ Corresponding author (heconghui@pjlab.org.cn).}
\renewcommand{\thefootnote}{\arabic{footnote}}
\begin{abstract}

Document content extraction is a critical task in computer vision, underpinning the data needs of large language models (LLMs) and retrieval-augmented generation (RAG) systems. 
Despite recent progress, current document parsing methods have not been fairly and comprehensively evaluated due to the narrow coverage of document types and the simplified, unrealistic evaluation procedures in existing benchmarks. 
To address these gaps, we introduce OmniDocBench, a novel benchmark featuring high-quality annotations across nine document sources, including academic papers, textbooks, and more challenging cases such as handwritten notes and densely typeset newspapers. 
OmniDocBench supports flexible, multi-level evaluations—ranging from an end-to-end assessment to the task-specific and attribute-based analysis—using 19 layout categories and 15 attribute labels. 
We conduct a thorough evaluation of both pipeline-based methods and end-to-end vision-language models, revealing their strengths and weaknesses across different document types. 
OmniDocBench sets a new standard for the fair, diverse, and fine-grained evaluation in document parsing. Dataset and code are available at \url{https://github.com/opendatalab/OmniDocBench}.

\end{abstract}
    
\section{Introduction}
\label{sec:intro}
\begin{figure}[h]
    \centering
    \includegraphics[width=0.87\linewidth]{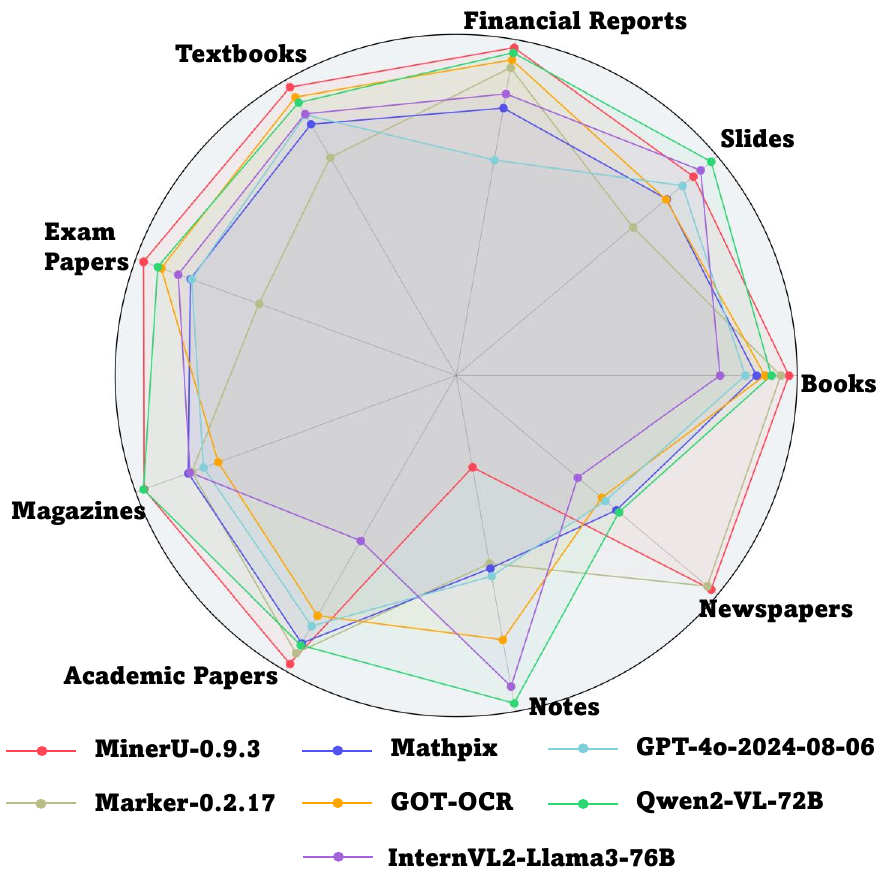}
    \caption{Results of End-to-End Text Recognition on OmniDocBench across 9 PDF page types.}
    \label{fig:radar}
\end{figure}

\begin{figure*}[t]
    \centering
    \includegraphics[width=0.9\textwidth]{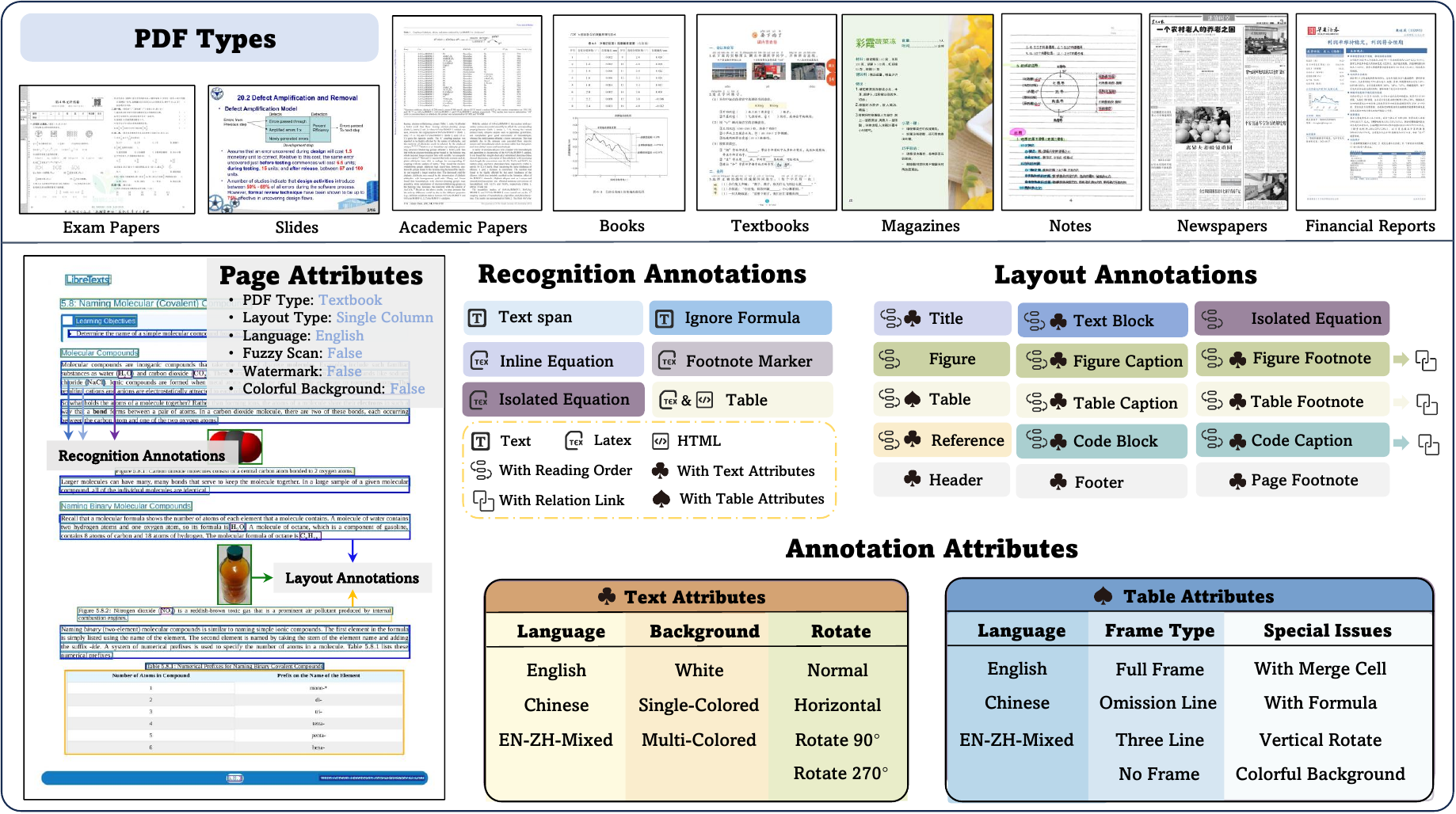}
    \caption{Overview of OmniDocBench Data Diversity. The benchmark includes 9 diverse PDF document types. It supports rich annotation types, including layout annotations (e.g., title, table, figure) and recognition annotations (e.g., text spans, equations, tables). Each page is annotated with 6 page-level attributes (e.g., PDF type, layout type), along with fine-grained 3 text attributes (e.g., language) and 6 tables attributes (Items under ``\textit{special issues}'' are treated as individual binary attributes (yes/no)), enabling detailed and robust evaluation.}
    \label{fig:data_diversity}
\end{figure*}

As large language models~\cite{GPT4,touvron2023llama,wang2024qwen2,liu2024deepseek} increasingly rely on high-quality, knowledge-rich data, the importance of accurate document parsing has grown substantially. Document parsing, a core task in computer vision and document intelligence, aims to extract structured, machine-readable content from unstructured documents such as PDFs. This task is particularly critical for ingesting academic papers, technical reports, textbooks, and other rich textual sources into large language models, thereby enhancing their factual accuracy and knowledge grounding~\cite{DocumentParsing1,MinerU,got2,xia2024docgenome,zhang2024document}.
Moreover, with the emergence of retrieval-augmented generation (RAG) systems~\cite{RAG2,RAG}, which retrieve and  generate answers conditionally with external documents, the demand for precise document understanding has further intensified. 

To address this challenging task, two main paradigms have emerged: 1) Pipeline-based approaches that decompose the task into layout analysis, OCR, formula/table recognition, and reading order estimation~\cite{MinerU,marker}; and 2) End-to-end vision-language models (VLMs) that 
directly output structured representations (e.g., Markdown)~\cite{Nougat,Vary,fox,got2,Qwen-VL,InternVL,xia2024chartx}. Although both approaches have demonstrated promising results, conducting a broad comparison of their effectiveness remains challenging due to the absence of a comprehensive and unified evaluation benchmark.

As shown in Table~\ref{tab:related_work}, for pipeline-based document parsing systems, dedicated benchmarks~\cite{DocBank,PubTabNet,Im2Latex-100K} have been developed to target specific sub-tasks. 
For end-to-end evaluation, works like Nougat~\cite{Nougat} and GOT-OCR~\cite{got2} provide relatively small validation sets and assess predictions using page-level metrics such as 
Edit Distance~\cite{levenshtein1966binary}.

However, these benchmarks present several key limitations:
1) \textbf{Limited document diversity:} Existing datasets primarily focus on academic papers, overlooking other real-world document types such as textbooks, exams, financial reports, and newspapers; 
2) \textbf{Inconsistent evaluation metrics:} Current benchmarks rely heavily on generic text similarity metrics (e.g., Edit Distance~\cite{levenshtein1966binary} and BLEU~\cite{papineni2002bleu}), which fail to fairly assess the accuracy of formulas and tables in LaTeX or HTML formats that allow for diverse syntactic expressions; and
3) \textbf{Lack of fine-grained evaluation:} Most evaluations report only an overall score, lacking insights into specific weaknesses, such as element-level score (e.g., text vs. formula) or per document-type performance 
(e.g., magazine or notes).


To address these limitations, we introduce \textbf{OmniDocBench}, a new benchmark designed to provide a rigorous and comprehensive evaluation for document parsing models across both pipeline-based and end-to-end paradigms.
In summary, our benchmark introduces the following key contributions:

\begin{itemize} \item \textbf{High-quality, diverse evaluation set:} We include pages from 9 distinct document types, ranging from textbooks to newspapers, annotated using a combination of automated tools, manual verification, and expert review. 
\item \textbf{Flexible, multi-dimensional evaluation:} 
We support comprehensive evaluation at three levels—end-to-end, task-specific, and attribute-based. End-to-end evaluation measures the overall quality of full-page parsing results. Task-specific evaluation allows users to assess individual components such as layout detection, OCR, table recognition, or formula parsing. Attribute-based evaluation provides fine-grained analysis across 9 document types, 6 page-level attributes and 9 bbox-level attributes. 
\item \textbf{Comprehensive benchmarking of state-of-the-art methods:} We systematically evaluate a suite of representative document parsing systems, including both pipeline-based tools and VLMs, providing the most comprehensive comparison and identifying performance bottlenecks across document types and content structures. \end{itemize}

\section{Related Work}

\begin{table*}[t]
\centering
\resizebox{0.9\textwidth}{!}{
\begin{tabular}{l|c|ccccc|cccc|cccc}
\toprule
\textbf{\multirow{2}{*}{Benchmark}} & \multirow{2}{*}{\makecell{\textbf{Document}\\\textbf{Domain}}} & \multicolumn{5}{c}{\textbf{Annotaion Type}} & \multicolumn{4}{c}{\textbf{Single-Task Eval}} & \multicolumn{4}{c}{\textbf{End-to-End Eval}} \\
& & BBox & Text & Table & Formula & Attributes & OCR & DLA & TR &  MFR & OCR & TR & MFR & ROD \\
\midrule
\multicolumn{15}{l}{\textbf{\textit{Single-Task Eval Benchmark}}} \\
\midrule
Robust Reading~\cite{RobustReading} & 1 & \CheckmarkBold & \CheckmarkBold & & & & \CheckmarkBold & & & & & &   \\
\midrule
\makecell[l]{PubLayNet~\cite{PubLayNet}, DocBank~\cite{DocBank}, \\DocLayNet~\cite{DocLayNet}, M\textsuperscript{6}Doc~\cite{M6Doc}} & \makecell[c]{1, 1, \\5, 6} & \CheckmarkBold & & & & & & \CheckmarkBold & & & & & &  \\
\midrule
PubTabNet~\cite{PubTabNet},TableX~\cite{TabLeX} & 1, 1 &  & & \CheckmarkBold & &  & & & \CheckmarkBold & & & & &  \\
TableBank~\cite{Tablebank} & 1 & \CheckmarkBold & & \CheckmarkBold & & &  &  & \CheckmarkBold & & & &  \\
\midrule
Im2Latex-100K~\cite{Im2Latex-100K},UniMER-Test~\cite{wang2024unimernetuniversalnetworkrealworld} & 1 &  &  & & \CheckmarkBold &  &  &  & & \CheckmarkBold &  & \\
\midrule
\multicolumn{15}{l}{\textbf{\textit{End-to-end Eval Benchmarks}}} \\
\midrule
Fox~\cite{fox} & 2 & \CheckmarkBold & \CheckmarkBold &  &  & & \CheckmarkBold &  & &  & \CheckmarkBold &  &  &  \\
Nougat~\cite{Nougat} & 1 &  & \CheckmarkBold & \CheckmarkBold & \CheckmarkBold &  &  &  & &  & \CheckmarkBold & \CheckmarkBold & \CheckmarkBold  &  \\
GOT OCR 2.0~\cite{got2} & 2 &  & \CheckmarkBold & \CheckmarkBold & \CheckmarkBold &  &  & & & & \CheckmarkBold & \CheckmarkBold & \CheckmarkBold  &  \\
\midrule
\rowcolor{gray!10} \textbf{OmniDocBench} & 9 & \CheckmarkBold & \CheckmarkBold & \CheckmarkBold & \CheckmarkBold & \CheckmarkBold & \CheckmarkBold & \CheckmarkBold & \CheckmarkBold & \CheckmarkBold & \CheckmarkBold & \CheckmarkBold & \CheckmarkBold & \CheckmarkBold  \\

\bottomrule
\end{tabular}
}
\caption{A Comparison between OmniDocBench and existing benchmarks. \textit{BBox}: Bounding boxes. \textit{Text}: Text in Unicode. \textit{Table}: Table in LaTeX/HTML/Markdown. \textit{Formula}: Formula in LaTeX. \textit{Attributes}: Page- and BBox-Level Attributes. \textit{OCR}: Optical Character Recognition; \textit{DLA}: Document Layout Analysis; \textit{TR}: Table Recognition; \textit{MFR}: Math Formula Recognition; \textit{ROD}: Reading Order Detection}
\label{tab:related_work}
\vspace{-12pt}
\end{table*}

\subsection{Pipeline-based Document Content Extraction}

Pipeline-based methods treat the document content extraction task as a collection of single modules, such as document layout detection~\cite{huang2022layoutlmv3pretrainingdocumentai, zhao2024doclayoutyoloenhancingdocumentlayout, pramanik2020towards, gu2021unidoc}, optical character recognition~\cite{Tesseract, li2022ppocrv3attemptsimprovementultra, liu2020abcnet, huang2022swintextspotter, wang2021pgnet}, formula recognition~\cite{zhang2018multi, li2020improving, pix2tex2022, wang2024unimernetuniversalnetworkrealworld}, and table recognition~\cite{huang2012tabtransformer, huang2023improving, li2022ppocrv3attemptsimprovementultra}. 
In this sense, such methods can utilize different expert models 
to address each specific task. 
Marker~\cite{marker} integrates open-source models to parse documents into structured formats such as Markdown, JSON, and HTML. To get higher accuracy, an optional LLM-enabled version can also be integrated to merge tables across pages, handle inline math, and so on.
Similarly, MinerU~\cite{MinerU} first utilizes a layout detection model to segment the document page into different regions, then applies task-specific models for corresponding regions. Finally, it outputs the complete content in Markdown format with a reading order algorithm.
By leveraging lightweight models and parallelized operations, pipeline-based methods can achieve efficient parsing speeds.

\subsection{VLM-based Document Content Extraction}

Document understanding and optical character recognition (OCR) are crucial tasks for evaluating the perception capabilities of vision-language models (VLMs). 
By incorporating extensive OCR corpus into the pretraining stage, VLMs like GPT4o~\cite{GPT4o} and Qwen2-VL~\cite{Qwen-VL} have demonstrated comparable performance in document content extraction tasks. 
Unlike pipeline-based methods, VLMs perform document parsing in an end-to-end manner. Furthermore, without requiring specialized data fine-tuning, these models are able to deal with diverse and even unseen document types for their generalization capabilities.

To integrate the efficiency of lightweight models and the generalizability of VLMs, many works~\cite{Nougat, Vary, fox, got2, hu2024mplug2, KOSMOS-2.5} have focus on training specialized end-to-end expert models for document parsing. 
These VLM-driven models excel at comprehending both visual layouts and textual contents, balancing a trade-off between accuracy and efficiency.


\subsection{Benchmarks for Document Content
Extraction}

Document content extraction requires the ability to understand document layouts and recognize various types of content. 
However, current benchmarks fall short of a comprehensive page-level evaluation, as they focus solely on evaluating the model's performance on module-level recognition.
PubLayNet~\cite{PubLayNet} and concurrent benchmarks~\cite{DocBank, DocLayNet, M6Doc} specialize in evaluating a model's ability to detect document page layouts.
OCRBench~\cite{Liu_2024} proposes five OCR-related tasks with a greater emphasis on evaluating the model's visual understanding and reasoning capabilities. Only line-level assessments are provided for text recognition and handwritten mathematical expression recognition (HMER). 
Similarly, single-module benchmarks~\cite{RobustReading, PubTabNet, DocBank, wang2024unimernetuniversalnetworkrealworld} disentangle the task into different dimensions and focus narrowly on specific parts. Such paradigm overlooks the importance of structural and semantic information like the reading order and fails to evaluate the model's overall ability when processing the full-page documents as a whole. 

Page-level benchmarks have been proposed alongside some recent VLM-driven expert models~\cite{fox, Nougat, got2}. 
However, the robustness of these benchmarks is compromised by limitations in data size, language, document type, and annotation. For example, Nougat~\cite{Nougat} evaluates models using only printed English documents collected from arXiv while the page-level benchmark introduced by GOT-OCR~\cite{got2} consists of only 90 pages of Chinese and English documents in total. 
Commonly-seen document types like handwritten notes, newspapers, and exam papers are further neglected.
Lacking detailed annotations, the benchmarks can only conduct naive evaluation between the full-page results of Ground Truths and predictions without special handling for different output formats and specialized metrics for different content types.  
The evaluation of the model performance can be severely biased due to limited document domains, unaligned output format and mismatched metrics. \textit{Therefore, there is an urgent need for a more finely annotated, diverse, and reasonable page-level document content extraction benchmark.}

\section{OmniDocBench Dataset}

\begin{figure}[t]
    \centering
    \includegraphics[width=0.5\textwidth]{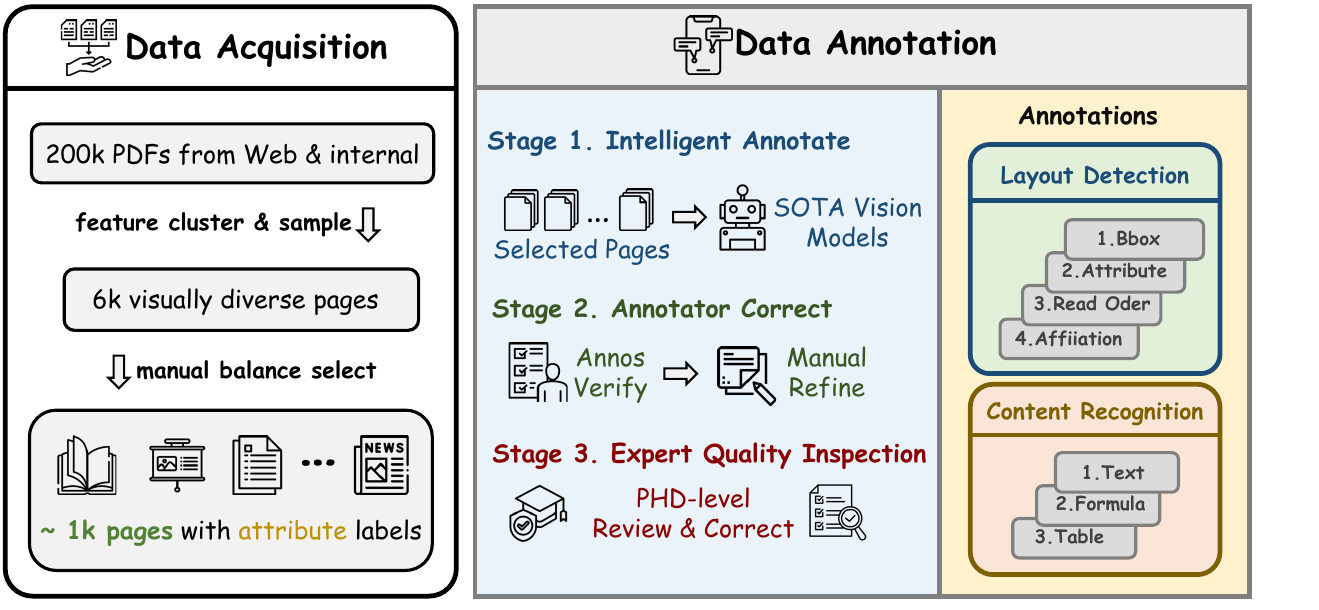}
    \caption{Overview of the OmniDocBench dataset construction.}    \label{fig:fig3_annotation}
\end{figure}

Constructing a diverse and comprehensive document parsing benchmark with precise annotations is a significant challenge. As illustrated in Figure \ref{fig:fig3_annotation}, we have designed a systematic and professional annotation framework for OmniDocBench, encompassing data acquisition, intelligent pre-annotation, and manual refinement. This ensures that OmniDocBench possesses the following key attributes:

\begin{itemize}
    \item \textbf{Page Diversity}. We sourced document pages from a variety of origins to ensure a wide range of document types.
    \item \textbf{Comprehensive Annotation}. We meticulously annotated all elements on the pages, including bounding boxes, specific contents, and various potential attributes.
    \item \textbf{Annotation Accuracy}. By integrating semi-automated annotation processes, annotator corrections, and expert quality checks, we ensure the reliability of all annotations.
\end{itemize}

The following sections detail the data acquisition process, the annotation methodology, and a statistical analysis of the final annotated dataset.

\subsection{Data Acquisition}
During the data acquisition phase, we sourced document pages from diverse origins and used clustering algorithms to initially select visually diverse pages, followed by manual annotation of page attributes to finalize the OmniDocBench pages. Specifically, we collected over 200,000 initial PDF documents from Common Crawl, Google, Baidu search engines, and internal data. Subsequently, we extracted visual features from these document pages using ResNet-50 and performed clustering using Faiss \footnote{\url{https://github.com/facebookresearch/faiss}}, sampling 6,000 visually diverse pages from 10 cluster centers. Finally, annotators provided page-level attribute annotations, including page type, layout type, and language type, and further balanced the selection to 981 samples for the final dataset. The OmniDocBench dataset includes pages from nine distinct types, multiple layout categories, and various attribute annotations, covering a wide range of real-world scenarios.

\subsection{Data Annotation}

To ensure the comprehensiveness of OmniDocBench's annotations, we conducted detailed annotations for layout detection and content recognition.

\subsubsection{Annotation Types}

\noindent\textbf{Layout Detection Annotations}: Unlike typical layout detection tasks, OmniDocBench includes four comprehensive types of annotations:
(1) Layout Bounding Box Annotations: Positioanl information for 19 distinct region categories such as titles, text paragraphs, tables, and images.
(2) Layout Attribute Annotations: Detailed attribute annotations for detected boxes, including 3 text box attribute categories, 6 table attribute categories, 9 bbox-level attribute labels in total.
(3) Reading Order Annotations: Annotating the reading sequence of detected boxes.
(4) Affiliation Annotations: For images, tables, formulas, and code blocks, we annotate captions and titles to distinguish them from main text. Similarly, for cross-page paragraphs, we annotate affiliation relationships.

\noindent\textbf{Content Recognition Annotations}: Based on the content type within each region, we conduct the following three types of annotations:
(1) Text Annotations: Pure text annotations for titles, text paragraphs, and other plain text content.
(2) Formula Annotations: LaTeX format annotations for inline formulas, display formulas, and subscripts.
(3) Table Annotations: Providing both HTML and LaTeX annotations for table data.

\subsubsection{Annotation Process}

For these annotation tasks on diverse pages, we design a standardized process to ensure quality and efficiency, comprising intelligent automatic annotation, annotator correction, and expert quality inspection.

\noindent\textbf{Automatic Annotation}. Manually annotating entire documents is time-consuming and costly. To enhance efficiency, we employ state-of-the-art detection and recognition models for pre-annotation of layout detection and content recognition. Specifically, we use fine-tuned LayoutLMv3~\cite{huang2022layoutlmv3pretrainingdocumentai} for layout detection annotations and PaddleOCR~\cite{li2022ppocrv3attemptsimprovementultra}, UniMERNet~\cite{wang2024unimernetuniversalnetworkrealworld}, and GPT-4o~\cite{GPT4o} for text, formula, and table annotations, respectively.

\noindent\textbf{Annotator Correction}. After the layout detection phase, annotators refine the detection boxes and enhance annotations with reading order and affiliation details. Each character is verified to ensure accuracy in content recognition. For complex annotations of tables and formulas, requiring LaTeX and HTML formats, annotators use tools like Tables Generator~\footnote{\url{https://www.tablesgenerator.com/}} and latexlive~\footnote{\url{https://www.latexlive.com/}} for verification and correction.

\noindent\textbf{Expert Quality Inspection}. Despite thorough annotator corrections, the complexity of formulas and tables may result in residual issues. To address these, we use CDM's rendering techniques~\cite{CDM} to identify unrenderable elements. These elements are then reviewed and corrected by three researchers to ensure accuracy in the final annotations.

\noindent\subsection{Dataset Statistics}

\noindent\textbf{Page Diversity.} OmniDocBench comprises a total of 981 PDF pages across 9 distinct types. Each page is annotated with global attributes, including text language, column layout type, and indicators for blurred scans, watermarks, and colored backgrounds.

\noindent\textbf{Annotation Diversity:} OmniDocBench contains over 100,000 annotations for page detection and recognition: (1) More than 20,000 block-level annotations across 15 categories, including over 15,979 text paragraphs, 989 image boxes, 428 table boxes, and so on. All document components except headers, footers, and page notes are labeled with reading order information, totaling over 16,000 annotations. (2) The dataset also includes more than 70,000 span-level annotations across 4 categories, with 4,009 inline formulas and 357 footnote markers represented in LaTeX format, while the remaining annotations are in text format.

\noindent\textbf{Annotation Attribute Diversity:} (1) \textit{Text Attributes:} All block-level annotations, except for tables and images, include text attribute tags. In addition to standard Chinese and English text, there are over 2,000 blocks with complex backgrounds and 493 with rotated text. (2) \textit{Table Attributes:} In addition to standard Chinese and English tables, there are 142 tables with complex backgrounds, 81 containing formulas, 150 with merged cells, and 7 vertical tables.

\section{OmniDocBench Evaluation Methodology}

\begin{figure}[t]
    \centering
    \includegraphics[width=0.48\textwidth]{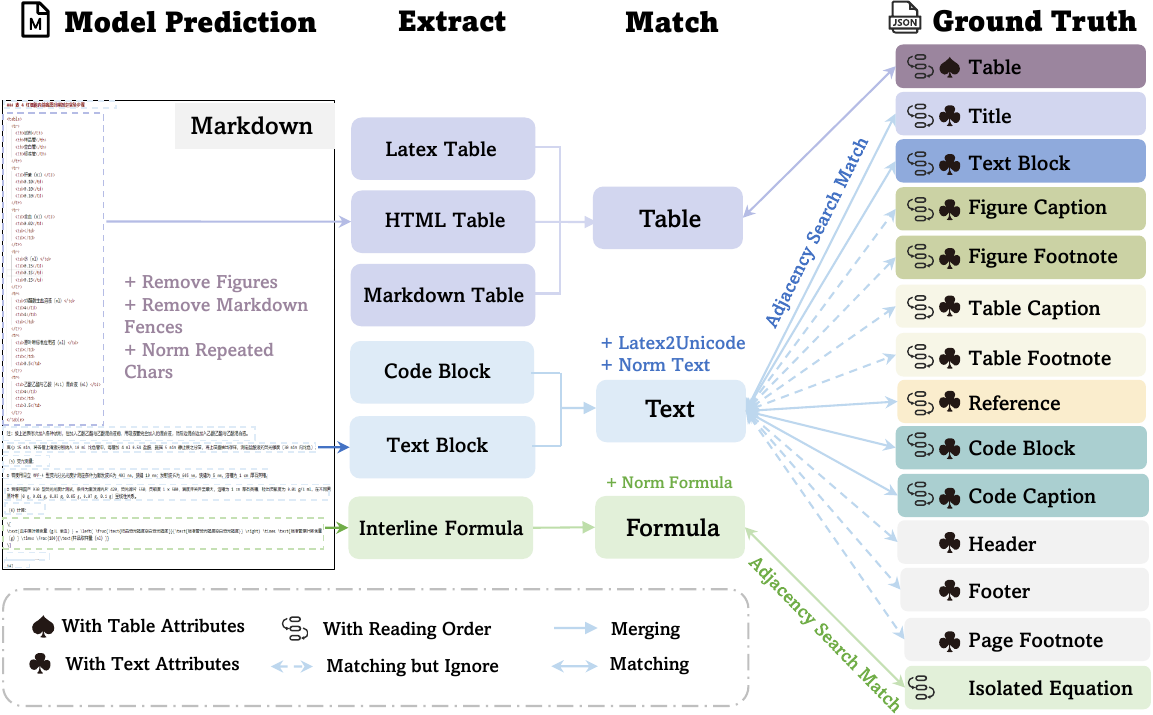}
    \caption{OmniDocBench Evaluation Pipeline. }
    \label{fig:eval}
\vspace{-5mm}
\end{figure}

To provide a fair and comprehensive evaluation for various models, we proposed an end-to-end evaluation pipeline consisting of several modules, including extraction, matching algorithm, and metric calculation, as shown in Figure~\ref{fig:eval}. It ensures that OmniDocBench automatically performs unified evaluation on document parsing, thereby producing reliable and effective evaluation results.

\subsection{Extraction}

\noindent\textbf{Preprocessing.} The model-generated markdown text should be preprocessed, which includes removing images, eliminating markdown tags at the beginning of the document, and standardizing the number of repeated characters.

\noindent\textbf{Elements Extraction.} Extraction is primarily carried out using regular expression matching. To ensure that the extraction of elements does not interfere with each other, it is necessary to follow a specific order. The extraction sequence is as follows: LaTeX tables, HTML tables, display formulas, markdown tables (which are then converted into HTML format), and code blocks.

\noindent\textbf{Pure Text Extraction.}  After extracting special components, the remaining content is considered pure text. Paragraphs are separated by double line breaks, allowing them to participate in subsequent matching processes, thus aligning with reading order annotation units in the GTs. If no double line break exists, single line breaks are used for paragraph separation. Additionally, previously extracted code blocks are merged into the text category for processing.

\noindent\textbf{Inline Formula Format Converting.}  We standardized inline formulas within paragraphs to Unicode format. This was necessary because different models produce inconsistent outputs for inline formulas. For formulas originally written in Unicode, it is hard to extract them using regular expressions. Therefore, to ensure a fair comparison, we do not extract inline formulas for separate evaluation. Instead, we include them in their Unicode format alongside the text paragraphs for evaluation.

\noindent\textbf{Reading Order Extraction.}  Upon completion of the extraction, the start and end positions of the extracted content in the original markdown are recorded for subsequent reading order calculation.

\begin{table*}[t]
  \centering
  \resizebox{0.92\textwidth}{!}{
    \begin{tabular}{c|l|cc|cc|cc|cc|cc|cc|cc}
    \toprule
    \multirow{2}{*}{\textbf{Method Type}} & \multirow{2}{*}{\textbf{Methods}} 
    & \multicolumn{2}{c|}{\textbf{{Text\textsuperscript{Edit}}}$\downarrow$} & \multicolumn{2}{c|}{\textbf{{Formula\textsuperscript{Edit}}}$\downarrow$} & \multicolumn{2}{c|}{\textbf{{Formula\textsuperscript{CDM}}}$\uparrow$} & \multicolumn{2}{c|}{\textbf{{Table\textsuperscript{TEDS}}}$\uparrow$} &  \multicolumn{2}{c|}{\textbf{{Table\textsuperscript{Edit}}}$\downarrow$} & \multicolumn{2}{c|}{\textbf{{Read Order\textsuperscript{Edit}}}$\downarrow$} & \multicolumn{2}{c}{\textbf{{Overall\textsuperscript{Edit}}}$\downarrow$} \\
    \cmidrule{3-16}
     & & {\textit{EN}} & {\textit{ZH}} & {\textit{EN}} & {\textit{ZH}} & {\textit{EN}} & {\textit{ZH}} & {\textit{EN}} & {\textit{ZH}} & {\textit{EN}} & {\textit{ZH}} & {\textit{EN}} & {\textit{ZH}} & {\textit{EN}} & {\textit{ZH}} \\
    \midrule
     \multirow{3}{*}{\textbf{Pipeline Tools}} & MinerU~\cite{MinerU} & \textbf{0.061} & \textbf{0.215} & \textbf{0.278} & 0.577 & 57.3 & 42.9 & \textbf{78.6} & 62.1 & \textbf{0.18} & 0.344 & \textbf{0.079} & 0.292 & \textbf{0.15} & \underline{0.357} \\
     & Marker~\cite{marker} & 0.08 & 0.315 & 0.53 & 0.883 & 17.6 & 11.7 & 67.6 & 49.2 & 0.619 & 0.685 & 0.114 & 0.34 & 0.336 & 0.556 \\
     & Mathpix~\tablefootnote{\label{mathpix}\url{https://mathpix.com/}} & \underline{0.105} & 0.384 & \underline{0.306} & \textbf{0.454} & 62.7 & \textbf{62.1} & \underline{77.0} & \textbf{67.1} & 0.243 & \textbf{0.32} & \underline{0.108} & 0.304 & \underline{0.191} & 0.365 \\
     \midrule
     \multirow{2}{*}{\makecell{\textbf{Expert VLMs}}} & GOT-OCR~\cite{got2} & 0.189 & 0.315 & 0.360 & \underline{0.528} & \underline{74.3} & 45.3 & 53.2 & 47.2 & 0.459 & 0.52 & 0.141 & 0.28 & 0.287 & 0.411 \\
     & Nougat~\cite{Nougat} & 0.365 & 0.998 & 0.488 & 0.941 & 15.1 & 16.8 & 39.9 & 0.0 & 0.572 & 1.000 & 0.382 & 0.954 & 0.452 & 0.973 \\
     \midrule
     \multirow{3}{*}{\makecell{\textbf{General VLMs}}} 
     & GPT4o~\cite{GPT4o} & 0.144 & 0.409 & 0.425 & 0.606 & \underline{72.8} & 42.8 & 72.0 & 62.9 & \underline{0.234} & \underline{0.329} & 0.128 & 0.251 & 0.233 & 0.399 \\
     & Qwen2-VL-72B~\cite{wang2024qwen2} & 0.096 & \underline{0.218} & 0.404 & 0.487 & \textbf{82.2} & \underline{61.2} & 76.8 & \underline{76.4} & 0.387 & 0.408 & 0.119 & \textbf{0.193} & 0.252 & \textbf{0.327} \\
     & InternVL2-76B~\cite{InternVL} & 0.353 & 0.290 & 0.543 & 0.701 & 67.4 & 44.1 & 63.0 & 60.2 & 0.547 & 0.555 & 0.317 & \underline{0.228} & 0.44 & 0.443 \\
    \bottomrule
    \end{tabular}%
  }
    \vspace{-4pt}
    \caption{Comprehensive evaluation of document parsing algorithms on OmniDocBench: performance metrics for text, formula, table, and reading order extraction, with overall scores derived from ground truth comparisons.}
  \label{result_label1}%
  \vspace{-4pt}
\end{table*}

\begin{table*}[t]
  \small
  \centering
  \resizebox{0.90\textwidth}{!}{
    \begin{tabular}{c|l|ccccccccc|c}
    \toprule

  \textbf{Model Type} & \textbf{Models} & \textbf{Book} & \textbf{Slides} & \makecell{\textbf{Financial}\\\textbf{Report}}  & \makecell{\textbf{Textbook}} & \makecell{\textbf{Exam}\\\textbf{Paper}}
  & \textbf{Magazine} & \makecell{\textbf{Academic}\\\textbf{Papers}} & \textbf{Notes}   & \textbf{Newspaper} & \textbf{Overall} \\
     \midrule
     \multirow{3}{*}{\makecell{\textbf{Pipeline Tools}}} & MinerU~\cite{MinerU} & \textbf{0.055} & 0.124 & \textbf{0.033} & \textbf{0.102} & \textbf{0.159} & \underline{0.072} & \textbf{0.025} & 0.984 & \textbf{0.171} & \underline{0.206} \\
     & Marker~\cite{marker} & \underline{0.074} & 0.34 & 0.089 & 0.319 & 0.452 & 0.153 & \underline{0.059} & 0.651 & \underline{0.192} & 0.274 \\
     & Mathpix~\textsuperscript{\ref{mathpix}} & 0.131 & 0.22 & 0.202 & 0.216 & 0.278 &0.147 & 0.091 & 0.634 & 0.69 & 0.3 \\
    \midrule  
     \multirow{2}{*}{\makecell{\textbf{Expert VLMs}}} & GOT-OCR~\cite{got2} & 0.111 & 0.222 & 0.067 & \underline{0.132} & 0.204 & 0.198 & 0.179 & 0.388 & 0.771 & 0.267 \\
     & Nougat~\cite{Nougat} & 0.734 & 0.958 & 1.000 & 0.820 & 0.930 & 0.83 & 0.214 & 0.991 & 0.871 & 0.806 \\
     \midrule
     \multirow{3}{*}{\makecell{\textbf{General VLMs}}} & GPT4o~\cite{GPT4o} & 0.157 & 0.163 & 0.348 & 0.187 & 0.281 & 0.173 & 0.146 & 0.607 & 0.751 & 0.316 \\
     & Qwen2-VL-72B~\cite{wang2024qwen2} & 0.096 & \textbf{0.061} & \underline{0.047} & 0.149 & \underline{0.195} & \textbf{0.071} & 0.085 & \textbf{0.168} & 0.676 & \textbf{0.179} \\
     & InternVL2-76B~\cite{InternVL} & 0.216 & \underline{0.098} & 0.162 & 0.184 & 0.247 & 0.150 & 0.419 & \underline{0.226} & 0.903 & 0.3 \\
        
    \bottomrule
    \end{tabular}%
  }
      \vspace{-2pt}
    \caption{End-to-end text recognition performance on OmniDocBench: evaluation using edit distance \textbf{across 9 PDF page types}.}    
    \vspace{-5pt}
  \label{result_label2}%
\end{table*}%

\begin{table}[t]
  \small
  \centering
  \renewcommand\tabcolsep{4.6pt}
  \renewcommand\arraystretch{0.9}
  \vspace{-0.1in}
  \resizebox{0.45\textwidth}{!}{
    \begin{tabular}{l|ccc|c}
    \toprule
    \textbf{Models} & \textbf{Fuzzy} & \textbf{Water} & \textbf{Color} & \textbf{None} \\
    \midrule
     MinerU~\cite{MinerU} & 0.15/0.048 & \textbf{0.151/0.031} & \underline{0.107}/\underline{0.052} & \underline{0.079}/\textbf{0.035} \\
     Marker~\cite{marker} & 0.333/0.092 & 0.484/0.126 & 0.319/0.127 & \textbf{0.062}/0.125 \\
     Mathpix~\textsuperscript{\ref{mathpix}} & 0.294/0.064 & 0.290/0.059 & 0.216/0.09 & 0.135/0.043 \\
    \midrule
     GOT-OCR~\cite{got2} & 0.175/0.05 & 0.190/0.056 & 0.186/0.097 & 0.177/0.081 \\
     Nougat~\cite{Nougat} & 0.934/0.051 & 0.915/0.071 & 0.873/0.096 & 0.615/0.208 \\
    \midrule
     GPT4o~\cite{GPT4o} & 0.263/0.078 & 0.195/0.057 & 0.184/0.078 & 0.186/0.072 \\
     Qwen2-VL-72B~\cite{wang2024qwen2} & \textbf{0.082} /\textbf{0.01} & \underline{0.172}/ 0.078 & \textbf{0.104}/\textbf{0.05} & 0.084/\underline{0.042} \\
     InternVL2-76B~\cite{InternVL} & \underline{0.120}/\underline{0.013} & 0.197/\underline{0.042} & 0.155/0.059 & 0.261/0.082 \\
    \bottomrule
    \end{tabular}%
  }
 \caption{End-to-end text recognition on OmniDocBench: evaluation \textbf{under various page attributes} using the edit distance metric. 
   The value is \textbf{Mean/Variance} of scores in the attribute group. Columns represent: \textit{Fuzzy} (Fuzzy scan), \textit{Water} (Watermark), \textit{Color} (Colorful background). \textit{None} (No special issue)}  \label{tab:special_situation2}%
\end{table}%

\begin{table}[t]
  \small
  \centering
  \renewcommand\tabcolsep{4.0pt}
  \renewcommand\arraystretch{0.9}
  \vspace{-0.1in}
  \resizebox{0.45\textwidth}{!}{
    \begin{tabular}{l|cccc}
    \toprule
    \textbf{Models} & \textbf{Single}  & \textbf{Double}  & \textbf{Three}   & \textbf{Complex} \\
    \midrule
     MinerU~\cite{MinerU} & 0.311/0.187 & \textbf{0.101}/\textbf{0.013} & \textbf{0.117}/\underline{0.046} & \underline{0.385}/\textbf{0.057} \\
     Marker~\cite{marker} & 0.299/0.143 & 0.299/0.299 & \underline{0.149}/0.063 & \textbf{0.363}/\underline{0.086} \\
     Mathpix~\textsuperscript{\ref{mathpix}} & 0.207/0.123 & 0.188/0.07 & 0.225/\textbf{0.029} & 0.452/0.177 \\
    \midrule
     GOT-OCR~\cite{got2} & 0.163/0.106 & \underline{0.145}/0.059 & 0.257/0.072 & 0.468/0.185 \\
     Nougat~\cite{Nougat} & 0.852/0.084 & 0.601/0.224 & 0.662/0.093 & 0.873/0.09  \\
    \midrule
     GPT4o~\cite{GPT4o} & 0.109/0.112 & 0.204/0.076 & 0.254/\underline{0.046} & 0.426/0.188 \\
     Qwen2-VL-72B~\cite{wang2024qwen2} & \textbf{0.066}/\textbf{0.048} & \underline{0.145}/\underline{0.049} & 0.204/0.055 & 0.394/0.203 \\
     InternVL2-76B~\cite{InternVL} & \underline{0.082}/\underline{0.052} & 0.312/0.069 & 0.682/0.098 & 0.444/0.174 \\
    \bottomrule
    \end{tabular}%
  }
  \caption{End-to-end reading order evaluation on OmniDocBench: results \textbf{across different column layout types} using Normalized Edit Distance. The value is \textbf{Mean/Variance} of scores in the attribute group.}
  \label{result_label3}%
\end{table}%

\subsection{Matching Algorithm}

\noindent\textbf{Adjacency Search Match.} To avoid the impact of paragraph splitting on the final results, we proposed Adjacency Search Match, that merges and splits paragraphs in both GTs and Preds to achieve the best possible match. The specific strategy involves: i) Calculate a metrix of Normalized Edit Distance between GTs and Preds. The Pred and GT pairs whose similarity exceeds a specific threshold are considered as successful match. ii) For the rest, we apply fuzzy matching to determine whether one string is a subset of another string. If so, we further apply the merging algorithm which would try to merge adjacent paragraph. This process would continue to merge more paragraph until the Normalized Edit Distance starts to decrease. After this process, the best match will be found for GTs and Preds.

\noindent\textbf{Ignore Handling.} We implement an ignore logic for certain components in PDF page content, meaning they participate in matching but are excluded from metric calculations. This is mainly because of inconsistent output standards among models, which should not affect the validation results. For fairness, we ignore: (1) Headers, footers, page numbers, and page footnotes, which are handled inconsistently by different models. (2) Captions for figures, tables, and footnotes often have uncertain placements, thus complicating the reading order. Additionally, some models embed table captions in HTML or LaTeX tables, while others treat them as plain text.

\begin{table*}[ht]
\centering
\resizebox{0.95\textwidth}{!}{
    \begin{tabular}{l|cc|ccccccccc|c}
    \toprule
    \textbf{Model} & \textbf{Backbone} & \textbf{Params} & \textbf{Book} & \textbf{Slides} & \makecell{\textbf{Research}\\\textbf{Report}} & \makecell{\textbf{Textbook}} & \makecell{\textbf{Exam}\\\textbf{Paper}} & \textbf{Magazine} & \makecell{\textbf{Academic}\\\textbf{Literature}} & \textbf{Notes} & \textbf{Newspaper} & \textbf{Average} \\
    \midrule
    DiT-L~\cite{DBLP:conf/mm/LiXL0ZW22}  & ViT-L & 361.6M & \underline{43.44} & 13.72  & 45.85   & 15.45   & 3.40   & 29.23  & \textbf{66.13}  & 0.21  & 23.65 & 26.90 \\
    LayoutLMv3~\cite{huang2022layoutlmv3pretrainingdocumentai} & RoBERTa-B & 138.4M & 42.12   & 13.63    & 43.22     & 21.00  & 5.48   & 31.81  & \underline{64.66}   & 0.80   & 30.84 & 28.84 \\
    DocLayout-YOLO~\cite{zhao2024doclayoutyoloenhancingdocumentlayout} & v10m & 19.6M & \textbf{43.71}   & \textbf{48.71}  & \textbf{72.83}   & \textbf{42.67} & \textbf{35.40} & \underline{51.44}   & \underline{64.64}  & \underline{9.54} & \textbf{57.54} & \textbf{47.38} \\
    SwinDocSegmenter~\cite{DBLP:conf/icdar/BanerjeeBLP23} & Swin-L & 223M & 42.91 &	\underline{28.20} &	\underline{47.29} &	\underline{32.44} &	\underline{20.81} &	\textbf{52.35} &	48.54 &	\textbf{12.38} &	\underline{38.06} &	\underline{35.89} \\
    GraphKD~\cite{DBLP:conf/icdar/BanerjeeBLP24} & R101 & 44.5M  & 39.03 & 16.18	& 39.92 & 22.82 &	14.31 &	37.61 &	44.43 &	5.71 &	23.86 &	27.10 \\
    DOCX-Chain~\cite{DocXChain2023} & - & - & 30.86   & 11.71  & 39.62   & 19.23   & 10.67  & 23.00  & 41.60   & 1.80 & 16.96 & 21.27 \\
    \bottomrule                  
    \end{tabular}
}
  \caption{Component-level layout detection evaluation on OmniDocBench layout subset: mAP results by PDF page type.}
 \label{tab:detection}
\end{table*}%


\begin{table*}[ht]
  \centering
  \renewcommand\tabcolsep{5.6pt}
  \renewcommand\arraystretch{0.98}
  \resizebox{0.95\textwidth}{!}{
    \begin{tabular}{c|l|ccc|cccc|cccc|c}
    \toprule
\multirow{2}{*}{\textbf{Model Type}} & \multirow{2}{*}{\textbf{Model}} & \multicolumn{3}{c|}{\textbf{Language}} & \multicolumn{4}{c|}{\textbf{Table Frame Type}} & \multicolumn{4}{c|}{\textbf{Special Situation}} &  \multirow{2}{*}{\textbf{Overall}}  \\
  &  & \textit{EN} & \textit{ZH} & \textit{Mixed}   & \textit{Full} & \textit{Omission} 
& \textit{Three} & \textit{Zero} & \textit{Merge Cell}(+/-) & \textit{Formula}(+/-) & \textit{\makecell{Colorful}}(+/-) & \textit{Rotate}(+/-) \\
  \midrule
   \multirow{2}{*}{\makecell{\textbf{OCR-based} \\ \textbf{ Models}}} 
   & PaddleOCR\cite{li2022ppocrv3attemptsimprovementultra} & \underline{76.8} & 71.8 & 80.1 & 67.9 & 74.3 & \underline{81.1} & 74.5 & \underline{70.6}/75.2 & \underline{71.3}/74.1 & 72.7/74.0 & 23.3/74.6 & 73.6 \\
   & RapidTable\cite{RapidTable2023} & \textbf{80.0} & \textbf{83.2} & \textbf{91.2} & \textbf{83.0} & \textbf{79.7} & \textbf{83.4} & 78.4 & \textbf{77.1/85.4} & \textbf{76.7/83.9} & \textbf{77.6/84.9} & \underline{25.2}/\textbf{83.7} & \textbf{82.5} \\
   \midrule
   \multirow{2}{*}{\textbf{Expert VLMs}} & StructEqTable\cite{StructEqTable2024} & 72.8 & \underline{75.9} & 83.4 & 72.9 & \underline{76.2} & 76.9 & \textbf{88.0} & 64.5/\underline{81.0} & 69.2/76.6 & \underline{72.8}/76.4 & \textbf{30.5}/76.2 & \underline{75.8} \\
   & GOT-OCR~\cite{got2} & 72.2 & 75.5 & \underline{85.4} & \underline{73.1} & 72.7 & 78.2 & 75.7 & 65.0/80.2 & 64.3/\underline{77.3} & 70.8/\underline{76.9} & 8.5/\underline{76.3} & 74.9 \\
   \midrule
   \multirow{2}{*}{\textbf{General VLMs}} & Qwen2-VL-7B~\cite{wang2024qwen2} & 70.2 & 70.7 & 82.4 & 70.2 & 62.8 & 74.5 & \underline{80.3} & 60.8/76.5 & 63.8/72.6 & 71.4/70.8 & 20.0/72.1 & 71.0 \\
   & InternVL2-8B~\cite{InternVL} & 70.9 & 71.5 & 77.4 & 69.5 & 69.2 & 74.8 & 75.8 & 58.7/78.4 & 62.4/73.6 & 68.2/73.1 & 20.4/72.6 & 71.5 \\
  \bottomrule
    \end{tabular}%
  }
  \caption{Component-level Table Recognition evaluation on OmniDocBench table subset. \textit{(+/-)} means \textit{with/without} special situation.}
  \label{tab:table_single_attr}%
\end{table*}%

\begin{table*}[t]
  \small
  \centering
  \renewcommand\tabcolsep{5.6pt}
  \renewcommand\arraystretch{0.92}
  \vspace{-0.1in}
  \resizebox{0.90\textwidth}{!}{
    \begin{tabular}{c|l|ccc|ccc|cccc}
    \toprule
    \multirow{2}{*}{\textbf{Model Type}} & \multirow{2}{*}{\textbf{Model}} & \multicolumn{3}{c|}{\textbf{Language}} & \multicolumn{3}{c|}{\textbf{Text background}} & \multicolumn{4}{c}{\textbf{Text Rotate}}  \\
    &  & \textit{EN} & \textit{ZH} & \textit{Mixed}   & \textit{White} & \textit{Single} 
  & \textit{Multi} & \textit{Normal} & \textit{Rotate90} & \textit{Rotate270} & \textit{Horizontal} \\
    \midrule
     \multirow{5}{*}{\makecell{\textbf{Expert Vision}\\\textbf{Models}}} 
     & PaddleOCR~\cite{li2022ppocrv3attemptsimprovementultra}    & 0.071 & \textbf{0.055} & \textbf{0.118} & \textbf{0.060} & \textbf{0.038} & \textbf{0.085} & \textbf{0.060} & \textbf{0.015} & \underline{0.285} & \textbf{0.021} \\
     & Tesseract OCR~\tablefootnote{\label{tesseract}\url{https://github.com/tesseract-ocr/tesseract}}    & 0.179 & 0.553 & 0.553 & 0.453 & 0.463 & 0.394 & 0.448 & 0.369 & 0.979 & 0.982 \\
     & Surya~\tablefootnote{\label{surya}\url{https://github.com/VikParuchuri/surya}}      & 0.057 & 0.123 & 0.164 & 0.093 & 0.186 & 0.235 & 0.104 & 0.634 & 0.767 & 0.255 \\
     & GOT-OCR~\cite{got2}     & 0.041 & \underline{0.112} & 0.135 & \underline{0.092} & \underline{0.052} & 0.155 & \underline{0.091} & 0.562 & 0.966 & 0.097 \\
     & Mathpix~\textsuperscript{\ref{mathpix}}   & \underline{0.033} & 0.240 & 0.261 & 0.185 & 0.121 & 0.166 & 0.180 & \underline{0.038} & \textbf{0.185} & 0.638 \\
     \midrule
     \multirow{3}{*}{\makecell{\textbf{Vision Language}\\\textbf{Models}}} 
     & Qwen2-VL-72B~\cite{wang2024qwen2} & 0.072 & 0.274 & 0.286 & 0.234 & 0.155 & \underline{0.148} & 0.223 & 0.273 & 0.721 & \underline{0.067} \\
     & InternVL2-76B~\cite{InternVL}     & 0.074 & 0.155 & 0.242 & 0.113 & 0.352 & 0.269 & 0.132 & 0.610 & 0.907 & 0.595 \\
     & GPT4o~\cite{GPT4o}     & \textbf{0.020} & 0.224 & \underline{0.125} & 0.167 & 0.140 & 0.220 & 0.168 & 0.115 & 0.718 & 0.132 \\
    \bottomrule
    \end{tabular}%
  }
    
    \caption{Component-level evaluation on OmniDocBench OCR subset: results grouped by text attributes using the edit distance metric.}
  \label{tab:OCR}%
  \vspace{-4pt}
\end{table*}%

\begin{table}[t]
  \centering
  \resizebox{0.47\textwidth}{!}{
    \begin{tabular}{l|cccc}
    \toprule
    \textbf{Models} & \textbf{CDM} & \textbf{ExpRate@CDM} & \textbf{BLEU}  & \textbf{Norm Edit} \\
    \midrule
     GOT-OCR~\cite{got2} & 74.1 & 28.0 & 55.07 & 0.290 \\
     Mathpix~\textsuperscript{\ref{mathpix}} & \underline{86.6} & 2.8 & \textbf{66.56} & 0.322 \\
     Pix2Tex~\tablefootnote{\label{Pix2Tex}\url{https://github.com/lukas-blecher/LaTeX-OCR}} & 73.9 & 39.5 & 46.00 & 0.337 \\
     UniMERNet-B~\cite{wang2024unimernetuniversalnetworkrealworld} & 85.0 & \underline{60.2} & \underline{60.84} & \textbf{0.238} \\
     \midrule
     GPT4o~\cite{GPT4o} & \textbf{86.8} & \textbf{65.5} & 45.17 & \underline{0.282} \\
     InternVL2-76B~\cite{InternVL} & 67.4 & 54.5 & 47.63 & 0.308 \\
     Qwen2-VL-72B~\cite{wang2024qwen2} & 83.8 & 55.4 & 53.71 & 0.285 \\
     
    \bottomrule
    \end{tabular}%
    }
  \caption{Component-level formula recognition evaluation on OmniDocBench formula subset.}
  \vspace{-0.5cm}
  \label{tab:FormulaRecog}%
\end{table}%

\subsection{Metric Calculation}

\noindent\textbf{Pure Text. } We calculate Normalized Edit Distance~\cite{levenshtein1966binary}, averaging these metrics at the sample level to obtain the final scores.

\noindent\textbf{Tables.} All tables are converted to HTML format before calculating the Tree-Edit-Distance-based Similarity (TEDS)~\cite{PubTabNet} metric and Normalized Edit Distance.

\noindent\textbf{Formulas.} Formulas are currently evaluated using the Character Detection Matching (CDM) metric~\cite{CDM}, Normalized Edit Distance, and BLEU~\cite{papineni2002bleu}.

\noindent\textbf{Reading Order.} Reading order is evaluated using the Normalized Edit Distance as metric. It only involves text components, with tables, images, and ignored components excluded from the final reading order calculation. 


\section{Benchmarks}

Based on the distinct characteristics of these algorithms, we categorize document content extraction methods into three main classes:
\begin{itemize}
    \item \textbf{Pipeline Tools:} These methods integrate layout detection and various content recognition tasks (such as OCR, table recognition, and formula recognition) into a document parsing pipeline for content extraction. Prominent examples include MinerU~\cite{MinerU} (v0.9.3), Marker~\cite{marker} (v1.2.3), and Mathpix\textsuperscript{\ref{mathpix}}.
    \item \textbf{Expert VLMs:} These are large multimodal models specifically trained for document parsing tasks. Representative models include GOT-OCR2.0~\cite{got2} and Nougat~\cite{Nougat}.
    \item \textbf{General VLMs:} These are general-purpose large multimodal models inherently capable of document parsing. Leading models in this category include GPT-4o~\cite{GPT4o}, Qwen2-VL-72B~\cite{wang2024qwen2}, and InternVL2-76B~\cite{InternVL}.
\end{itemize}

\subsection{End-to-End Evaluation Results}

\noindent\textbf{Overall Evaluation Results}. As illustrated in Table~\ref{result_label1}, pipeline tools such as MinerU and Mathpix,  demonstrate superior performance across sub-tasks like text recognition, formula recognition, and table recognition. Moreover, the general Vision Language Models (VLMs), Qwen2-VL, and GPT4o, also exhibit competitive performance. Almost all algorithms score higher on English than on Chinese pages. 

\noindent\textbf{Performance Across Diverse Page Types}. To gain deeper insights into model performance on diverse document types, we evaluated text recognition tasks across different page types. Intriguingly, as shown in Table~\ref{result_label2}, pipeline tools perform well for commonly used data, such as academic papers and financial reports. Meanwhile, for more specialized data, such as slides and handwritten notes, general VLMs demonstrate stronger generalization. Notably, most VLMs fail to recognize when dealing with the Newspapers, while pipeline tools achieve significantly better performance.

\noindent\textbf{Performance on Pages with Visual Degradations}. 
In Table~\ref{tab:special_situation2}, we further analyze performance on pages containing common document-specific challenges, including fuzzy scans, watermarks, and colorful backgrounds. 
VLMs like InternVL2 and Qwen2-VL exhibit higher robustness in these scenarios despite visual noise. Among pipeline tools, MinerU remains competitive due to its strong layout segmentation and preprocessing capabilities.

\noindent\textbf{Performance on Different Layout Types}. 
Page layout is a critical factor in document understanding, especially for tasks involving reading order. OmniDocBench annotates layout attributes such as single-column, multi-column, and complex custom formats. Across all models, we observe a clear drop in accuracy on multi-column and complex layouts. MinerU shows the most consistent reading order prediction, though its performance dips on handwritten single-column pages due to recognition noise.

\noindent\textbf{Discussion on End-to-End Results.} 
1) While general VLMs often lag behind specialized pipelines and expert models on standard documents (e.g., academic papers), they generalize better to unconventional formats (e.g., notes) and perform more robustly under degraded conditions (e.g., fuzzy scans). This is largely due to their broader training data, enabling better handling of long-tail scenarios compared to models trained on narrow domains.
2) VLMs, however, struggle with high-density documents like newspapers due to limitations in input resolution and token length. In contrast, pipeline tools leverage layout-based segmentation to process components individually, maintaining accuracy in complex layouts.
Enhancing VLMs with layout-aware designs and domain-specific fine-tuning offers a promising path forward. OmniDocBench facilitates this by providing detailed annotations for layout, text, formulas, and tables, enabling comprehensive benchmarking and modular tool development for diverse document parsing tasks.

\subsection{Single Task Evaluation Results}

\noindent\textbf{Layout Detection Results}. Layout detection is the first step in document parsing using pipeline tools. A robust layout detection algorithm should perform well across a variety of document types. Table~\ref{tab:detection} presents an evaluation of leading layout detection models. The DocLayout-YOLO method, which is pre-trained on diverse synthetic document data, significantly outperforms other approaches. This superiority is a key factor in MinerU's integration of DocLayout-YOLO, contributing to its outstanding overall performance. Other methods perform well on books and academic literature but struggle with more diverse formats due to limited training data.

\noindent\textbf{Table Recognition Results}. 
In Table~\ref{tab:table_single_attr}, We evaluate table recognition models across three dimensions on our OmniDocBench table subset: language diversity, table frame types, and special situations. Among all models, OCR-based models demonstrate superior overall performance, with RapidTable achieving the highest scores in language diversity and maintaining stable performance across different frame types. Expert VLMs show competitive results in specific scenarios, with StructEqTable~\cite{StructEqTable2024} excelling in no-frame tables and showing better rotation robustness. General VLMs (Qwen2-VL-7B and InternVL2-8B) exhibit relatively lower but consistent performance, suggesting that while general-purpose VLMs have made progress in table understanding, they still lag behind specialized solutions.

\noindent\textbf{Text Recognition Results}. 
Table~\ref{tab:OCR} compares OCR tools across languages, backgrounds, and rotations using Edit Distance. PaddleOCR outperforms all competitors, followed by GOT-OCR and Mathpix. General VLMs struggle to handle text rotation or mixed-language scenarios.

\noindent\textbf{Formula Recognition Results}. 
Table~\ref{tab:FormulaRecog} presents results on formula parsing, using CDM, BLEU, and normalized Edit Distance. 
GPT-4o, Mathpix, and UniMERNet achieve results of 86.8\%, 86.6\%, and 85.0\%, respectively. Notably, GPT-4o excels with a recall rate of 65.5\% under strict conditions requiring perfect character accuracy. Although Mathpix shows high character-level precision, it occasionally omits punctuation, such as commas, leading to a lower overall correctness rate. Nonetheless, all three models are strong candidates for formula recognition tasks.

\section{Conclusion}

This paper addresses the lack of diverse and realistic benchmarks in document parsing research by introducing OmniDocBench, a dataset featuring a variety of page types with comprehensive annotations, along with a flexible and reliable evaluation framework. OmniDocBench enables systematic and fair assessments of document parsing methods, providing crucial insights for advancing the field. Its task-specific and attribute-level evaluations facilitate targeted model optimization, promoting more robust and effective parsing solutions.


\clearpage

{
    \small
    \bibliographystyle{ieeenat_fullname}
    \bibliography{main}
}


\clearpage
\setcounter{page}{1}
\maketitlesupplementary
\setcounter{section}{0}
\setcounter{table}{0}
\setcounter{figure}{0}
\renewcommand{\thesection}{\Roman{section}}
\renewcommand{\thetable}{S\arabic{table}}
\renewcommand{\thefigure}{S\arabic{figure}}

\section{More End-to-End Evaluation Results}

\Cref{tab:end2end_table} presents the evaluation results of End2End Tables grouped by Table Attributes. As it shows, most of the models perform better in English Tables rather than Chinese ones. Most models perform relatively poorly with Full Frame and No Frame tables. The accuracy of most models is affected by special conditions. Merged cells and formulas mainly test the breadth of data the model can recognize, while colored backgrounds and table rotation test their robustness. The results show that table rotation significantly impacts the accuracy of all models. Pipeline Tools' performance would not be affected by more challenging tables (e.g., merge cell), but colored backgrounds can affect recognition accuracy. Several Vision Language Models (VLMs) tend to perform worse on tables with merged cells, but colored backgrounds do not significantly impact table recognition accuracy.

\Cref{tab:end2end_text} shows the evaluation results of End2End Text blocks grouped by Text Attributes. Almost all models have lower recognition accuracy in Chinese compared to English. Some models, such as MinerU and Marker, experience a further decrease in accuracy when recognizing mixed Chinese and English content. The main reason is that minerU's text recognition module is PaddleOCR model. According to the performance of the PaddleOCR model in text recognition module, its accuracy will decline in the case of mixed language. Moreover, complex background colors significantly affect the recognition accuracy of pipeline tools, but it has only little impact on accuracy for VLMs.

\begin{table*}[t]
  \small
  \centering
  \renewcommand\tabcolsep{5.6pt}
  \renewcommand\arraystretch{0.9}
  \vspace{-0.1in}
  \resizebox{1.0\textwidth}{!}{
    \begin{tabular}{c|l|ccc|cccc|cccc}
    \toprule
    \multirow{2}{*}{\textbf{Model Type}} & \multirow{2}{*}{\textbf{Model}} & \multicolumn{3}{c|}{\textbf{Language}} & \multicolumn{4}{c|}{\textbf{Table Frame Type}} & \multicolumn{4}{c}{\textbf{Special Situation}}  \\
    &  & \textit{EN} & \textit{ZH} & \textit{Mixed}   & \textit{Full} & \textit{Omission} 
  & \textit{Three} & \textit{Zero} & \textit{Merge Cell}(+/-) & \textit{Formula}(+/-) & \textit{\makecell{Colorful}}(+/-) & \textit{Rotate}(+/-) \\
    \midrule
     \multirow{3}{*}{\textbf{Pipeline Tools}} & MinerU & \underline{75.1} & 59.3 & \textbf{79.1} & 59.4 & 71.6 & \underline{69.7} & 60.0 & 63.6/65.3 & \underline{66.0}/64.4 & 59.2/67.5 & 3.0/65.8 \\
     & Marker & 64.9 & 47.3 & 49.8 & 44.5 & 61.8 & 59.0 & \textbf{63.6} & 52.6/52.7 & 53.2/52.5 & 48.0/54.9 & \underline{35.5}/52.9 \\
     & Mathpix & \textbf{75.4} & \textbf{63.2} & 71.3 & \underline{67.4} & \underline{77.3} & 66.3 & 25.5 & \textbf{70.3}/\underline{65.4} & \underline{68.7}/\underline{66.7} & 59.7/\underline{70.8} & 19.2/\underline{67.9} \\
    \midrule
     \multirow{2}{*}{\makecell{\textbf{Expert Vision}\\\textbf{Models}}} & GOT-OCR & 51.7 & 46.2 & 49.0 & 45.5 & 48.3 & 51.3 & 46.2 & 46.0/48.9 & 45.7/48.4 & 39.8/51.9 & 0.0/48.7 \\
     & Nougat & 36.2 & 0.3 & 0.0 & 6.1 & 3.5 & 22.1 & 0.0 & 15.0/8.9 & 21/8.7 & 2.6/15.2 & 0.0/11.2 \\
     \midrule
     \multirow{3}{*}{\makecell{\textbf{Vision Language}\\\textbf{Models}}} & GPT4o & 71.1 & 58.0 & 57.3 & 62.5 & 68.7 & 61.3 & 31.2 & 56.8/64.7 & 60.8/62.2 & \underline{61.4}/62.2 & 14.2/62.7 \\
     & Qwen2-VL-72B & 73.2 & \textbf{75.1} & \underline{76.1} & \textbf{72.0} & \textbf{79.0} & \textbf{77.5} & \underline{63.2} & \underline{67.9}/\textbf{78.1} & \textbf{71.6}/\textbf{75.3} & \textbf{77.9}/\textbf{72.9} & \textbf{42.7}/\textbf{75.1} \\
     & InterVL2-76B & 60.9 & 58.5 & 65.4 & 58.8 & 65.3 & 58.3 & 55.6 & 49.0/65.1 & 53.3/60.9 & 58.8/59.8 & 6.9/60.3 \\ 
    \bottomrule
    \end{tabular}%
  }
     \caption{End-to-End Table TEDS Result grouped by Table Attributes
  }
  \label{tab:end2end_table}%
\end{table*}%

\begin{table}[t]
  \centering
  \resizebox{0.47\textwidth}{!}{
    \begin{tabular}{c|l|ccc|ccc}
    \toprule
    \multirow{2}{*}{\textbf{Model Type}} & \multirow{2}{*}{\textbf{Model}} & \multicolumn{3}{c|}{\textbf{Language}
} & \multicolumn{3}{c}{\textbf{Text background}	} \\
    & & \textit{EN} & \textit{ZH} & \textit{Mixed}  & \textit{White} & \textit{Single} & \textit{Multi}
    \\
    \midrule
     \multirow{3}{*}{\textbf{Pipeline Tools}} & MinerU & \textbf{0.124} & \textbf{0.234} & 0.742 & \textbf{0.188} & \textbf{0.15} & 0.514
  \\
     & Marker & 0.163 & \underline{0.379} & 0.747 & \underline{0.303} & 0.396 & 0.594 \\
     & Mathpix & 0.175 & 0.793 & 0.538 & 0.698 & 0.587 & 0.583 \\
    \midrule
     \multirow{2}{*}{\makecell{\textbf{Expert Vision}\\\textbf{Models}}} & GOT-OCR & 0.251 & 0.763 & 0.266 & 0.669 & 0.595 & 0.440 \\
     & Nougat & 0.587 & 0.991 & 0.983 & 0.874 & 0.935 & 0.972 \\
     \midrule
     \multirow{3}{*}{\makecell{\textbf{Vision Language}\\\textbf{Models}}} & GPT4o & 0.170 & 0.647 & 0.322 & 0.536 & 0.423 & 0.406 \\
     & Qwen2-VL-72B & \underline{0.128} & 0.582 & \textbf{0.209} & 0.494 & 0.388 & \textbf{0.217} \\
     & InternVL2-76B & 0.418 & 0.606 & \underline{0.251} & 0.589 & \underline{0.366} & \underline{0.221} \\
        
    \bottomrule
    \end{tabular}%
  }
     \caption{End-to-End Text Normalized Edit Distance results grouped by Text Attributes. “Mixed” represents a mixture of Chinese and English, “Single” and “Multi” represent single color and multi color.
  }
  \label{tab:end2end_text}%
\end{table}%
\begin{table}[t]
    \centering
      \renewcommand\tabcolsep{5.6pt}
  \renewcommand\arraystretch{0.9}
    \resizebox{0.35\textwidth}{!}{
    \begin{tabular}{llr}
        \hline
        \textbf{Category} & \textbf{Attribute Name} & \textbf{Count} \\ \hline
        \textbf{PDF Type} & Book & 104 \\
         & PPT2PDF & 133 \\
         & Research Report & 81 \\ 
         & Colorful Textbook & 96 \\
         & Exam Paper & 114 \\ 
         & Magazine & 97 \\
         & Academic Literature & 129 \\ 
         & Notes & 116 \\
         & Newspaper & 111 \\
         \hline
        \textbf{Layout Type} & Single Column & 477 \\ 
         & Double Column & 126 \\ 
         & Three Column & 45 \\ 
         & One\&More Mixed & 120 \\ 
         & Complex Layout & 213 \\
         \hline
        \textbf{Language} & English & 290 \\ 
         & Simplified Chinese & 612 \\ 
         & Mixed & 79 \\ \hline
        \textbf{Special Issues} & Fuzzy Scan & 28 \\ 
         & Watermark & 65 \\ 
         & Colorful Background & 246 \\
         \bottomrule
    \end{tabular}
    }
    \caption{The Page Attributes Statistics of OmniDocBench.}
    \label{tab:page_attribute_count}
\end{table}
\begin{table}[t]
    \centering
      \renewcommand\tabcolsep{5.6pt}
  \renewcommand\arraystretch{0.9}
    \resizebox{0.35\textwidth}{!}{

    \begin{tabular}{lll}
        \toprule
        \textbf{Attribute Category} & \textbf{Category Name} & \textbf{Count} \\ 
        \midrule
        \textbf{Language} & English & 5857 \\ 
         & Simplified Chinese & 16073 \\ 
         & EN\&CH Mixed & 1080 \\ 
        \cmidrule(lr){1-3}
        \textbf{Text Background} & White & 19465 \\ 
         & Single-Colored & 1116 \\ 
         & Multi-Colored & 2429 \\ 
        \cmidrule(lr){1-3}
        \textbf{Text Rotate} & Normal & 22865 \\ 
         & Rotate90 & 14 \\ 
         & Rotate270 & 58 \\ 
         & Horizontal & 421 \\ 
        \bottomrule
    \end{tabular}}
    \caption{Text Attributes Statistics of OmniDocBench.}
    \label{tab:text_attribute_count}
\end{table}
\begin{table}[t]
    \centering
      \renewcommand\tabcolsep{5.6pt}
  \renewcommand\arraystretch{0.9}
      \resizebox{0.35\textwidth}{!}{
    \begin{tabular}{lll}
        \toprule
        \textbf{Attribute Category} & \textbf{Category Name} & \textbf{Count} \\ 
        \midrule
        \textbf{Language} & English & 128 \\ 
         & Simplified Chinese & 285 \\ 
         & EN\&CH Mixed & 15 \\ 
        \cmidrule(lr){1-3}
        \textbf{Table Frame Type} & Full Frame & 205 \\ 
         & Omission Line & 62 \\ 
         & Three Line & 147 \\ 
         & No Frame & 14 \\ 
        \cmidrule(lr){1-3}
        \textbf{Special Issues} & Merge Cell & 150 \\ 
         & Colorful Background & 142 \\ 
         & Contain Formula & 81 \\ 
         & Rotate & 7 \\ 
        \bottomrule
    \end{tabular}}
    \caption{Table Attributes Statistics of OmniDocBench.}
    \label{tab:table_attribute_count}
\end{table}

\begin{table*}[h!]
\centering
\resizebox{1\textwidth}{!}{
\begin{tabular}{ l l l c c }
\hline
No. & Category Name & Explaination & Total \\
\hline
1 & Title & Include main titles, chapter titles, etc. & 2972 \\
2 & Text Block & Text paragraphs, which are usually separated by double line breaks in Markdown. & 15979 \\
3 & Figure & Including images, visual charts, etc. & 989 \\
4 & Figure Caption  & Typically starts with 'Figure' followed by a number, or just descriptive language below the figure. & 651 \\
5 & Figure Footnotes & Descriptive language, apart from the figure caption, usually starts with an asterisk (*). & 133 \\
6 & Table & Content organized in table form usually includes borders or a clear table structure. & 428 \\
7 & Table Caption & Typically starts with 'Table' followed by a number, or just descriptive language above the Table. & 299 \\
8 & Table Footnotes & Descriptive language, apart from the table caption, usually starts with an asterisk (*). & 132 \\
9 & Header & Information located at the top of a PDF page or in the sidebar, separate from the main content, typically includes chapter names and other details. & 1271 \\
10 & Footer & Information located at the bottom of a PDF page, separate from the main content, typically includes the publisher's name and other details. & 541 \\
11 & Page Number & It is usually represented by numbers, which may be located at the top, in the sidebar, or at the bottom of the page. & 669 \\
12 & Page Footnote & It provides further explanation of the footnotes marked within the page content. For example, information about the authors' affiliations. & 92 \\
13 & Code Block & In Markdown, a code block is typically defined using triple backticks (```). & 13 \\
14 & Code Block Caption & Descriptive language above the Code Block. & / \\
15 & Reference & Typically found only in academic literature. & 260 \\
\hline
16 & Text Span & Span-Level text box, which is the plain text content can be directly written in Markdown format. & 73143 \\
17 & Equation Inline & Formulas that need to be represented using LaTeX format and embedded within the text. & 4009 \\
18 & Equation Ignore & Some formulas that can be displayed correctly without using LaTeX formatting, such as \textit{15 kg}. & 3685 \\
19 & Footnote Mark & Typically embedded within the text as superscripts or subscripts, and their numbering usually corresponds to page footnotes. & 357 \\
\hline
20 & Other Abandoned Categories & (Masked) Some uncategorizable, irrelevant page information, such as small icons, etc. & 538 \\
21 & Masked Text Block & (Masked) Some difficult-to-recognize information that disrupts text flow, such as pinyin annotations above Chinese characters. & 34 \\
22 & Organic Chemical Formula & (Masked) Organic chemistry formulas, which are difficult to write using Markdown and are easily recognized as Figures. & 24 \\
\hline
\end{tabular}
}
\caption{Annotation Explanations and Statistics.}
\label{tab:annotations}
\end{table*}

\section{Dataset Statistics and Visualization}
\begin{figure*}[t]
    \centering
    \includegraphics[width=1\textwidth]{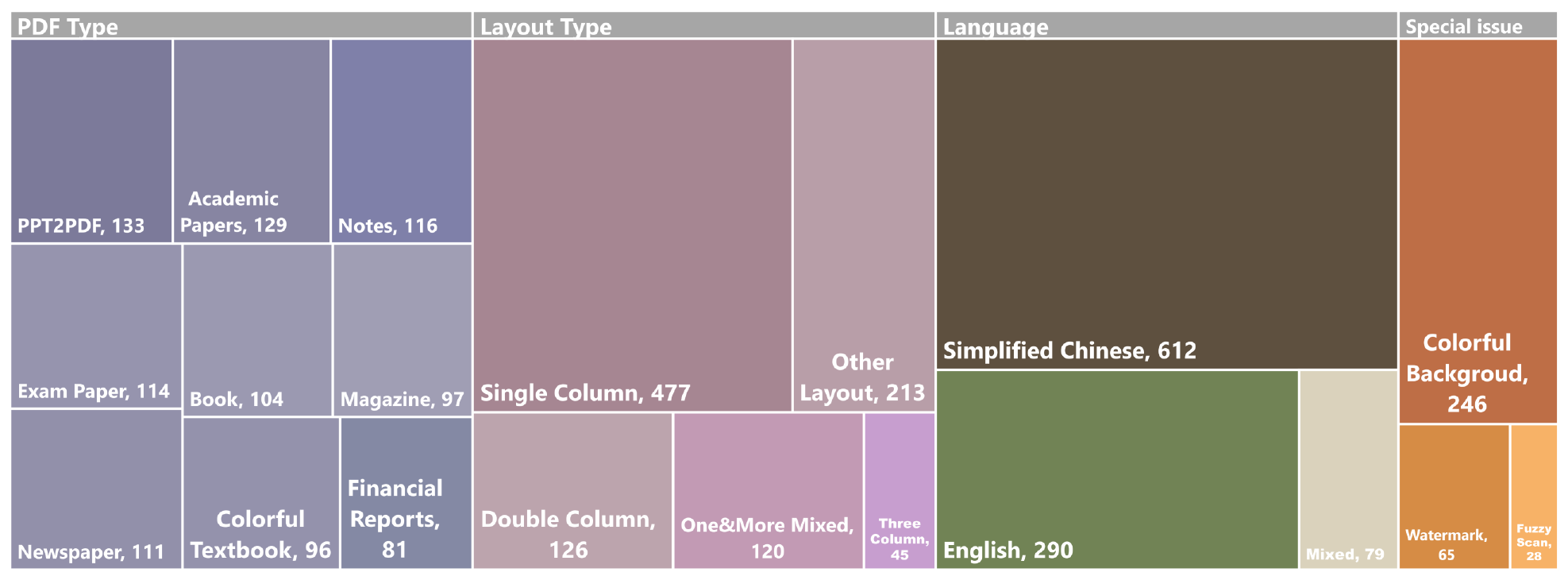}
    \caption{The Data Proportion of Pages for each Attribute in OmniDocBench.}
    \label{fig:page_count}
\vspace{-5mm}
\end{figure*}

OmniDocBench contains 981 pages, including 9 types of PDF pages, 4 types of layouts, 3 types of languages, and 3 special issues in visual degradations (e.g., watermarks). Table~\ref{tab:page_attribute_count} and Figure~\ref{fig:page_count} show the number of pages with each page attribute. \Cref{fig:show_pdf_types_1,fig:show_pdf_types_2,fig:showcase_layout,fig:showcase_special_issue} are examples of PDF pages with different PDF types, Layout Types, and Special Issues.

\begin{figure*}[t]
    \centering
    \includegraphics[width=1\textwidth]{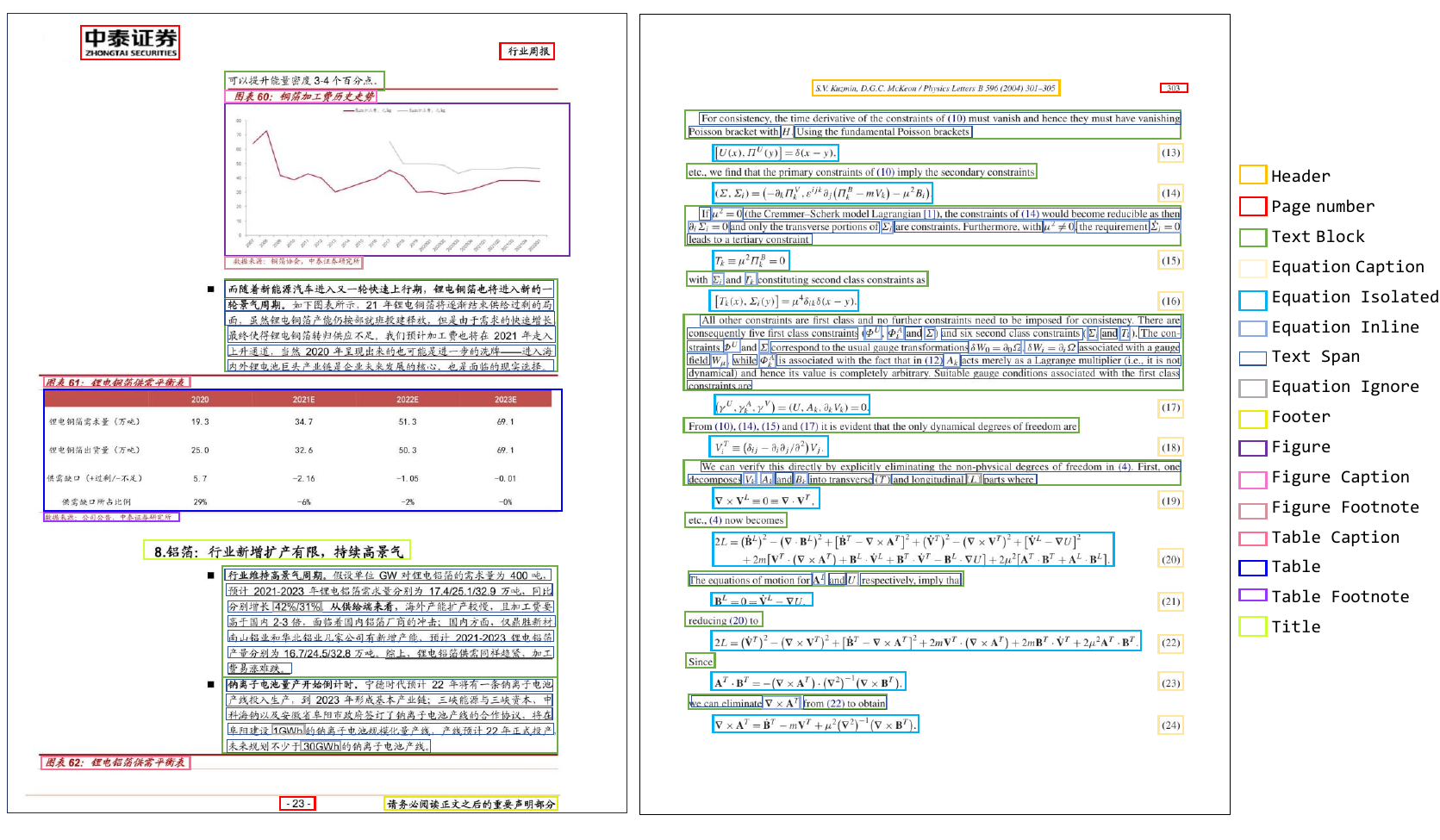}
    \vspace{-4mm}
    \caption{The Visualization of vary Annotations in OmniDocBench.}
    \label{fig:page_anno_show}
\vspace{-5mm}
\end{figure*}

Table~\ref{tab:annotations} and \Cref{fig:page_anno_show} show all annotation categories included in OmniDocBench. All of them are annotated by bounding boxes. There are 15 types of block-level annotations and 4 types of span-level annotations, with span-level annotations nested within the block-level ones. In addition, there are 3 types of annotations marked as page interference information (No.20-22), whose bounding boxes are used to mask the specific regions of the PDF pages to avoid affecting the evaluation results. The recognition annotations are also provided for each annotation category except for Figures. Formulas is written in LaTeX format and Table is annotated in both HTML and LaTeX formats. Others are annotated in plain text.

Furthermore, the Text Attributes are also annotated for each block-level category that contains text. There are 3 types of Text Attributes that might influent OCR accuracy: Language, Text Background Color, and Text Rotation. Table~\ref{tab:table_attribute_count} shows the statistics of annotations with specific text attributes. There are 23,010 block-level annotations are labeled with text attributes. 

Tables are also annotated with Table Attributes. There are 6 types of Table Attributes that might influent the Table Recognition accuracy: Language, Table Frame Type, Merge Cell, Colorful Background, Contain Formula, and Rotation. Table~\ref{tab:table_attribute_count} shows the numbers of annotations with specific table attributes. \Cref{fig:showcase_table_frame,fig:showcase_table_issue} are the examples of Tables with different Frames and Special Issues.


\section{Discussion on Model Predictions}
\noindent\textbf{Conclusion} Combining scattered results from tasks and sub-attributes, it can be concluded that pipeline tools and expert models have better performance on common data like academic papers and challenging cases such as tables with merged cells compared to VLMs. However, VLMs demonstrate stronger generalization on uncommon PDF types like slides and exam papers, and they show greater robustness in special page situations, such as fuzzy scans.
The low accuracy of VLMs is mainly due to:1) Missing Content in dense pages(\Cref{fig:papers}); 2) Hallucinations in hard-to-recognize pages(\Cref{fig:text_rotate}).
The low accuracy of Pipeline tools mainly due to: 1) Lower robustness in special page situations, e.g., watermark(\Cref{fig:watermark}); 2) Weak generalization on uncommon PDF types, e.g., handwriting notes(\Cref{fig:notes}).

\Cref{fig:book,fig:exam,fig:magazine,fig:newspaper,fig:notes,fig:papers,fig:Research_report,fig:Slides,fig:textbook} show the examples of Good model outputs and Bad model outputs of Document Parsing \textbf{among different PDF types}. As it shown, different models exhibit varying performance across different PDF types. For example, MinerU detects all handwritten notes as figures, resulting in very low recognition accuracy in Notes. Marker and InternVL2 experience missed detections, leading to lower scores. InternVL2 and Qwen2-VL, in specific PDF types (such as slides or financial reports), tend to merge multi-column text. 

\Cref{fig:colorful_background,fig:fuzzy_scan,fig:watermark} show the examples of Good model outputs and Bad model outputs \textbf{under special issues} of the PDF pages. It shows that Marker tends to generate typos when the PDF pages are fuzzy scanned or with watermarks, while GOT-OCR fails to recognize content on pages with colored backgrounds. MinerU performs well under special situations, while Mathpix occasionally generates typos.

\Cref{fig:single_col,fig:double_col,fig:three_col,fig:complex_layout} show examples of Good model outputs and Bad model outputs for PDF pages \textbf{with different layouts}. MinerU has a low reading order score for single-column layouts primarily because most notes are single-column, and MinerU performs poorly in recognizing Notes, leading to a low reading order score accordingly. InternVL2 scores high in Single-Column layouts but scores poorly on Double-Column and Three-Column layouts. It is mainly due to frequent missed content recognition and errors in reading order judgment in multi-column layouts pages. MinerU's reading order and recognition accuracy decrease with complex layouts, primarily because it incorrectly merges multiple columns during recognition.

\Cref{fig:text_colorful_background,fig:text_rotate} show the model's recognition ability \textbf{under special issues of text}. In text recognition with complex background colors, Marker may produce errors or miss content, whereas Qwen2-VL still performs well. Most models fail to recognize text when it is rotated 270 degrees. Some vision language models generate hallucinated information based on the content they can recognize.

\Cref{fig:badcase_table_threeline,fig:badcase_table_noframe,fig:badcase_table_rotate,fig:badcase_table_formula} show the examples of good and bad model results for \textbf{tables with different attributes}. For three-line tables, RapidTable demonstrates a good performance with accurate structure recognition, while PaddleOCR shows limitations by missing the last column in its outputs. Interestingly, in tables without frames, PaddleOCR performs well with accurate table predictions, while Qwen2-VL-7B exhibits errors in the last two columns. This indicates that the presence or absence of table frames can significantly impact different models' performance in different ways. Rotated tables prove to be particularly challenging, with most models, including GOT-OCR, failing to recognize the table structure. However, StructEqTable shows promising results by correctly identifying most of the table content, though with a few detail errors. For tables containing formula, Qwen2-VL-7B shows more accurate table structure recognition compared to InternVL2-8B.

\section{Model Settings}
For pipeline tools such as MinerU, Marker, and Mathpix, default settings are used for evaluation. Specifically, MinerU with Version 0.9.3\footnote{\url{https://github.com/opendatalab/MinerU/releases/tag/magic_pdf-0.9.3-released}} is employed. 
For Marker, Version 1.2.3\footnote{\url{https://github.com/VikParuchuri/marker/releases/tag/v1.2.3}} is evaluated. 
For Nougat, we utilize its 0.1.0-base model (350M).
For GOT-OCR, we employ its format OCR mode to output structured data.

For general VLMs, we used the GPT4o, Qwen2-VL-72B, and InternVL2-Llama3-76B by setting the \textit{do\_sample$=$False} to ensure the reproducibility. After testing the different setting of \textit{max\_token}, the best setting is chosen for each VLMs. Specifically, \textit{max\_token$=$32000} is set for Qwen2-VL-72B, and \textit{max\_token$=$4096} is set for InternVL2-Llama3-76B. For GPT-4o, the default setting is used.

\section{More Details on Methods}
\begin{figure*}[t]
    \centering
    \includegraphics[width=1\textwidth]{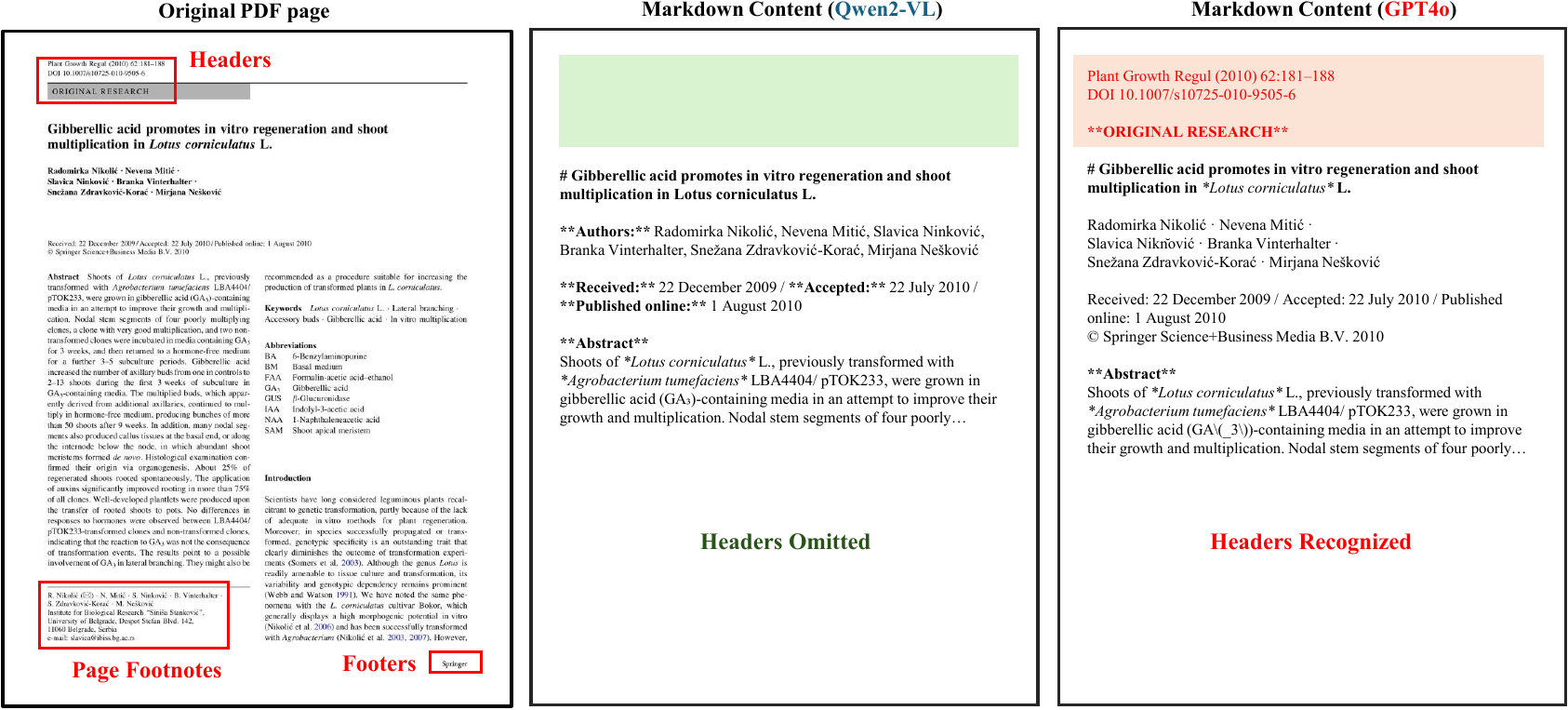}
    \vspace{-3mm}
    \caption{The Vary Standards in parsing Header, Footers, and so on.}
    \label{fig:abandon_standard}
\vspace{-5mm}
\end{figure*}

\begin{figure*}[t]
    \centering
    \includegraphics[width=1\textwidth]{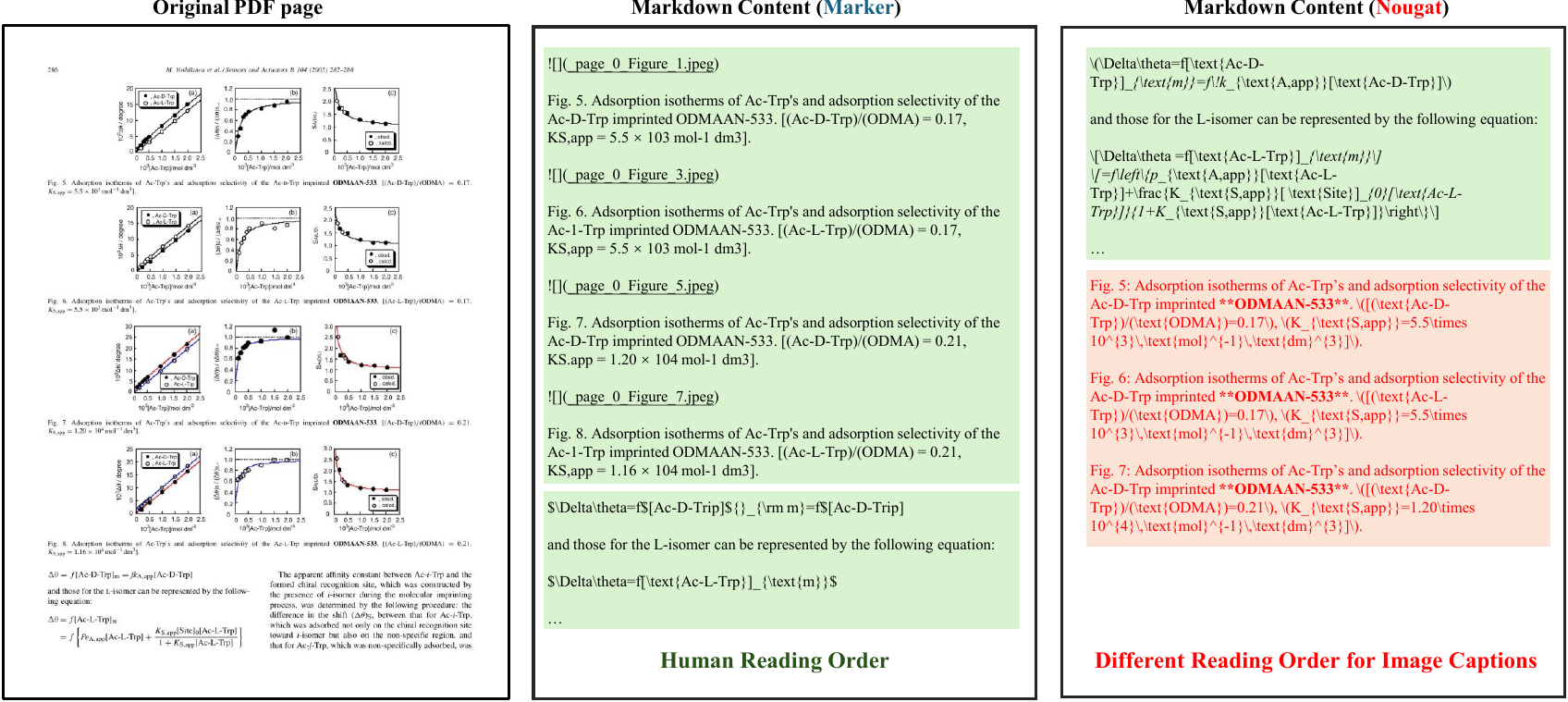}
    \vspace{-3mm}
    \caption{The Vary Standards in parsing Captions.}
    \label{fig:captions_standard}
\vspace{-5mm}
\end{figure*}

\begin{figure*}[t]
    \centering
    \includegraphics[width=0.95\textwidth]{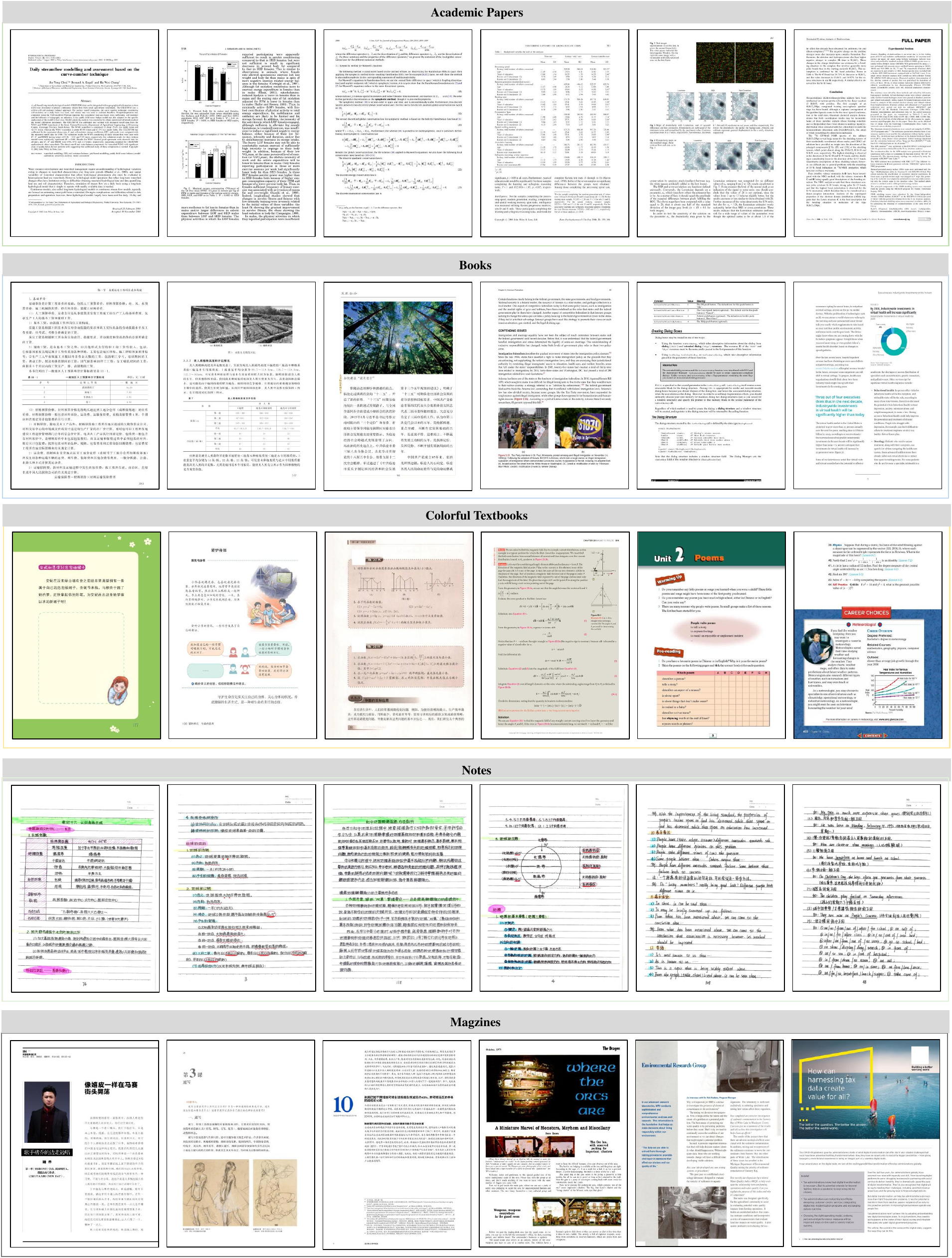}
    \caption{The Examples of Academic Papers, Books, Textbooks, Notes, and Magazines in OmniDocBench.}
    \label{fig:show_pdf_types_1}
\vspace{-5mm}
\end{figure*}

\begin{figure*}[t]
    \centering
    \includegraphics[width=1\textwidth]{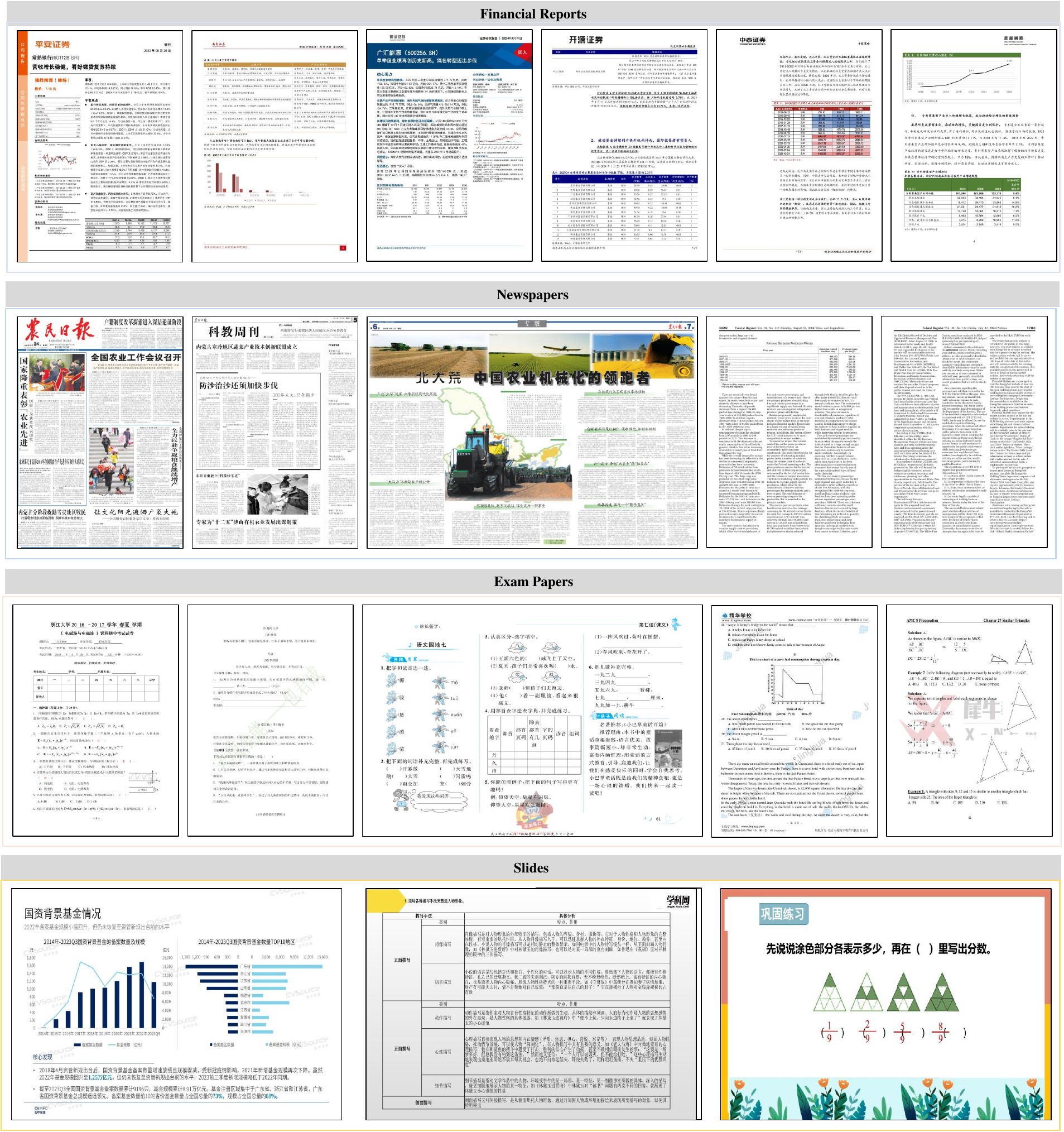}
    \caption{The Examples of Finacial Reports, Newspapers, Example Papers, and Slides in OmniDocBench.}
    \label{fig:show_pdf_types_2}
\vspace{-5mm}
\end{figure*}

\begin{figure*}[t]
    \centering
    \includegraphics[width=1\textwidth]{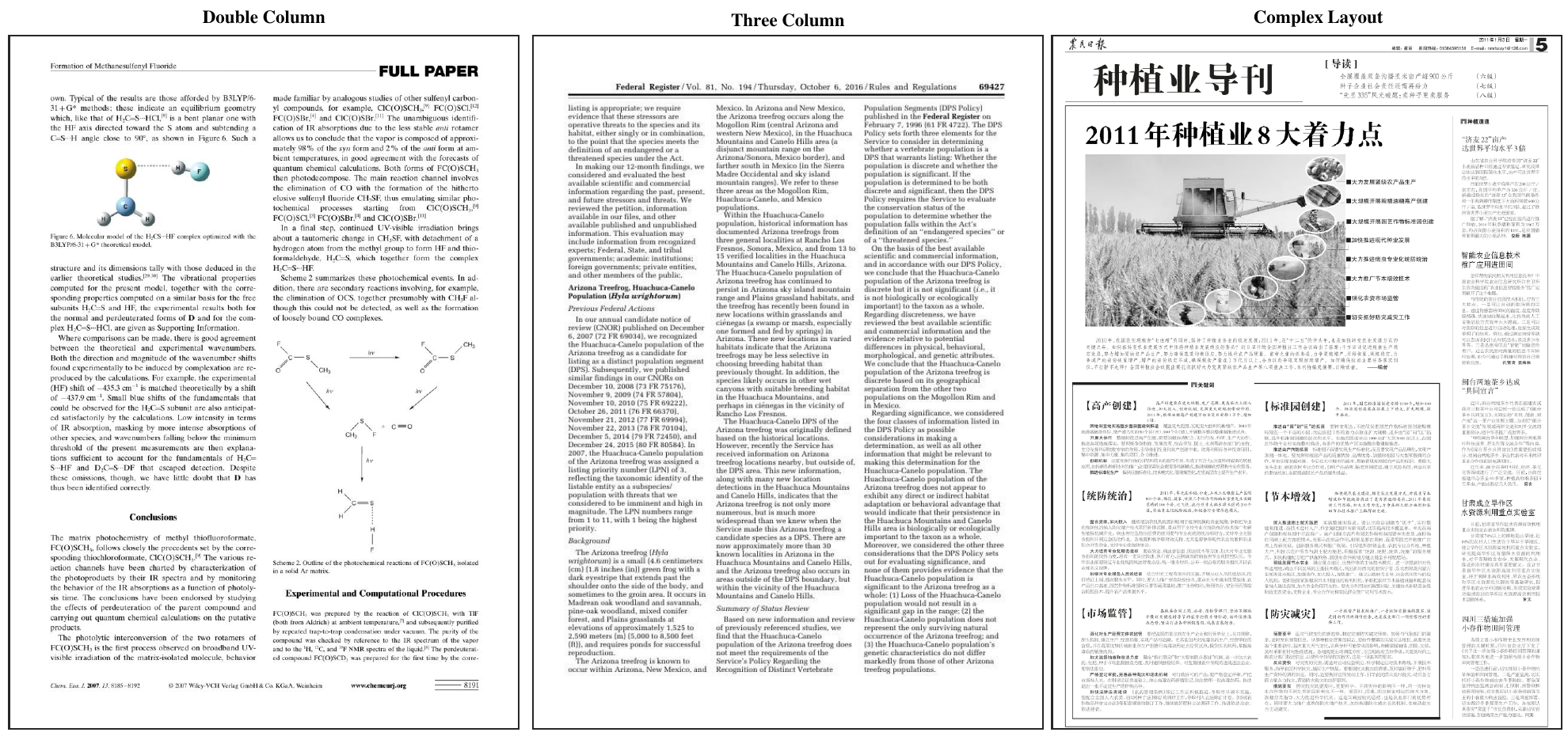}
    \caption{The Examples of PDF pages with different Layout Types in OmniDocBench.}
    \label{fig:showcase_layout}
\vspace{-5mm}
\end{figure*}

\begin{figure*}[t]
    \centering
    \includegraphics[width=1\textwidth]{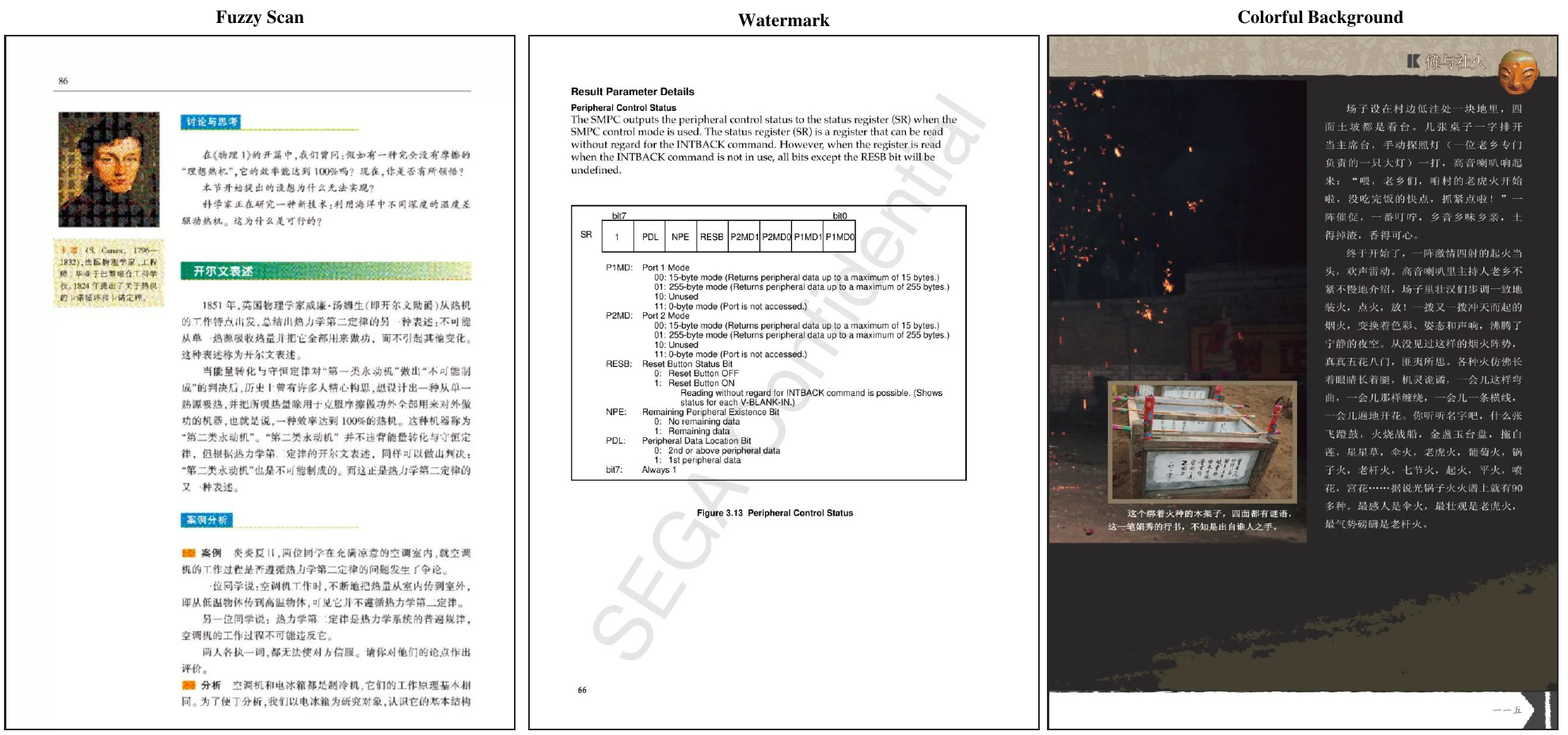}
    \caption{The Examples of PDF pages under Special Issues in OmniDocBench.}
    \label{fig:showcase_special_issue}
\vspace{-5mm}
\end{figure*}

\begin{figure*}[t]
    \centering
    \includegraphics[width=1\textwidth]{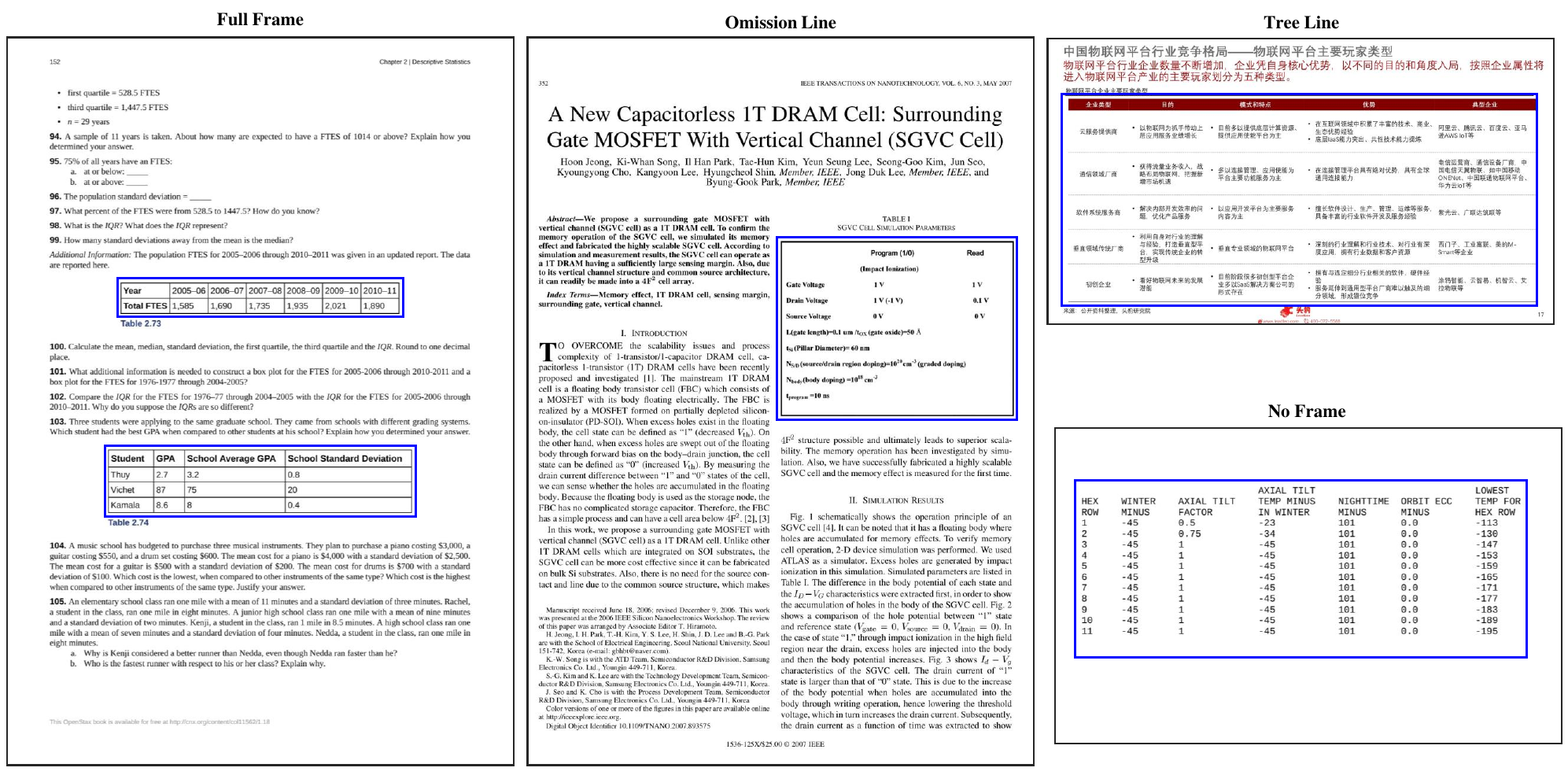}
    \caption{The Examples of Tables with different Frame in OmniDocBench.}
    \label{fig:showcase_table_frame}
\vspace{-5mm}
\end{figure*}

\begin{figure*}[t]
    \centering
    \includegraphics[width=1\textwidth]{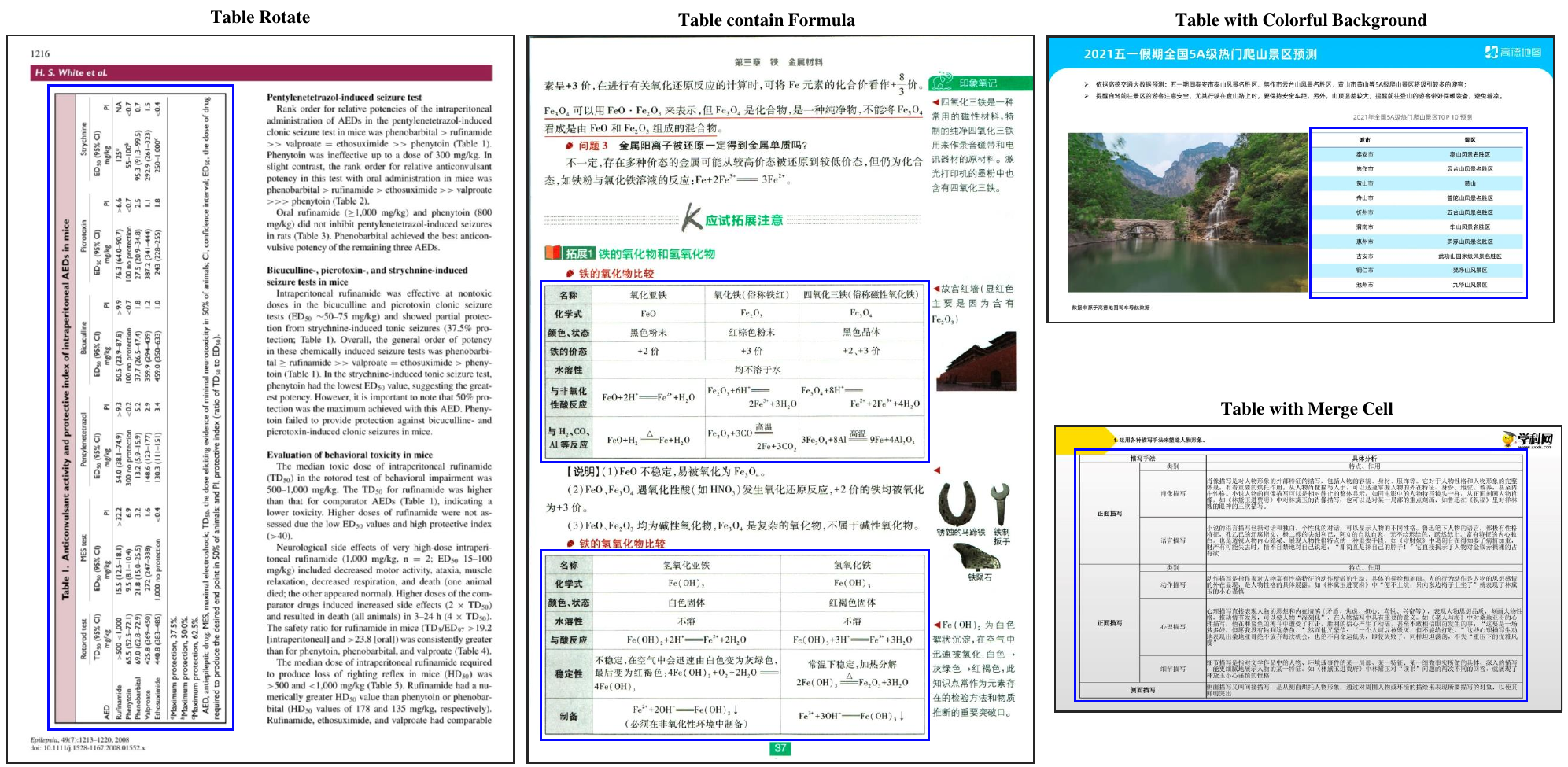}
    \caption{The Examples of Tables under Special Issues in OmniDocBench.}
    \label{fig:showcase_table_issue}
\vspace{-5mm}
\end{figure*}

\begin{figure*}[t]
    \centering
    \includegraphics[width=1\textwidth]{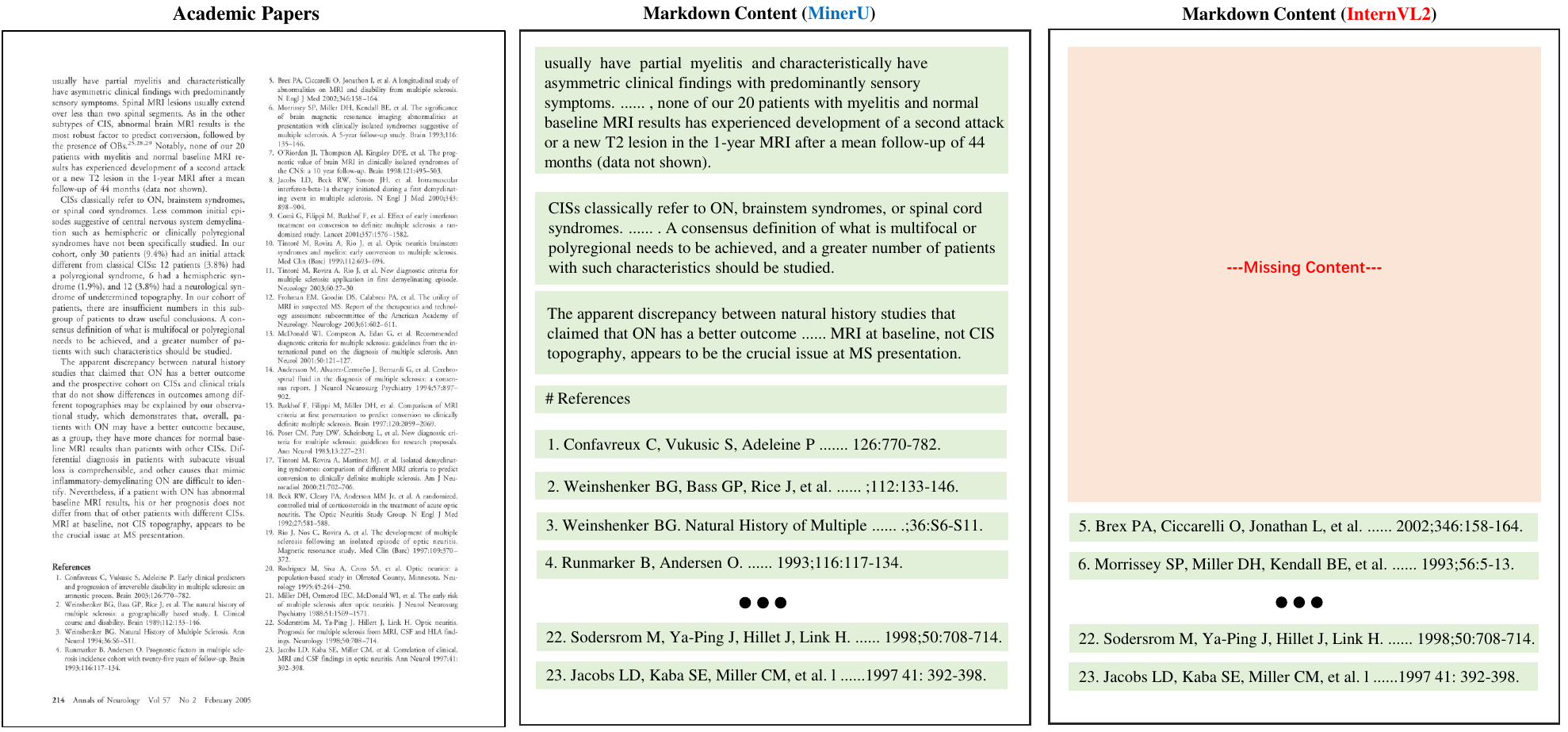}
    \caption{The \textcolor{blue}{Good} Model Result and \textcolor{red}{Bad} Model Result for Academic Papers.}
    \label{fig:papers}
\vspace{-5mm}
\end{figure*}

\begin{figure*}[t]
    \centering
    \includegraphics[width=1\textwidth]{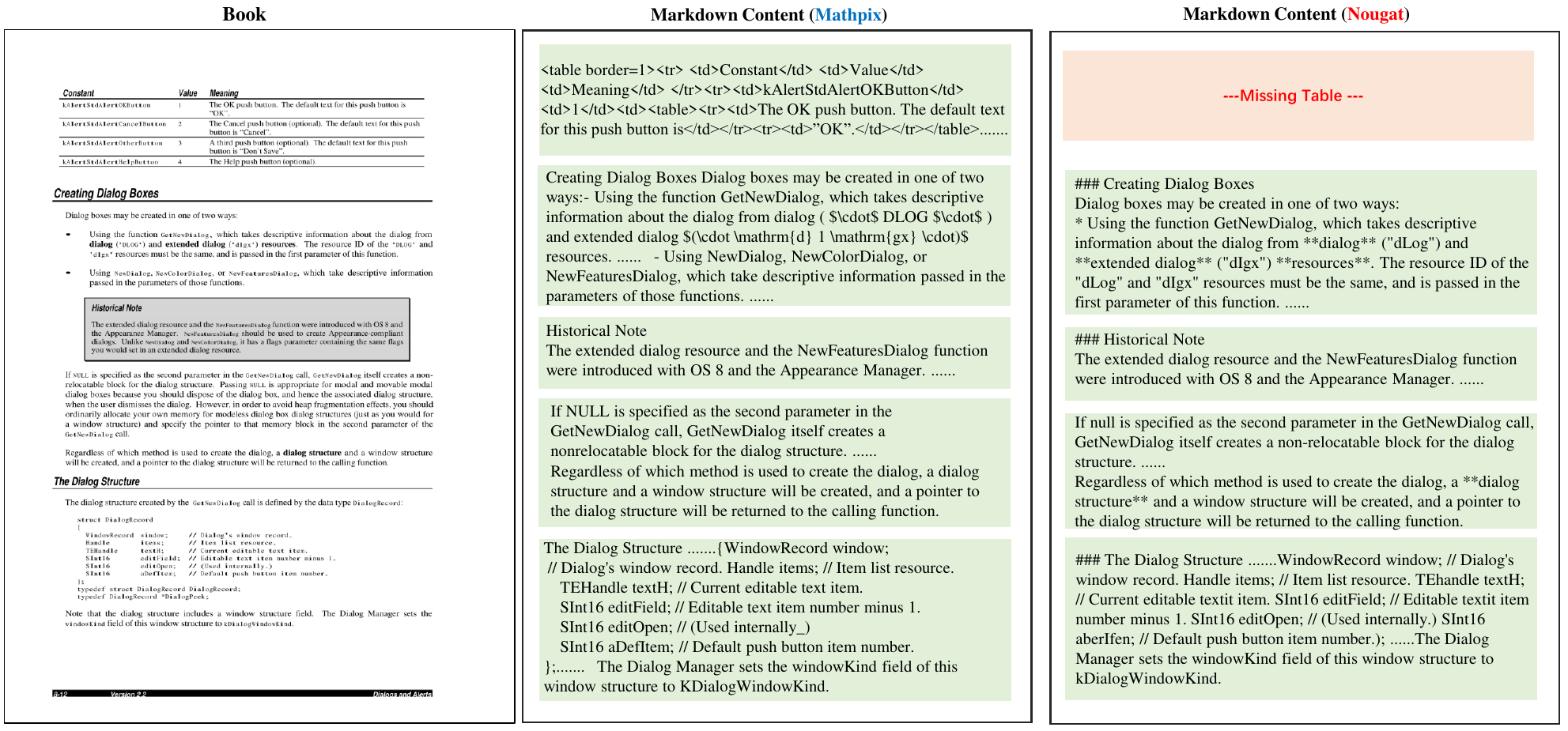}
    \caption{The \textcolor{blue}{Good} Model Result and \textcolor{red}{Bad} Model Result for Books.}
    \label{fig:book}
\vspace{-5mm}
\end{figure*}

\begin{figure*}[t]
    \centering
    \includegraphics[width=1\textwidth]{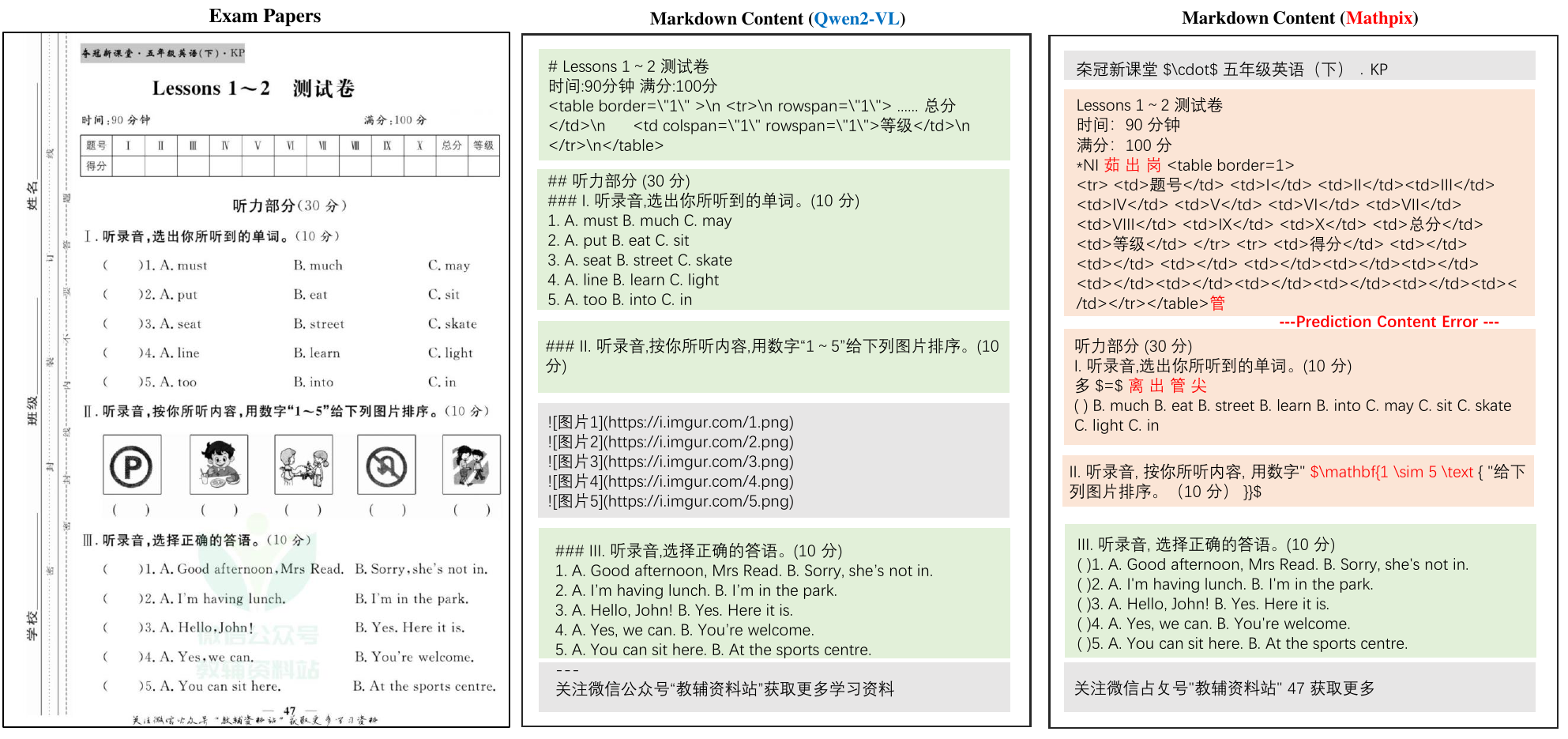}
    \caption{The \textcolor{blue}{Good} Model Result and \textcolor{red}{Bad} Model Result for Exam Papers.}
    \label{fig:exam}
\vspace{-5mm}
\end{figure*}

\begin{figure*}[t]
    \centering
    \includegraphics[width=1\textwidth]{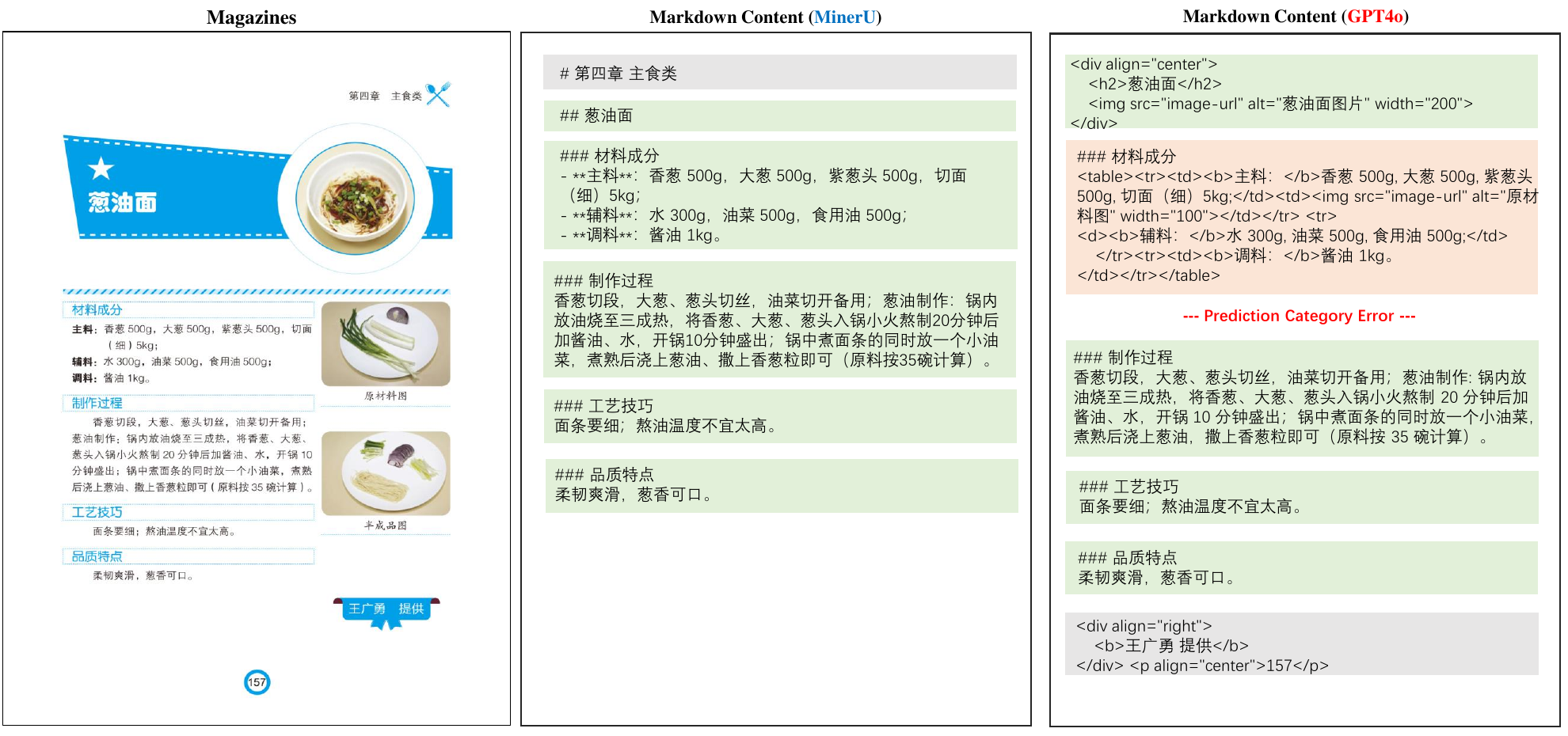}
    \caption{The \textcolor{blue}{Good} Model Result and \textcolor{red}{Bad} Model Result for Magazines.}
    \label{fig:magazine}
\vspace{-5mm}
\end{figure*}

\begin{figure*}[t]
    \centering
    \includegraphics[width=1\textwidth]{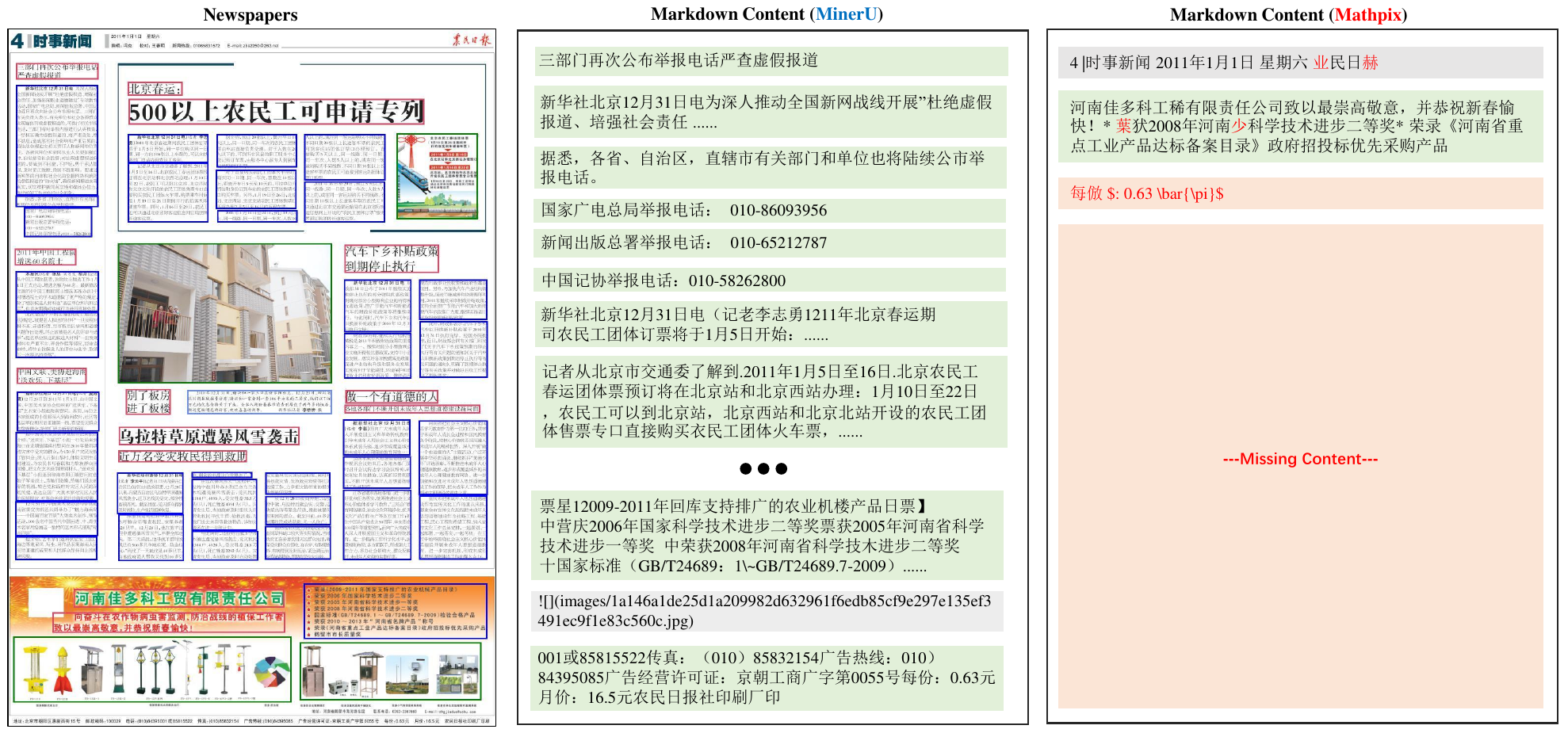}
    \caption{The \textcolor{blue}{Good} Model Result and \textcolor{red}{Bad} Model Result for Newspaper.}
    \label{fig:newspaper}
\vspace{-5mm}
\end{figure*}

\begin{figure*}[t]
    \centering
    \includegraphics[width=1\textwidth]{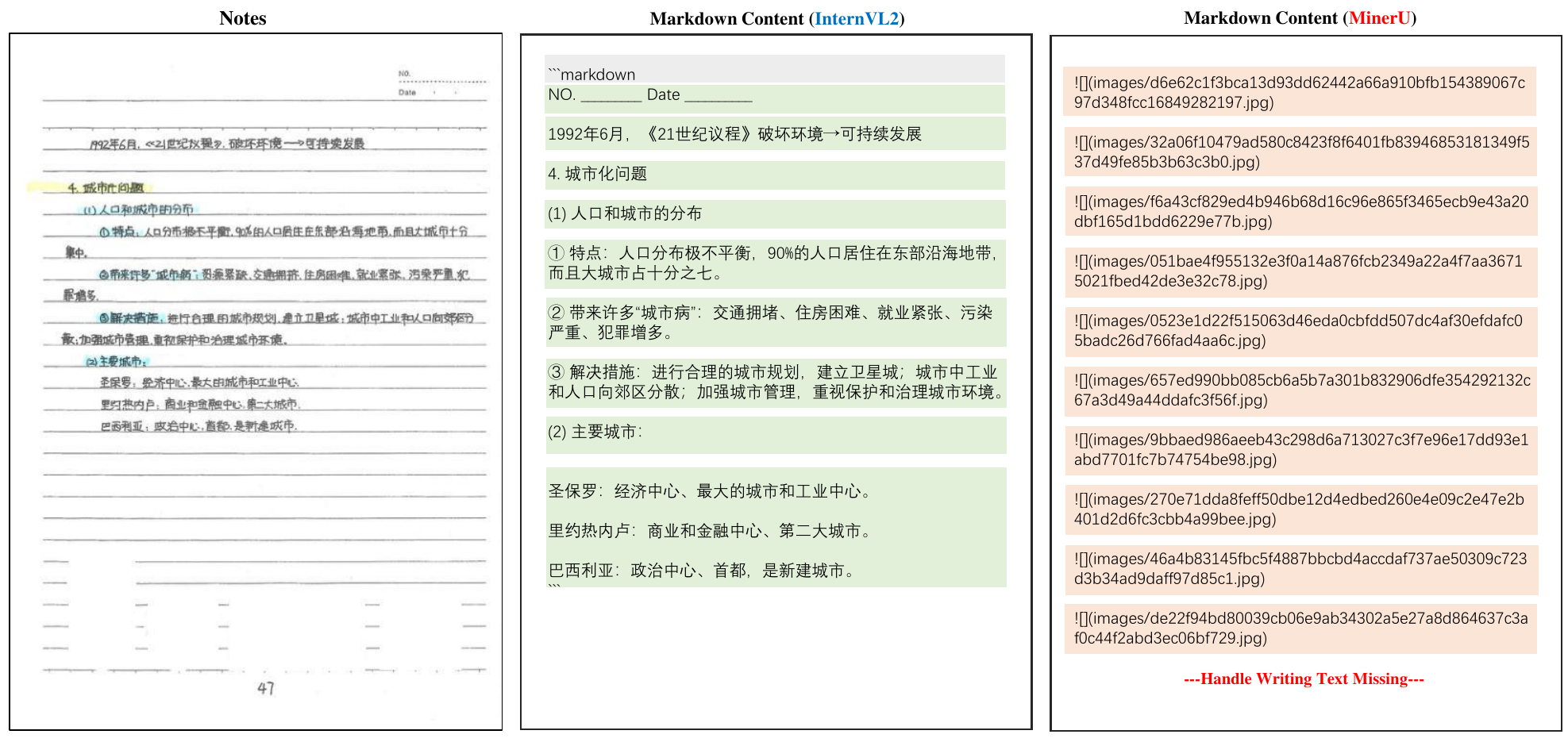}
    \caption{The \textcolor{blue}{Good} Model Result and \textcolor{red}{Bad} Model Result for Handwriting Notes.}
    \label{fig:notes}
\vspace{-5mm}
\end{figure*}

\begin{figure*}[t]
    \centering
    \includegraphics[width=1\textwidth]{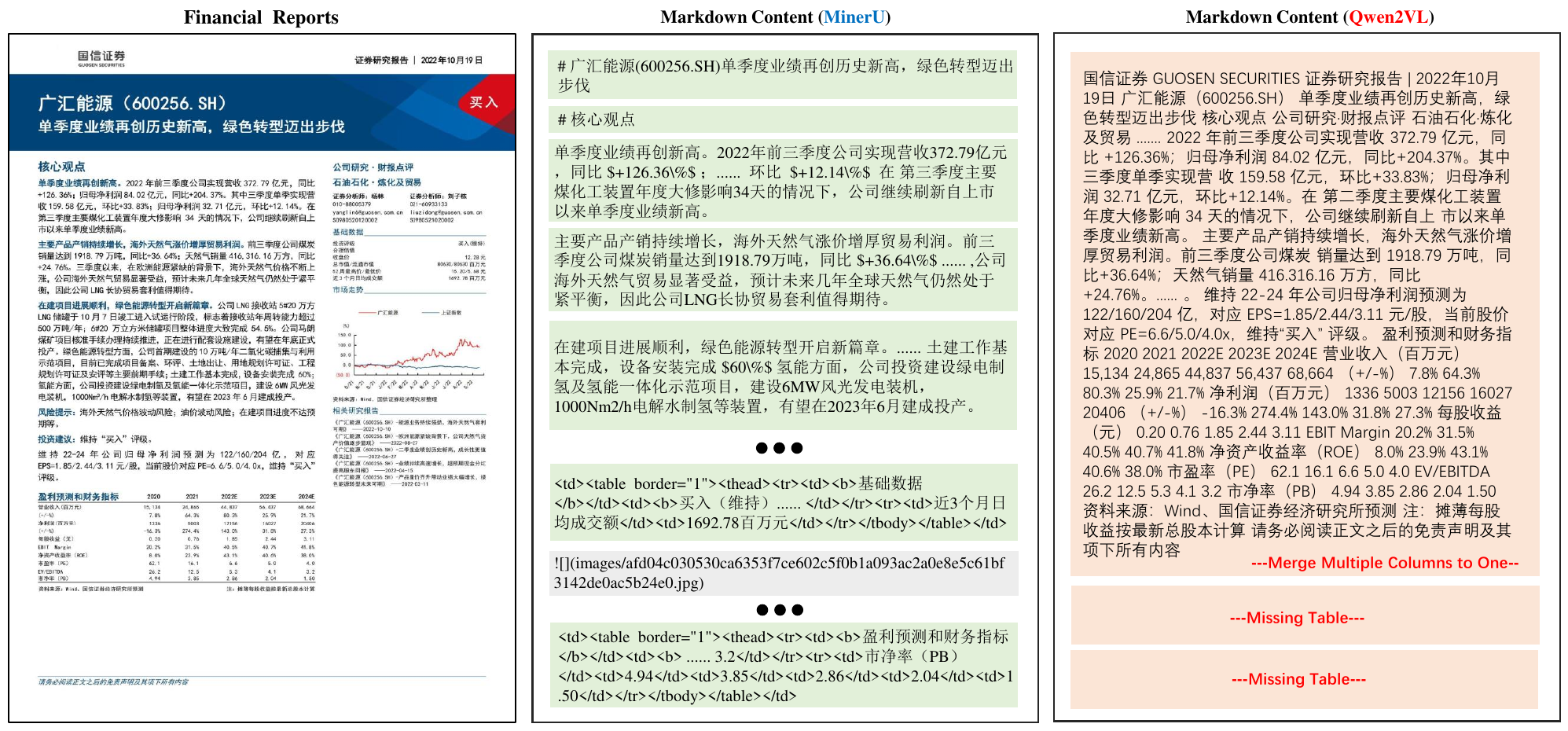}
    \caption{The \textcolor{blue}{Good} Model Result and \textcolor{red}{Bad} Model Result for Financial Reports.}
    \label{fig:Research_report}
\vspace{-5mm}
\end{figure*}

\begin{figure*}[t]
    \centering
    \includegraphics[width=1\textwidth]{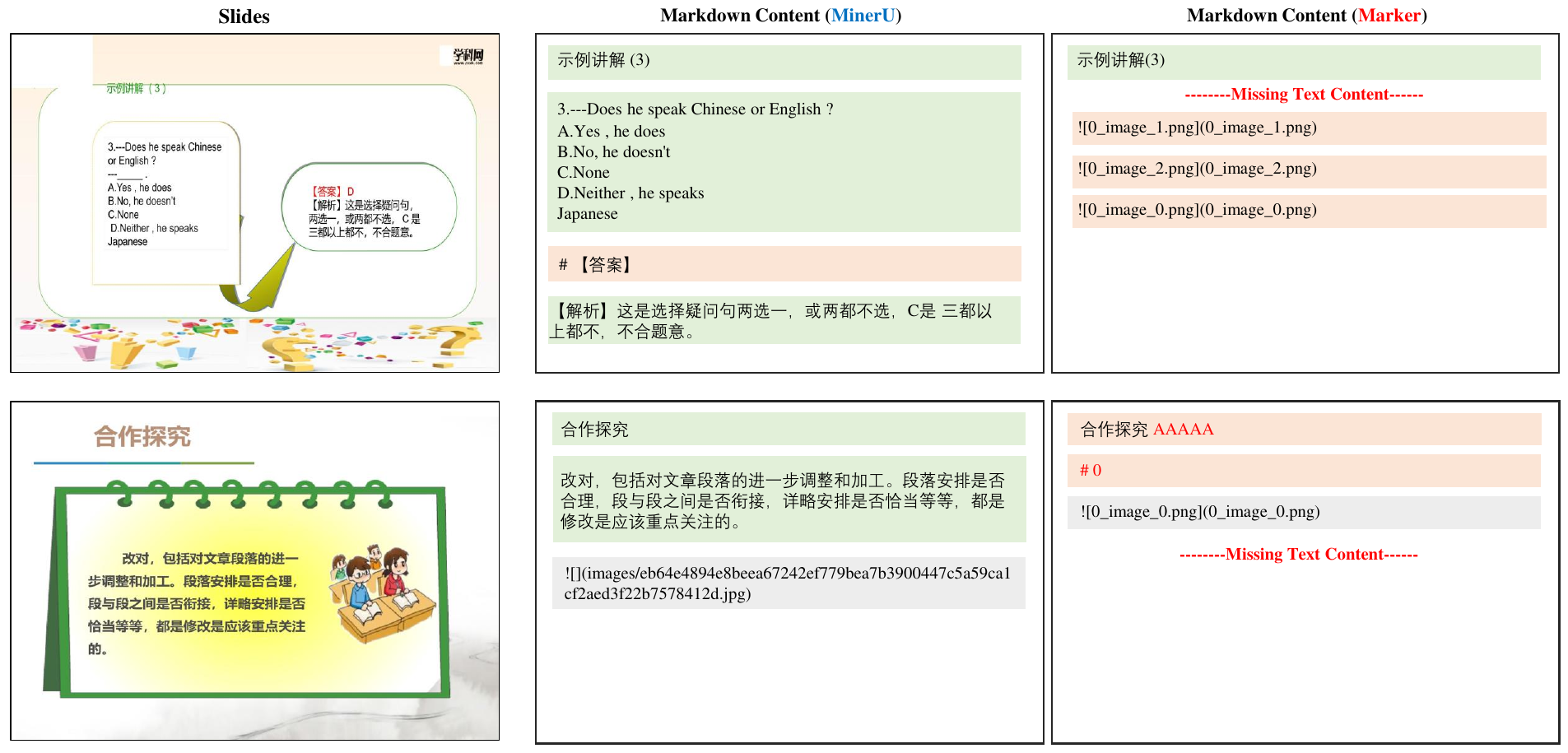}
    \caption{The \textcolor{blue}{Good} Model Result and \textcolor{red}{Bad} Model Result for Slides.}
    \label{fig:Slides}
\vspace{-5mm}
\end{figure*}

\begin{figure*}[t]
    \centering
    \includegraphics[width=1\textwidth]{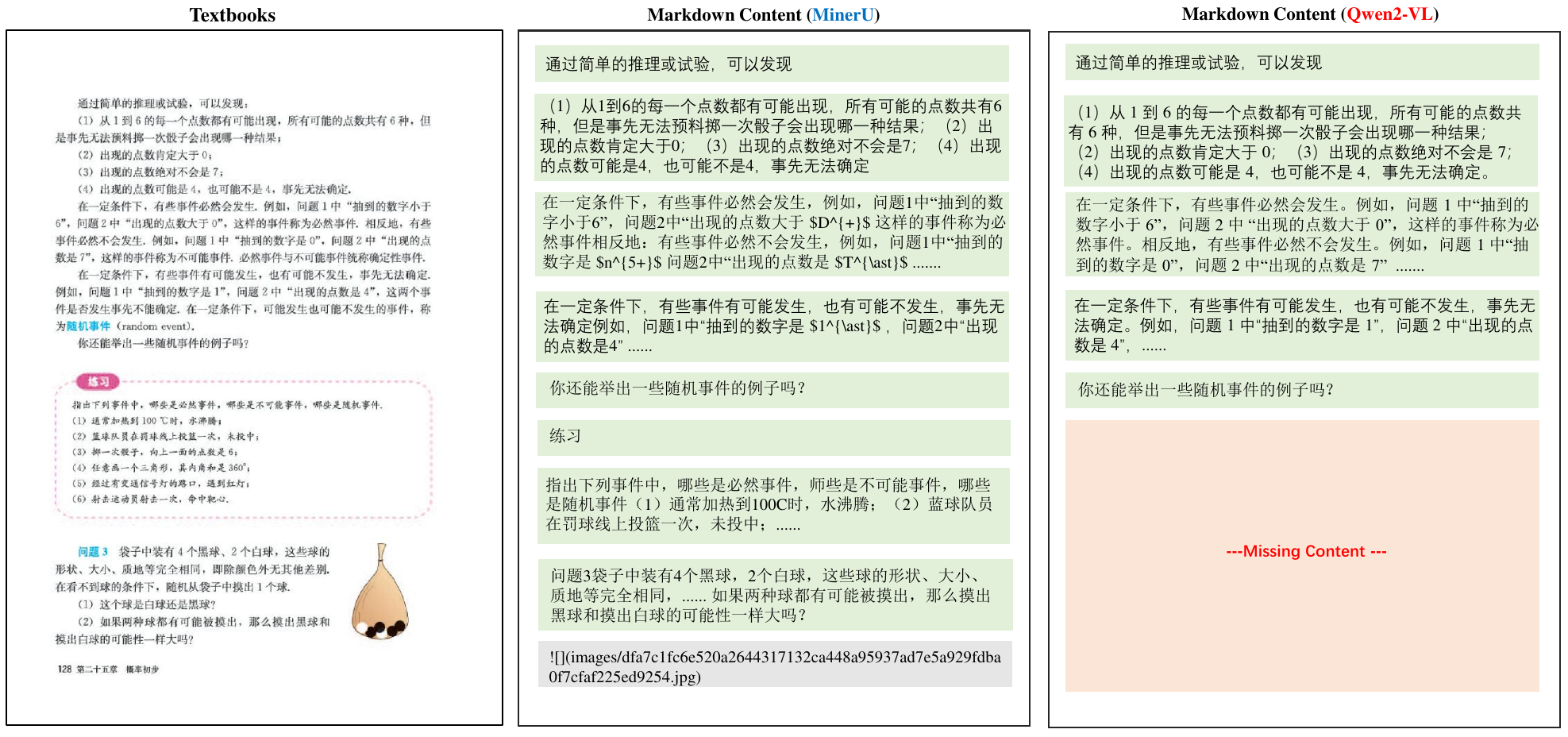}
    \caption{The \textcolor{blue}{Good} Model Result and \textcolor{red}{Bad} Model Result for Textbooks.}
    \label{fig:textbook}
\vspace{-5mm}
\end{figure*}

\begin{figure*}[t]
    \centering
    \includegraphics[width=1\textwidth]{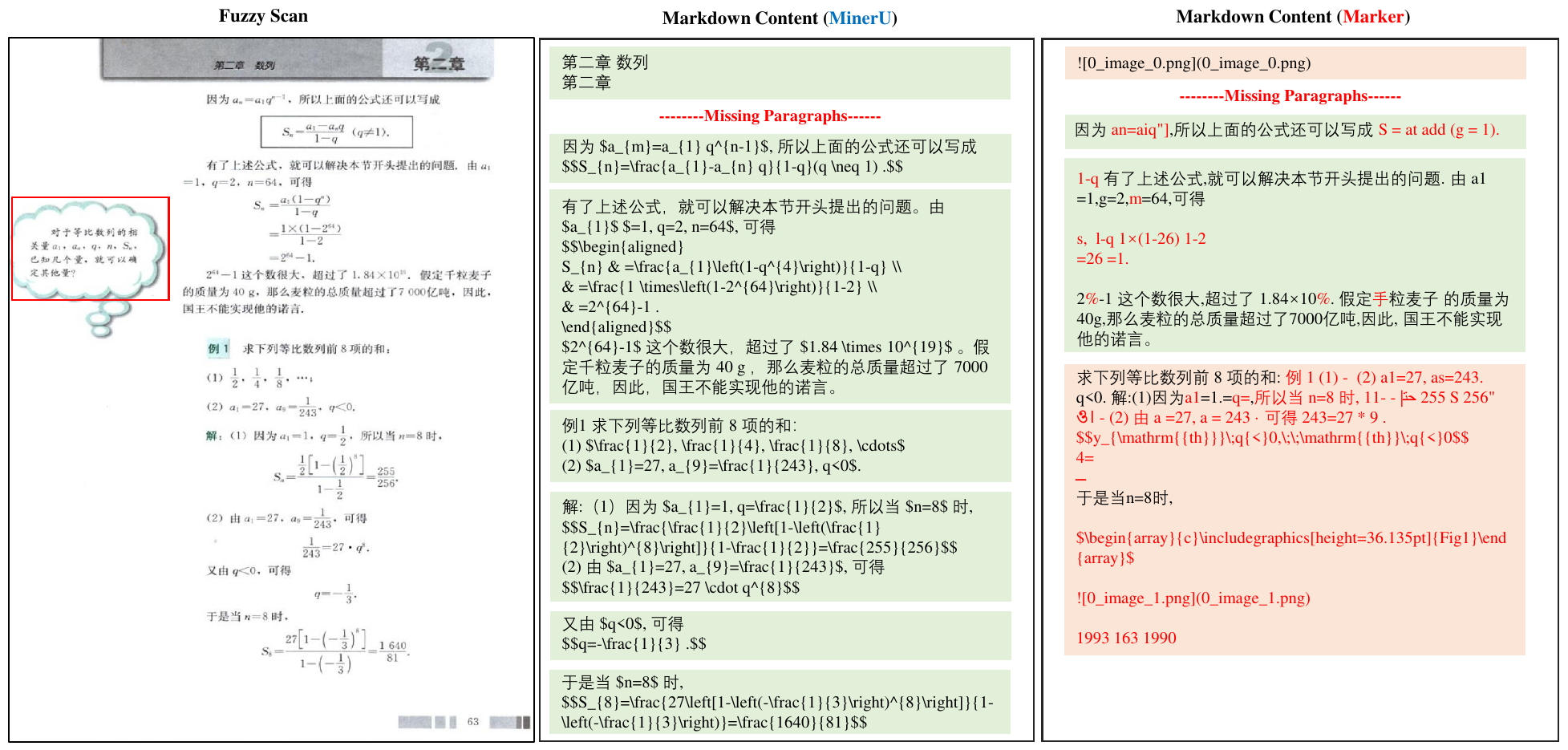}
    \caption{The \textcolor{blue}{Good} Model Result and \textcolor{red}{Bad} Model Result for Fuzzy Scan Pages.}
    \label{fig:fuzzy_scan}
\vspace{-5mm}
\end{figure*}

\begin{figure*}[t]
    \centering
    \includegraphics[width=1\textwidth]{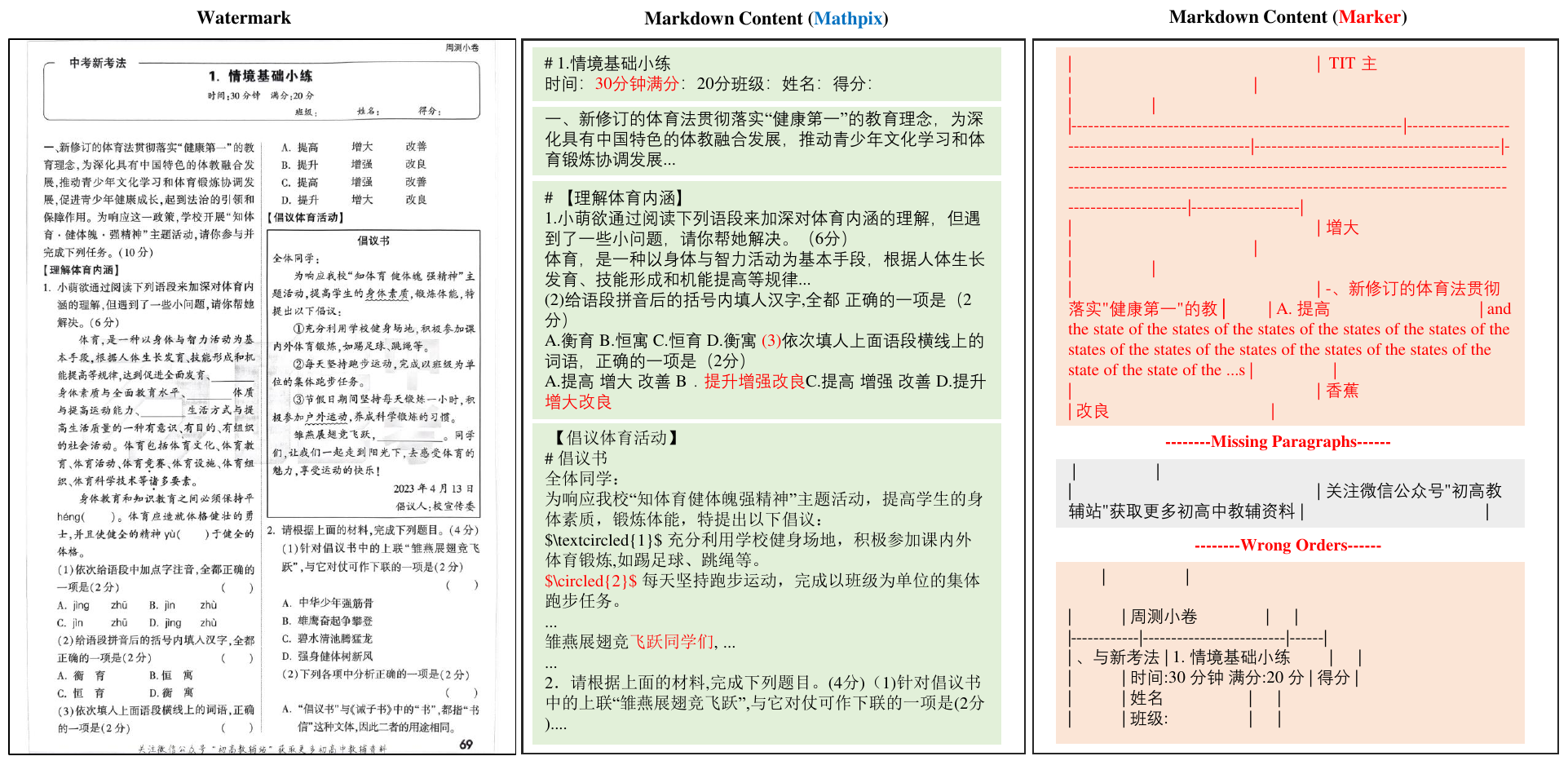}
    \caption{The \textcolor{blue}{Good} Model Result and \textcolor{red}{Bad} Model Result for Pages with Watermark.}
    \label{fig:watermark}
\vspace{-5mm}
\end{figure*}

\begin{figure*}[t]
    \centering
    \includegraphics[width=1\textwidth]{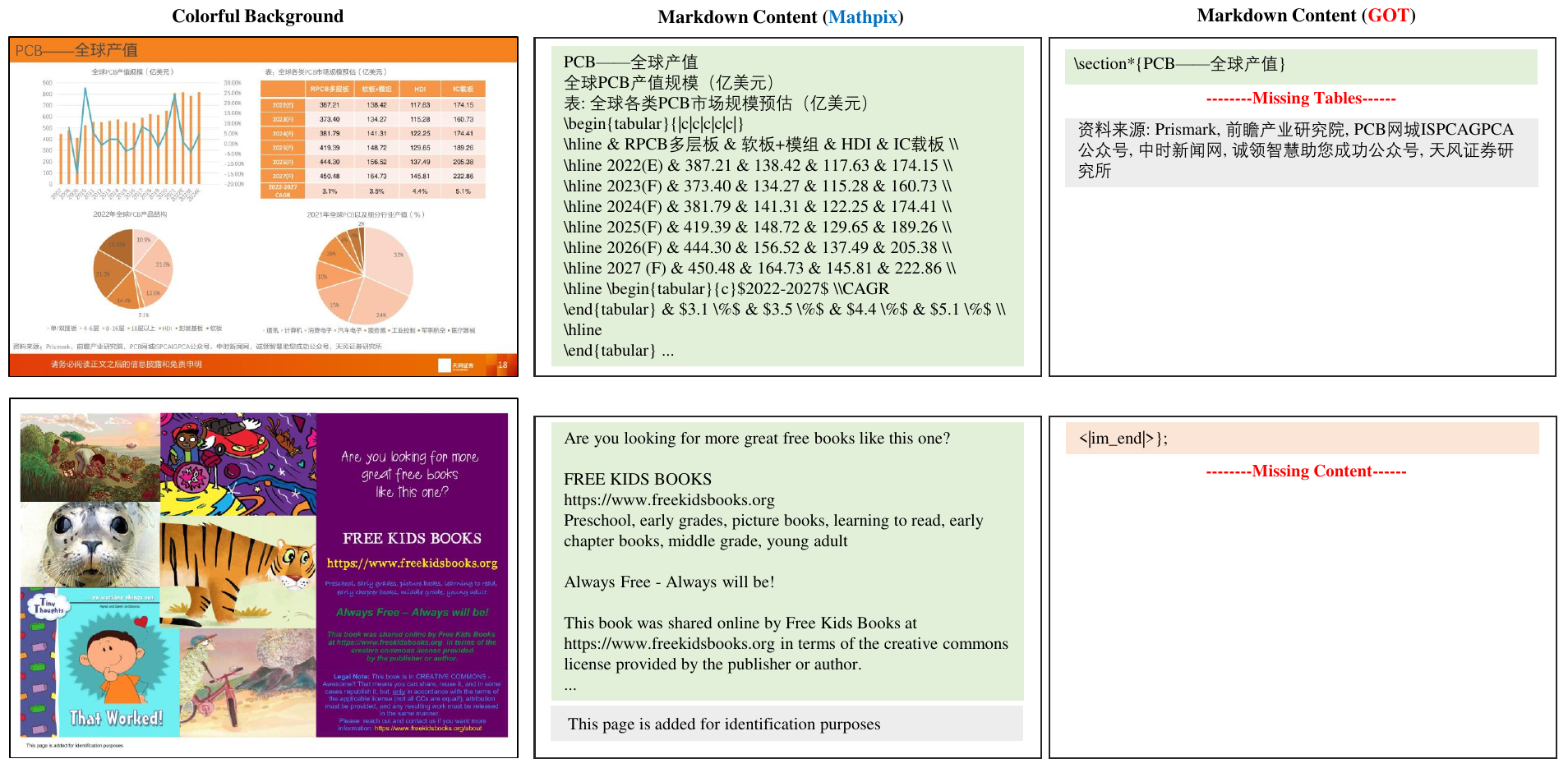}
    \caption{The \textcolor{blue}{Good} Model Result and \textcolor{red}{Bad} Model Result for Colorful Background Pages.}
    \label{fig:colorful_background}
\vspace{-5mm}
\end{figure*}

\begin{figure*}[t]
    \centering
    \includegraphics[width=1\textwidth]{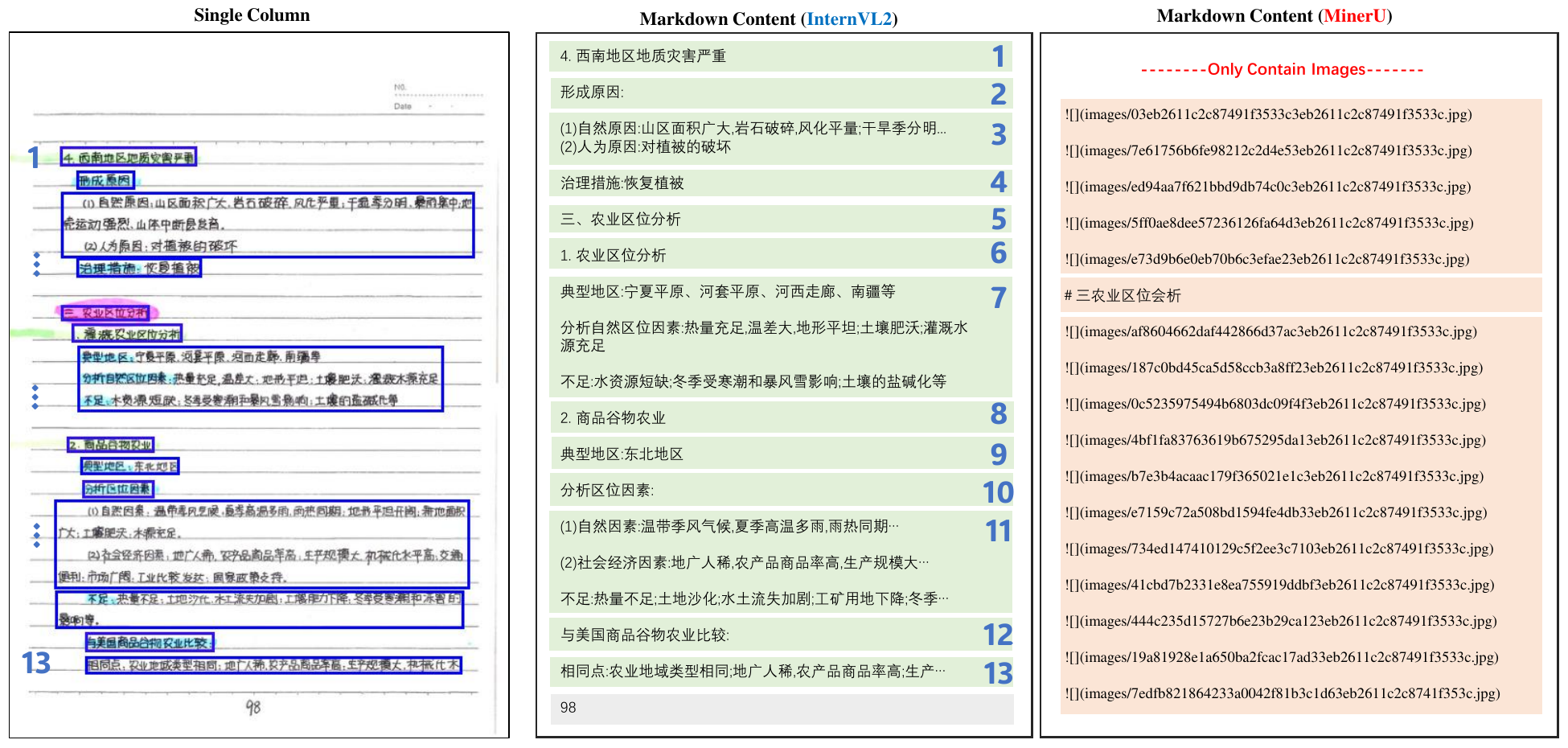}
    \caption{The \textcolor{blue}{Good} Model Result and \textcolor{red}{Bad} Model Result for Single Column Pages.}
    \label{fig:single_col}
\vspace{-5mm}
\end{figure*}

\begin{figure*}[t]
    \centering
    \includegraphics[width=1\textwidth]{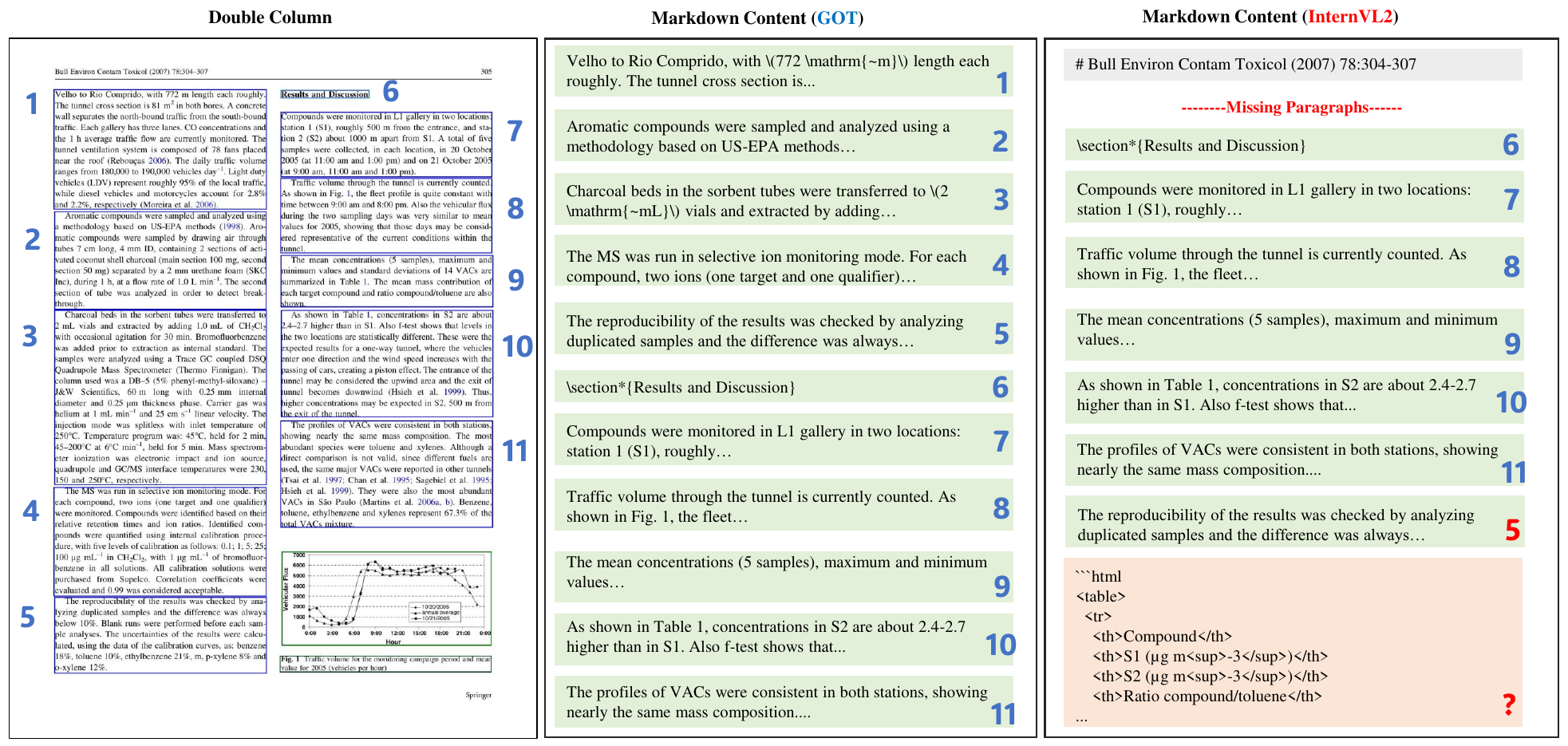}
    \caption{The \textcolor{blue}{Good} Model Result and \textcolor{red}{Bad} Model Result for Double Column Pages.}
    \label{fig:double_col}
\vspace{-5mm}
\end{figure*}

\begin{figure*}[t]
    \centering
    \includegraphics[width=1\textwidth]{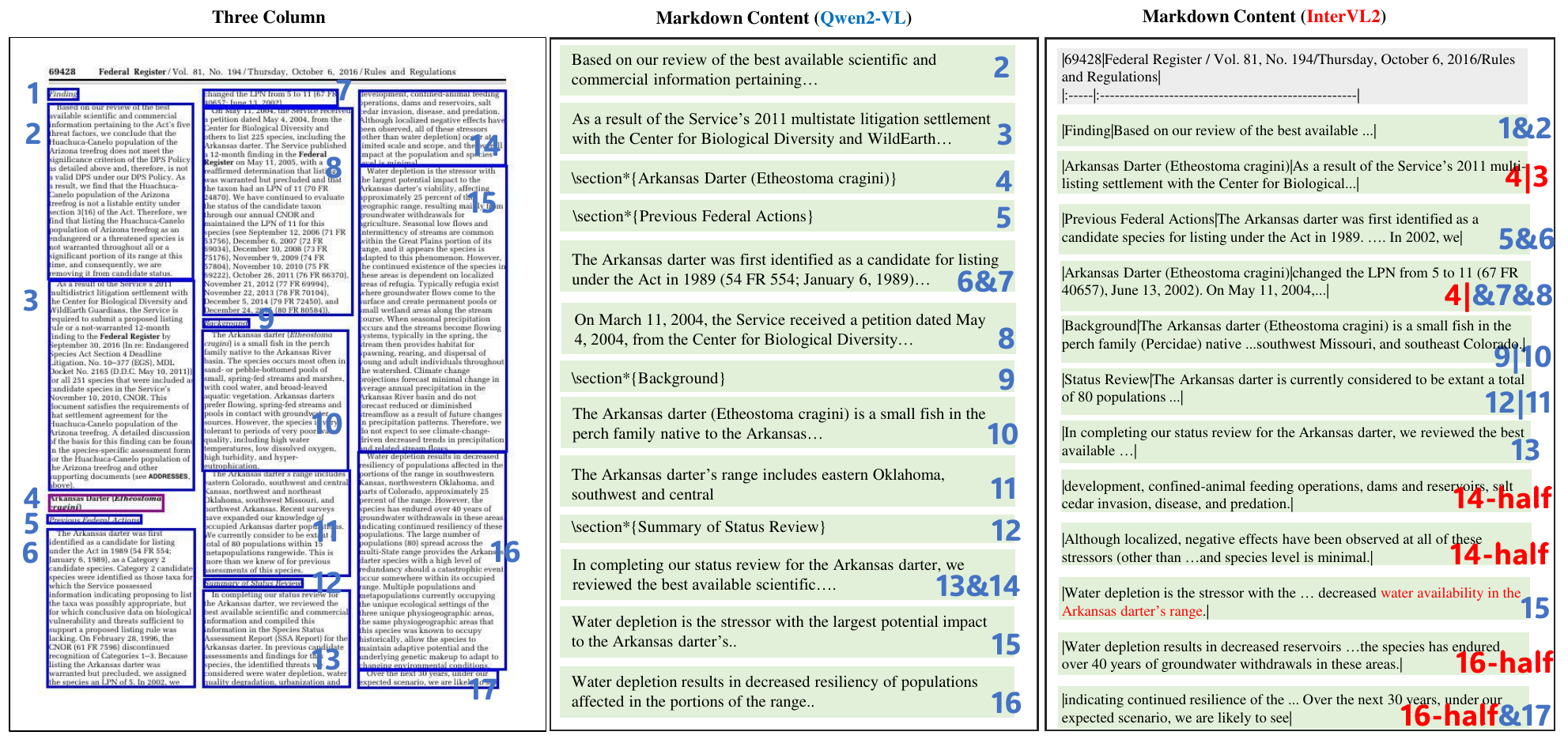}
    \caption{The \textcolor{blue}{Good} Model Result and \textcolor{red}{Bad} Model Result for Three Column Pages.}
    \label{fig:three_col}
\vspace{-5mm}
\end{figure*}

\begin{figure*}[t]
    \centering
    \includegraphics[width=1\textwidth]{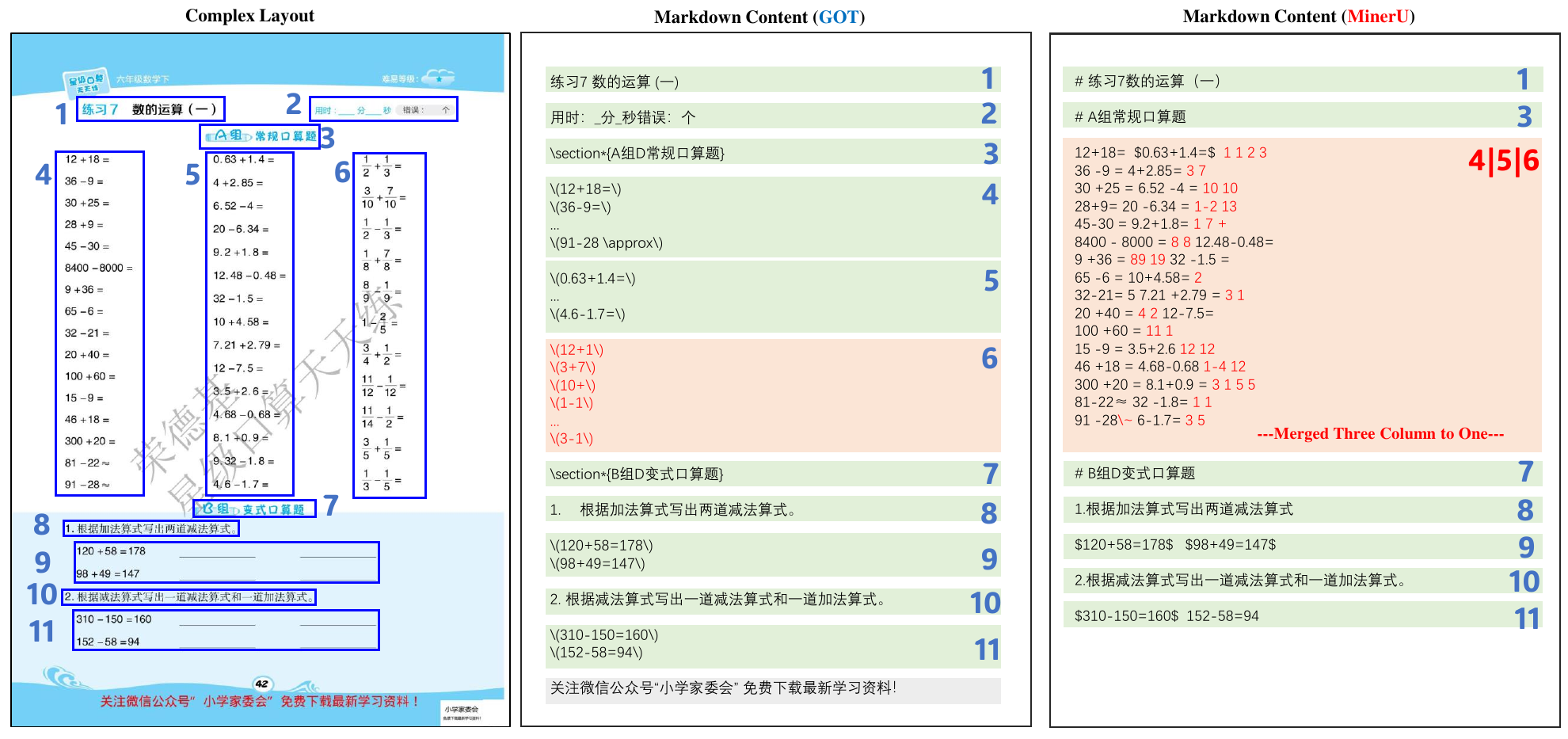}
    \caption{The \textcolor{blue}{Good} Model Result and \textcolor{red}{Bad} Model Result for Complex Layout Pages.}
    \label{fig:complex_layout}
\vspace{-5mm}
\end{figure*}

\noindent\textbf{Ignore handling.} The purpose of this process is to avoid fluctuations in accuracy caused by the lack of uniformity in the output standards among document parsing algorithm. (1) Some algorithm (e.g., GPT-OCR, Qwen2-VL) tends to remove headers and footers, while others (e.g., GPT4o) prefers to retain them (~\Cref{fig:abandon_standard}). (2) Moreover, the reading order mismatch cause by captions and footnotes is also considered. For example, Nougat would put the image captions in the end of the page content(~\Cref{fig:captions_standard}), while others tend to put the image captions in human reading order.

Ignore handling is to minimize the impact of varying standards of document parsing on evaluation. Our evaluation dataset aims to more fairly assess the parsing accuracy of various algorithms, and these trivial issues regarding standards are not within our scope of consideration.

\begin{figure*}[t]
    \centering
    \includegraphics[width=1\textwidth]{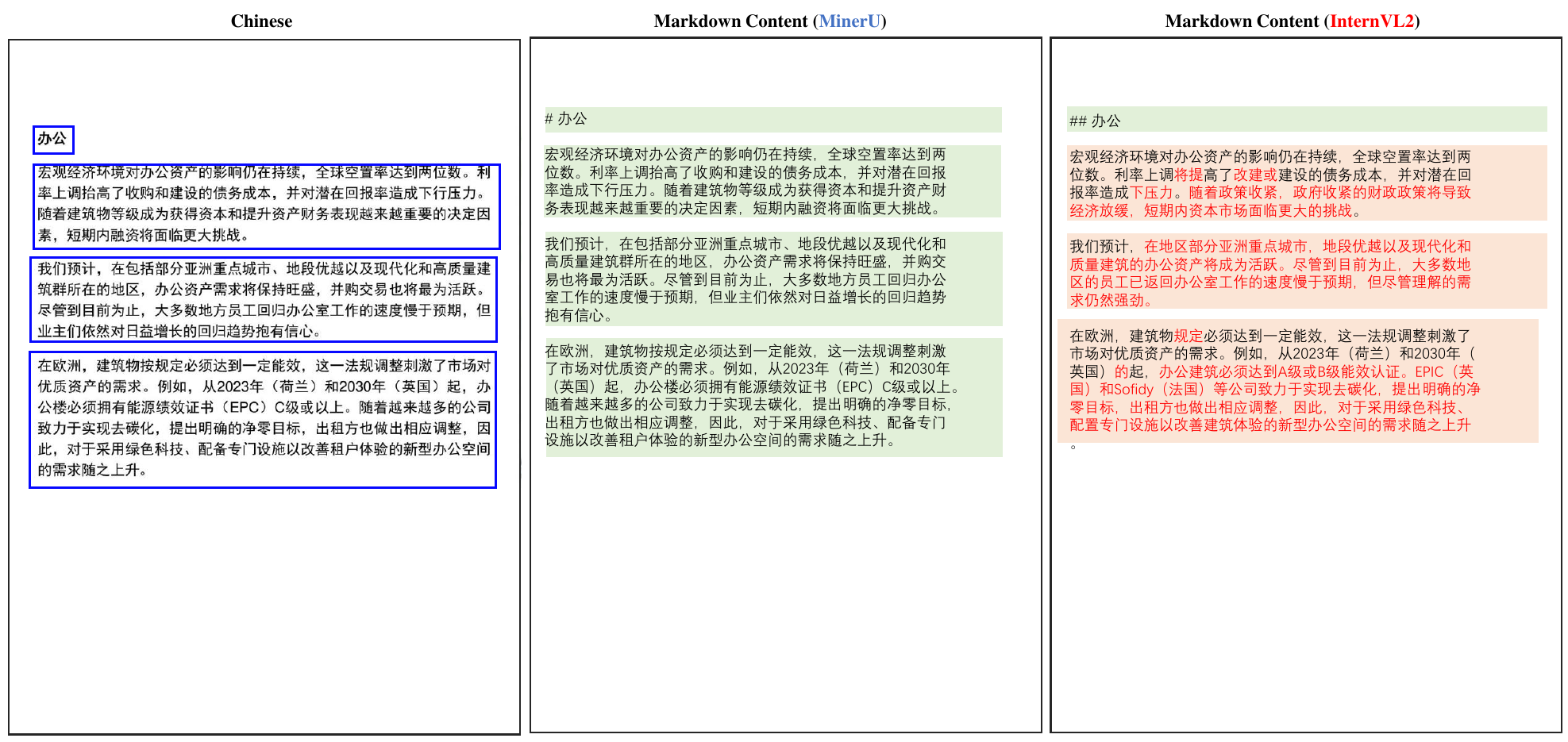}
    \caption{The \textcolor{blue}{Good} Model Result and \textcolor{red}{Bad} Model Result for Text Language in Chinese.}
    \label{fig:text_chinese}
\vspace{-5mm}
\end{figure*}

\begin{figure*}[t]
    \centering
    \includegraphics[width=1\textwidth]{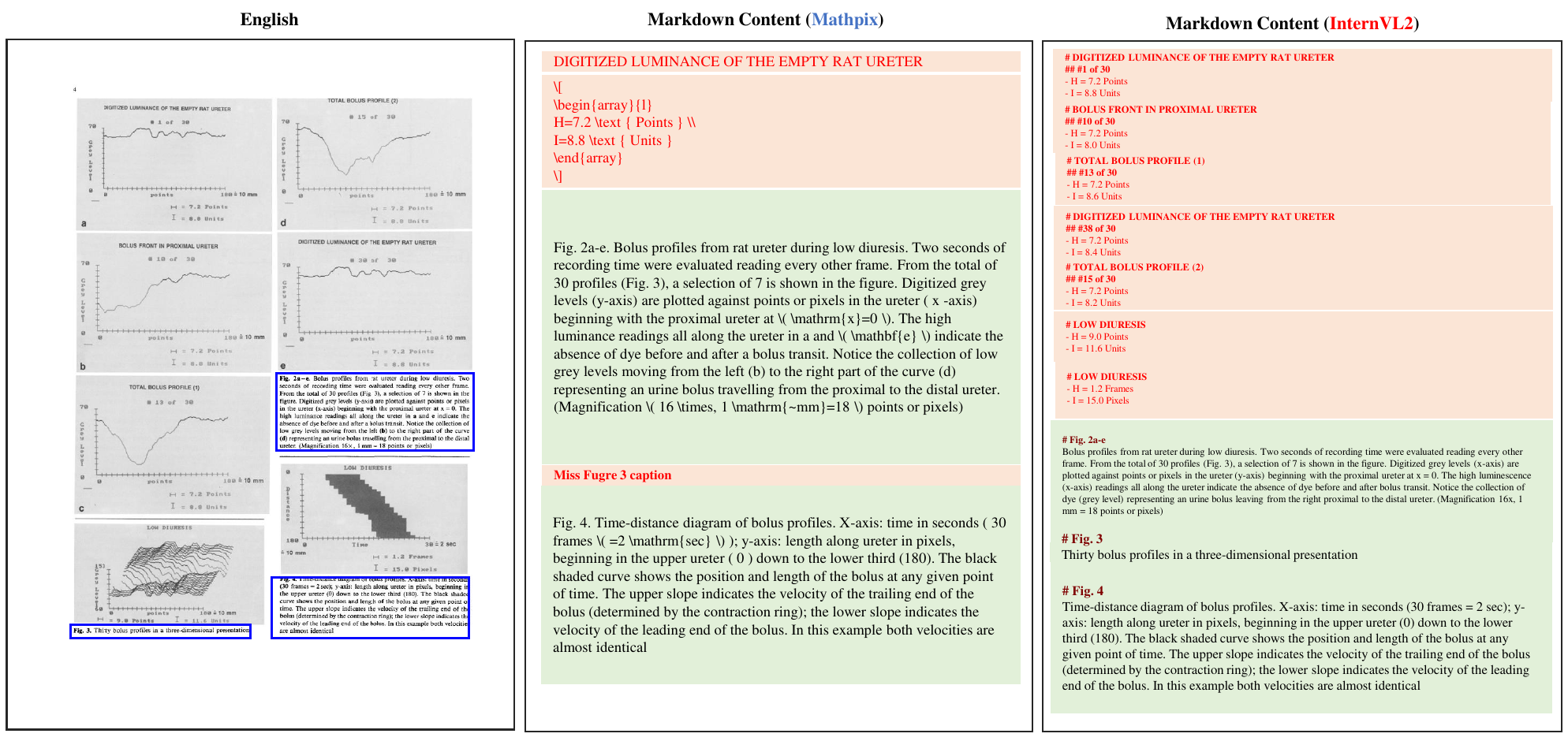}
    \caption{The \textcolor{blue}{Good} Model Result and \textcolor{red}{Bad} Model Result for Text Language in English.}
    \label{fig:text_english}
\vspace{-5mm}
\end{figure*}

\begin{figure*}[t]
    \centering
    \includegraphics[width=1\textwidth]{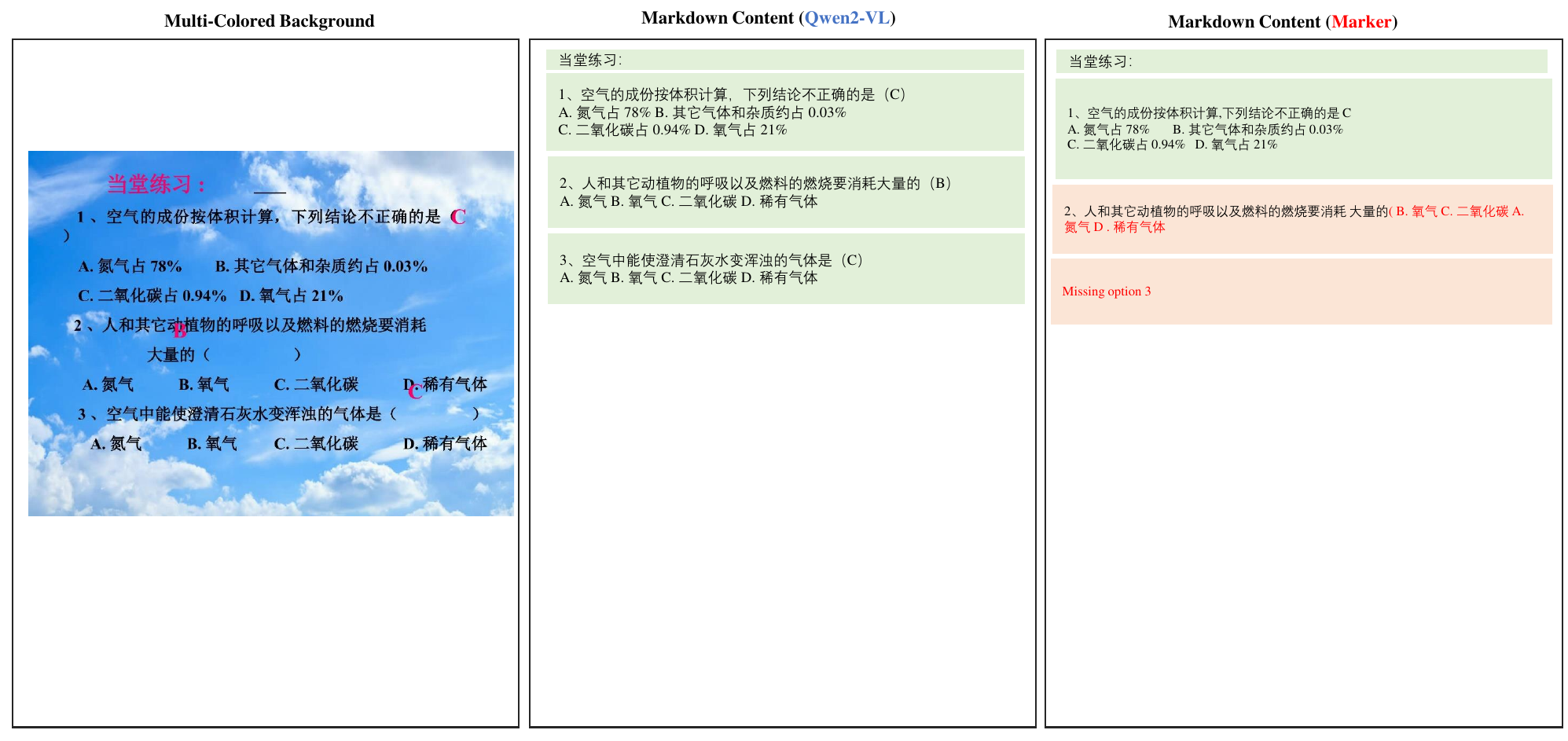}
    \caption{The \textcolor{blue}{Good} Model Result and \textcolor{red}{Bad} Model Result for Text with Colorful Background.}
    \label{fig:text_colorful_background}
\vspace{-5mm}
\end{figure*}

\begin{figure*}[t]
    \centering
    \includegraphics[width=1\textwidth]{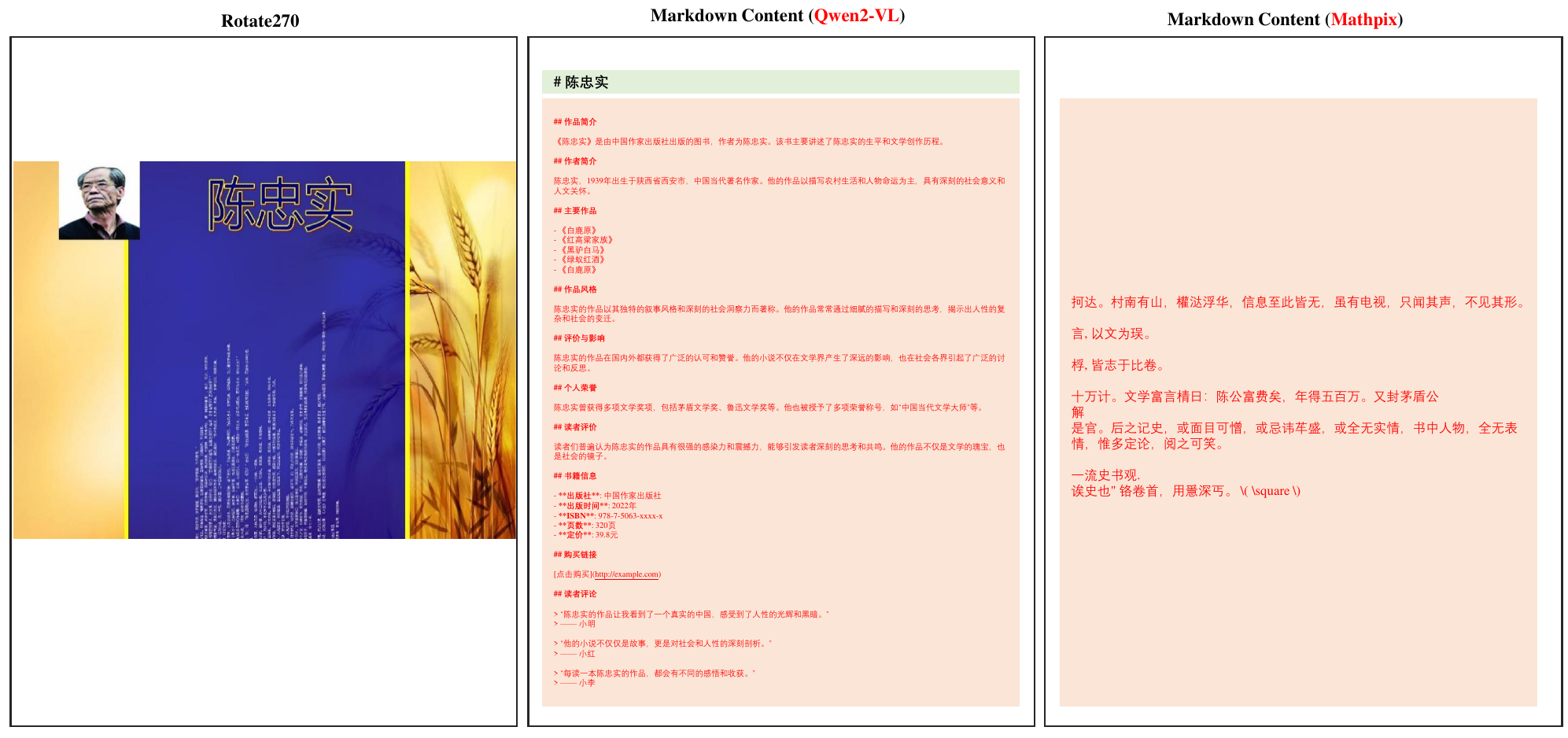}
    \caption{The \textcolor{red}{Bad} Model Result for Text with Rotation.}
    \label{fig:text_rotate}
\vspace{-5mm}
\end{figure*}

\begin{figure*}[t]
    \centering
    \includegraphics[width=0.8\textwidth]{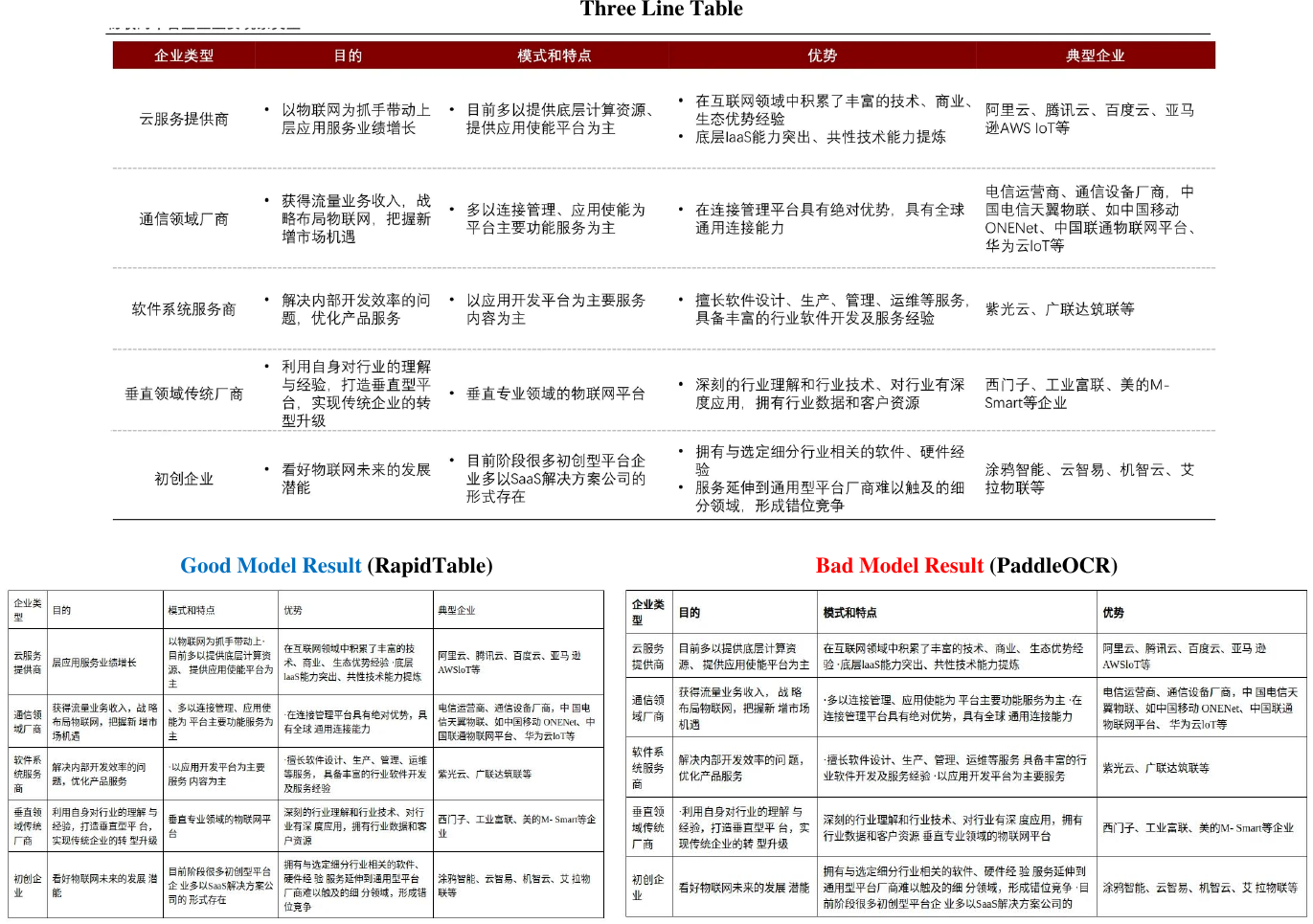}
    \caption{The \textcolor{blue}{Good} Model Result and \textcolor{red}{Bad} Model Result for Three Line Frame Table.}
    \label{fig:badcase_table_threeline}
\vspace{-5mm}
\end{figure*}

\begin{figure*}[t]
    \centering
    \includegraphics[width=0.8\textwidth]{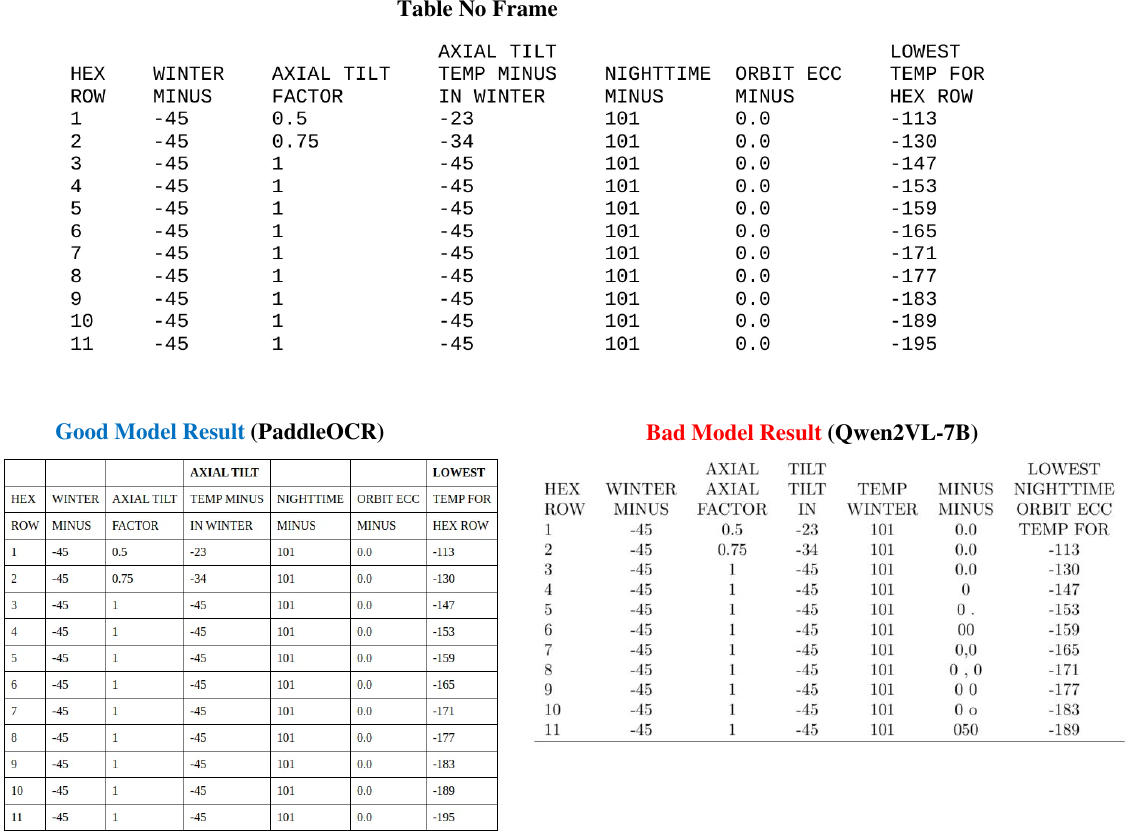}
    \caption{The \textcolor{blue}{Good} Model Result and \textcolor{red}{Bad} Model Result for No Frame Table.}
    \label{fig:badcase_table_noframe}
\vspace{-5mm}
\end{figure*}

\begin{figure*}[t]
    \centering
    \includegraphics[width=0.8\textwidth]{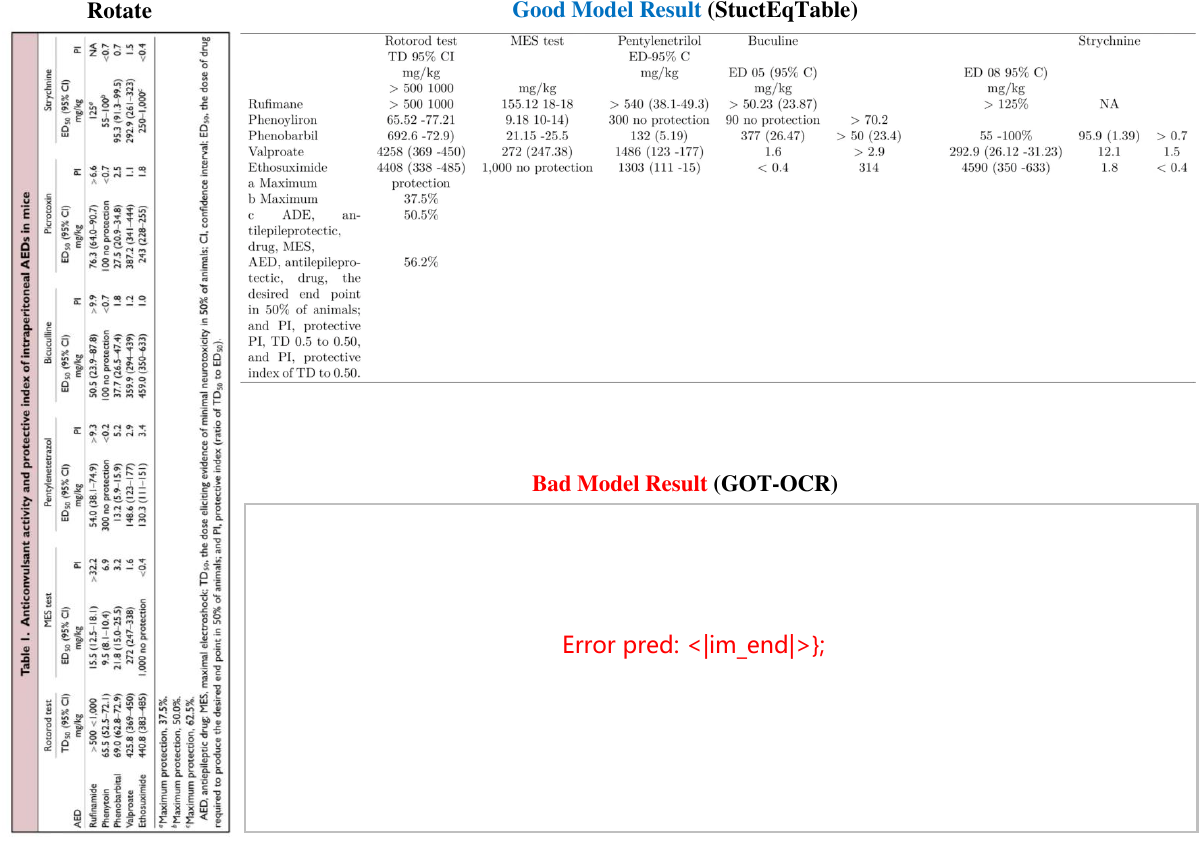}
    \caption{The \textcolor{blue}{Good} Model Result and \textcolor{red}{Bad} Model Result for Rotated Table.}
    \label{fig:badcase_table_rotate}
\vspace{-5mm}
\end{figure*}

\begin{figure*}[t]
    \centering
    \includegraphics[width=1\textwidth]{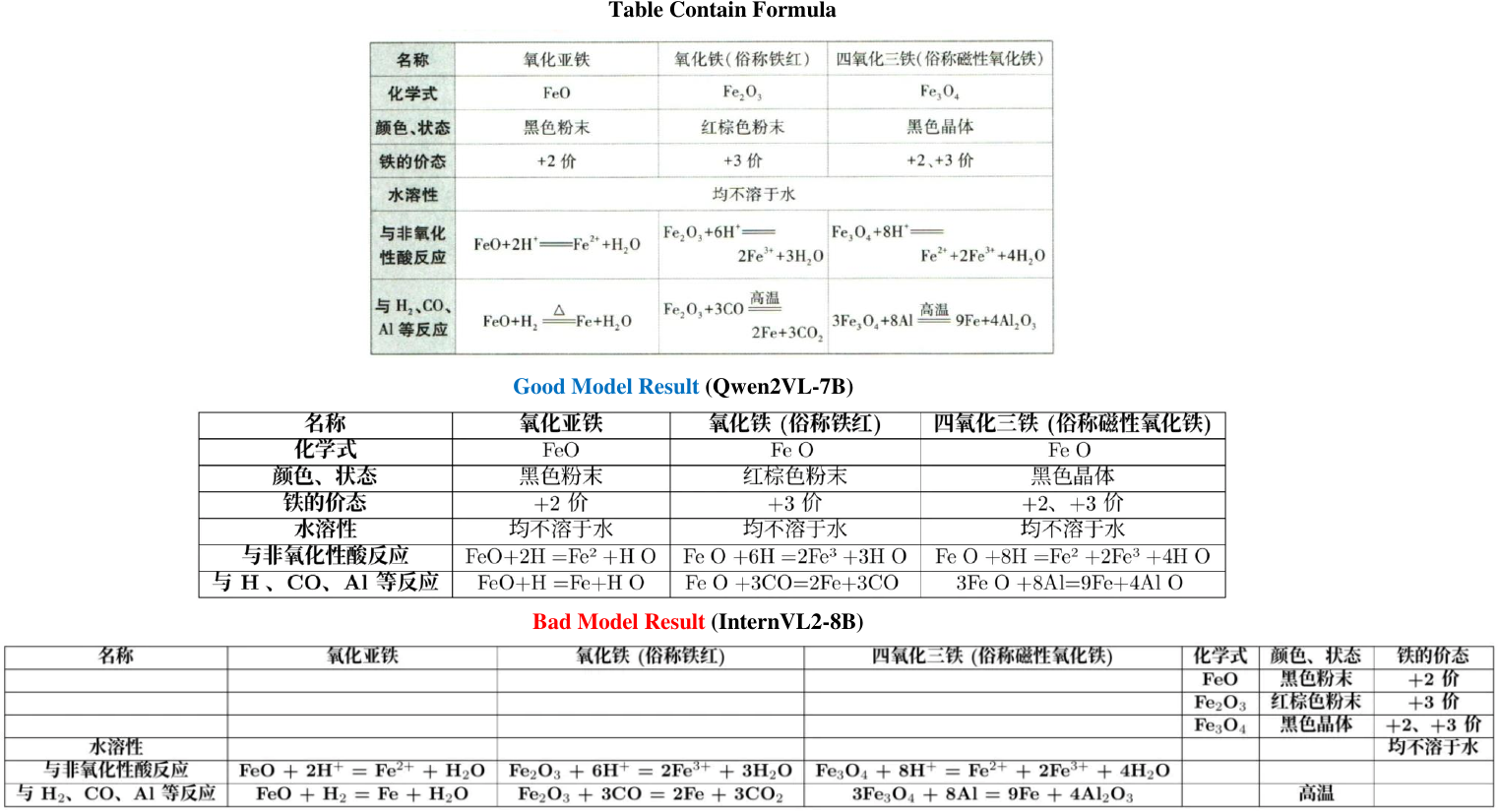}
    \caption{The \textcolor{blue}{Good} Model Result and \textcolor{red}{Bad} Model Result for Table with Formula.}
    \label{fig:badcase_table_formula}
\vspace{-5mm}
\end{figure*}

\end{document}